\documentclass[Afour,sageh,times]{sagej}
\usepackage{moreverb,url}
\usepackage{algorithm}
\usepackage{algpseudocode}
\usepackage{booktabs}
\usepackage{multirow}
\usepackage[table]{xcolor}
\usepackage{pifont}
\usepackage{makecell}
\newcommand{\cmark}{\ding{51}} % check
\newcommand{\xmark}{\ding{55}} % cross
\algrenewcommand\algorithmicrequire{\textbf{Input:}}
\algrenewcommand\algorithmicensure{\textbf{Output:}}
\usepackage[colorlinks,bookmarksopen,bookmarksnumbered,citecolor=red,urlcolor=red]{hyperref}

\newcommand\BibTeX{{\rmfamily B\kern-.05em \textsc{i\kern-.025em b}\kern-.08em
T\kern-.1667em\lower.7ex\hbox{E}\kern-.125emX}}

\makeatletter
\renewcommand\paragraph[1]{\textbf{#1}}
\makeatother

\renewcommand{\titlesize}{\fontsize{17}{19pt}\selectfont} % slightly larger title font while keeping one-line fit

\begin{document}

\runninghead{Wu et al.}

\title{Synergizing Efficiency and Reliability for\\Continuous Mobile Manipulation
}

% todo: check all

\author{Chengkai Wu\affilnum{1}, 
    Ruilin Wang\affilnum{2},
    Yixin Zeng\affilnum{2},
    Jiayuan Wang\affilnum{2},
    Mingjie Zhang\affilnum{1},
    Guiyong Zheng\affilnum{2},
    Qun Niu\affilnum{2},
    Juepeng Zheng\affilnum{2},
    Jun Ma\affilnum{1}, and 
    Boyu Zhou\affilnum{3}
    \vspace{-1cm}
    }

\affiliation{\affilnum{1}Robotics and Autonomous Systems Thrust, The Hong Kong University of Science and Technology (Guangzhou), Guangzhou, China\\
\affilnum{2}School of Artificial Intelligence, Sun Yat-sen University, Zhuhai, China\\
\affilnum{3}Department of Mechanical and Energy Engineering, Southern University of Science and Technology, Shenzhen, China\\
% \affilnum{4}Differential Robotics, Hangzhou, China\\
Ruilin Wang and Yixin Zeng contributed equally to this work.\\
Chengkai Wu, Ruilin Wang and Yixin Zeng are core contributors.
}

\corrauth{Boyu Zhou, Southern University of Science and Technology, China.}
\email{zhouby@sustech.edu.cn}

\begin{abstract}
Humans seamlessly fuse anticipatory planning with immediate feedback to perform successive mobile manipulation tasks without stopping, achieving both high efficiency and reliability. Replicating this fluid and reliable behavior in robots remains fundamentally challenging, not only due to conflicts between long-horizon planning and real-time reactivity, but also because excessively pursuing efficiency undermines reliability in uncertain environments: it impairs stable perception and the potential for compensation, while also increasing the risk of unintended contact. 
In this work, we present a unified framework that synergizes efficiency and reliability for continuous mobile manipulation. It features a reliability-aware trajectory planner that embeds essential elements for reliable execution into spatiotemporal optimization, generating efficient and reliability-promising global trajectories. It is coupled with a phase-dependent switching controller that seamlessly transitions between global trajectory tracking for efficiency and task-error compensation for reliability. We also investigate a hierarchical initialization that facilitates online replanning despite the complexity of long-horizon planning problems. Real-world evaluations demonstrate that our approach enables efficient and reliable completion of successive tasks under uncertainty (e.g., dynamic disturbances, perception and control errors). Moreover, the framework generalizes to tasks with diverse end-effector constraints. Compared with state-of-the-art baselines, our method consistently achieves the highest efficiency while improving the task success rate by 26.67\%--81.67\%. Comprehensive ablation studies further validate the contribution of each component. The source code will be released.
\vspace{-.3cm}
\end{abstract}


\maketitle

\section{Introduction}
Mobile Manipulators (MMs) hold great potential for scalable automation across manufacturing~\cite{johns2023framework}, healthcare~\cite{zhang2022learning}, and scientific laboratories~\cite{burger2020mobile,dai2024autonomous}. However, deploying MMs for continuous, tightly arranged tasks in complex real-world environments faces a fundamental challenge: replicating the human-like synergy of efficiency and reliability. Efficiency requires anticipatory planning of long-horizon whole-body motions across multiple tasks to minimize the overall execution time of successive tasks. Reliability, in contrast, demands (i) the capability for real-time error compensation to mitigate misalignment between the end-effector and the desired operational pose, caused by inaccurate environmental priors, perception or control errors, and even target movement, as well as (ii) the avoidance of unintended interactions between the end-effector, the target, and the environment. However, existing paradigms struggle to reconcile these two seemingly incompatible objectives, often encountering three critical bottlenecks.

\begin{figure}[h]
\centering
\includegraphics[width=0.95\linewidth]{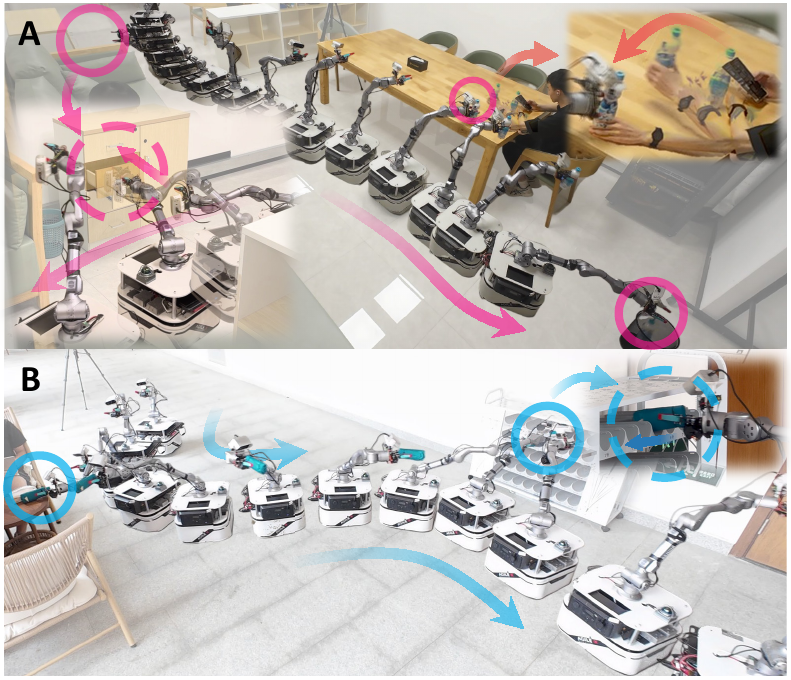}
\vspace{-0.1cm}
\caption{Teaser. Continuous on-the-move mobile manipulation for tightly arranged task sequences, synergizing efficiency and reliability under real-world uncertainty using onboard sensing.
}
\label{fig:head}
\vspace{-0.8cm}
\end{figure}

\begin{figure*}[t]
\centering
\includegraphics[width=0.95\linewidth]{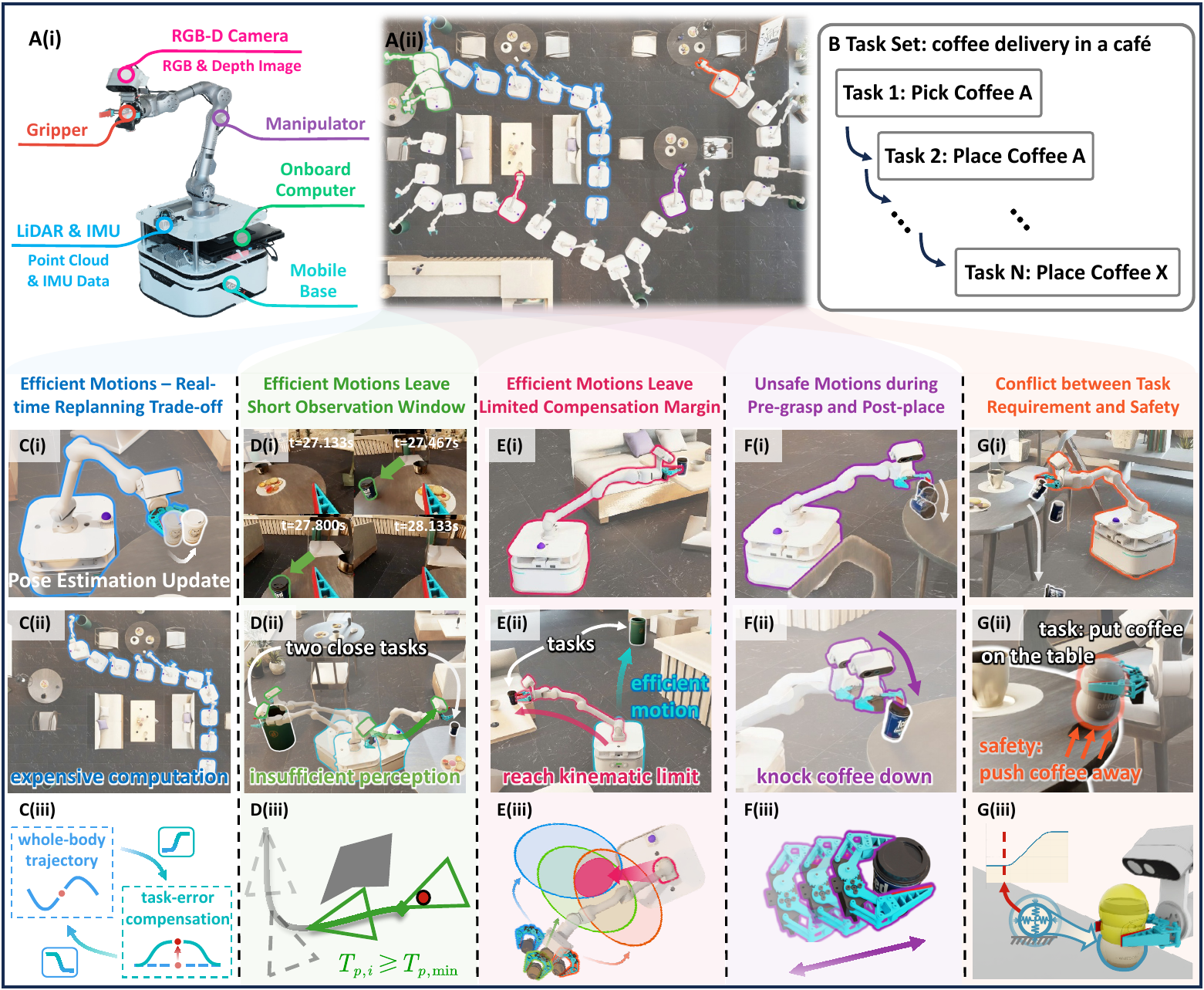}
\vspace{-0.1cm}
\caption{Overview of challenges and solutions for efficient and reliable mobile manipulation.
(\textbf{A}) (i) A mobile manipulator robot capable of coupled locomotion, manipulation, and perception, and (ii) a composite top-view of the execution trajectory for a mobile manipulator performing continuous tasks, where colored segments correspond to specific challenges in (C)–(G). 
(\textbf{B}) Example task set.
(\textbf{C})--(\textbf{G}) illustrate five challenges while pursuing both efficiency and reliability (rows i--ii) and our corresponding solutions (row iii): (\textbf{C}) To bridge long-horizon planning and real-time reactivity, we propose a high-frequency controller that bridges whole-body trajectory tracking with real-time task-error compensation using a phase-dependent switching scheme. (\textbf{D}) To prevent insufficient observation windows due to fast motion, we introduce Time-assured Active Perception to ensure sufficient observation time. (\textbf{E}) To avoid limited compensation margins while pursuing efficiency, we present Compensation Margin Zone. (\textbf{F}) To resolve unsafe end-effector motions, we propose Efficient Safe Interaction for pre-grasp and post-place phases. (\textbf{G}) To handle conflicts between task and safety requirements, we propose Elastic Collision Spheres.
}
\label{fig:problem_overview}
\vspace{-0.4cm}
\end{figure*}

First, a fundamental trade-off exists between long-horizon planning and reactivity (Fig.~\ref{fig:problem_overview}C). Efficiency-oriented planners generate long-horizon whole-body trajectories that consider multiple tasks to minimize inter-task transition time~\cite{thakar2018towards,thakar2020manipulator,zimmermann2021go,reister2022combining}. However, their high computational complexity prevents reactive planning, making them vulnerable to real-world uncertainties (e.g., uncertainties in target poses, control errors, and dynamic disturbances). In contrast, high-frequency reactive controllers prioritize local reactivity to compensate end-effector error in real-time~\cite{haviland2022holistic,burgess2023architecture,wang2025ehc}. But they typically operate on short horizons, leading to greedy behavior that significantly reduces global efficiency.
Second, the pursuit of efficiency introduces two critical flaws that undermine reliable task execution. To minimize task time, efficiency-oriented trajectories often result in insufficient observation windows for modern perception models~\cite{wen2024foundationpose} to estimate accurate target poses (Fig.~\ref{fig:problem_overview}D), leaving the robot to execute based on stale pose information and thereby causing misalignment between the end-effector and the target. Furthermore, such trajectories often drive MMs close to their kinematic limits (Fig.~\ref{fig:problem_overview}E), eliminating the kinematic margins required to compensate for end-effector errors induced by the real-world uncertainties.
Third, safely establishing necessary contact without sacrificing efficiency is a key bottleneck. To avoid unintended collisions among the end-effector, target objects, and the environment (Fig.~\ref{fig:problem_overview}F-G), most methods use rigid end-effector motion primitives that cause unnecessary halts for MMs~\cite{thakar2018towards,reister2022combining,ShenW-RSS-25}, disrupting operational continuity and reducing overall efficiency. Moreover, standard collision-avoidance models struggle to handle tasks requiring intentional environmental contact (e.g., placing the bottle on the table)~\cite{ichnowski2020deep} and exhibit two critical failure modes: (i) overly conservative behavior, which misclassifies intentional contact as a dangerous collision and blocks MMs from reaching the desired task pose, or (ii) overly aggressive behavior, which causes collisions between the manipulated object and the environment during transport, even leading to object slippage from the gripper.

To break this efficiency-reliability dilemma, we propose a unified framework for continuous mobile manipulation in complex environments, which minimizes the time required for successive tightly arranged tasks while maintaining high reliability under real-world uncertainties. Rather than treating efficiency and reliability as a zero-sum trade-off, our framework explicitly embeds reliability awareness into the spatiotemporal trajectory optimization paradigm. During execution, a phase-dependent switching controller (Fig.~\ref{fig:problem_overview}C(iii)) seamlessly transitions between (i) global trajectory tracking during non-critical phases for efficiency and (ii) real-time error compensation during task-critical phases to improve reliability under real-world uncertainties, while preserving the safe interaction structure and overall efficiency of the planned trajectory.

% The two components synergize reciprocally, allowing previously incompatible capabilities to coexist.

The reliability-aware long-horizon planner formalizes efficiency maximization as a constrained spatiotemporal optimization problem and employs a compact trajectory representation for mobile manipulators. To address reliability gaps in efficiency-oriented planning, we encode reliability awareness via meticulously designed differentiable metrics that incorporate core requirements. Specifically, the Time-assured Active Perception strategy (Fig.~\ref{fig:problem_overview}D(iii)) ensures adequate onboard perception for reliable target pose estimation. To effectively capture the potential of corrective motions, Compensation Margin Zone (CMZ) (Fig.~\ref{fig:problem_overview}E(iii)) is designed to promote compensation capability within kinematic limits. For object contact safety, two mechanisms are adopted: (i) Efficient Safe Interaction (ESI) (Fig.~\ref{fig:problem_overview}F(iii)) optimizes pre-grasp/post-placement motions to avoid end-effector–object collisions without redundant stops; (ii) Elastic Collision Sphere (ECS) (Fig.~\ref{fig:problem_overview}G(iii)) with a contact-adaptive radius enables precise, safe, and efficient manipulation during object–support surface contact. These constraints are flexibly blended through spatiotemporal joint optimization, establishing a reliability-aware optimization paradigm that prioritizes efficiency while promoting reliable execution. 

The above optimization is computationally intensive due to the high dimensionality of the MM, long planning horizon, and complex nonlinear constraints. For online solving, we design a hierarchical initialization strategy that efficiently generates high-quality feasible initializations. First, a Sequential-Progress Hybrid A* algorithm generates a feasible base path, ensuring the end-effector reaches all keypoints, where a tailored reachability map and a progress-aware function accelerate reachability checks and search progress. Subsequently, the manipulator path is searched over a compact configuration graph that captures potential safe motions between key base waypoints.

The proposed framework is rigorously validated in challenging real-world and simulation settings, demonstrating effective regulation of efficiency and reliability in continuous mobile manipulation (Extension~1). Specifically, we deployed the robot in tightly arranged tasks across two scenarios: a confined office lounge and a dynamic setting with moving obstacles and a non-stationary target. Its robustness was further verified through an uninterrupted marathon test, during which the robot completed consecutive tasks with ad hoc object placements without failure. Furthermore, the framework extends to complex skills across varied end-effector trajectories, such as valve turning and umbrella insertion into a stand hole. Extensive benchmarks against state-of-the-art (SOTA) methods show our framework yields significantly more fluid, time-efficient motions and higher reliability. Furthermore, comprehensive ablation studies confirm the necessity of each component. In summary, our main contributions are as follows:

\begin{enumerate}
  \item[1)] A unified hierarchical framework for continuous mobile manipulation that systematically reconciles operational efficiency and execution reliability.

  \item[2)] A reliability-aware spatiotemporal optimization formulation for time-efficient long-horizon whole-body trajectory planning, while promoting reliable execution.

  \item[3)] A hierarchical initialization strategy for online long-horizon trajectory optimization under high dimensionality and complex nonlinear constraints.

  \item[4)] A safe-warping-based phase-dependent execution layer bridging long-horizon trajectory tracking and real-time task-error compensation while preserving the planned safe interaction structure.

  \item[5)] Comprehensive real-world and simulation experiments, demonstrating the superior performance and validity of the proposed framework. The source code will be made publicly available.
\end{enumerate}

% \begin{enumerate}
%   \item[3)] A hierarchical online solving strategy, consisting of Sequential-Progress Hybrid A* and layered-graph manipulator search, that provides high-quality feasible initialization for the proposed optimization despite the high dimensionality and complex constraints of mobile manipulators.

% More recent works extend this idea to successive or long-horizon task settings, exploiting kinematic redundancy to overlap task execution with preparation for upcoming tasks, thereby reducing transition time and overall mission duration.

\section{Related Work}
\subsection{Efficiency-oriented Long-Horizon Planning}
% Whole-body motion planning for mobile manipulators has been widely studied to coordinate the mobile base and the manipulator for efficient mobile manipulation. Several studies~\cite{zucker2013chomp,schulman2014motion,thakar2020accelerating,spahn2021coupled,wu2024real,deng2025whole,xu2025topay} focus primarily on the execution of isolated, single tasks, such as reaching a goal state. These methods simultaneously optimize whole-body decision variables to generate coordinated motion of the base and the arm to reach a specific target. These approaches improve execution efficiency compared to decoupled planning~\cite{pilania2015hierarchical,wu2023tidybot}, which generates motions where one component must stop while the other is moving. Despite achieving coordinated motion during the reaching phase, these methods typically cause the system to stop upon reaching the goal. This behavior is highly inefficient for real-world applications that require executing sequential mobile manipulation tasks.

To achieve efficient task execution, early methods typically adopt long-horizon whole-body motion planning to coordinate the mobile base and manipulator~\cite{zucker2013chomp,schulman2014motion,thakar2020accelerating,spahn2021coupled,wu2024real,deng2025whole,xu2025topay}, which avoids redundant time consumption caused by decoupled planning~\cite{pilania2015hierarchical,wu2023tidybot}, where the mobile base and manipulator move separately. However, these methods primarily focus on optimizing execution for single tasks, resulting in low efficiency in sequential mobile manipulation tasks. The inefficiency is mainly due to the failure to fully leverage the kinematic redundancy of the mobile manipulator to (i) reduce the base movement distance between tasks and (ii) avoid increased total operation time caused by the mobile base coming to a halt during task execution. To address these two key inefficiencies, recent works have shifted toward sequential mobile manipulation, leveraging the robot’s kinematic redundancy to target both aspects. Specifically, to tackle the first aspect (reducing inter-task base movement distance), some methods optimize base placements so that a single placement supports as many subsequent tasks as possible, thereby reducing the need for repeated base repositioning~\cite{xu2020planning,xu2021planning}. Others determine the optimal next placement by integrating manipulation costs with navigation costs evaluated over the next two consecutive tasks~\cite{reister2022combining,burgess2024reactive}. Both strategies minimize unnecessary base movement and improve overall efficiency. To address the second aspect (avoiding base halts during task execution), other planners generate continuous paths that not only minimize total base travel distance but also enable the mobile base to keep moving toward the next task while executing the current one, thereby eliminating time losses from unnecessary stops~\cite{thakar2018towards,thakar2020manipulator,zimmermann2021go,burgess2024reactive}.

Despite significant efficiency gains, these methods have critical limitations in real-world deployments, often falling into two extremes. On one hand, long-horizon whole-body planners incur high computational overhead that inherently restricts reactive planning~\cite{thakar2018towards,thakar2020manipulator,zimmermann2021go,reister2022combining}. Unable to adjust trajectories in real time when task poses are updated online, the robot is forced to execute outdated trajectories generated from stale task-pose information, which can ultimately lead to task failure. On the other hand, methods that pursue real-time performance often achieve it by severely simplifying the problem, such as solely optimizing the base path while completely ignoring the manipulator’s 3D kinematic feasibility and collision avoidance~\cite{burgess2024reactive}. This decoupled over-simplification frequently leads to unreachable target poses or unexpected collisions in complex environments.

% More crucially, these efficiency-driven planners tend to drive the mobile manipulator close to its kinematic limits to maximize global efficiency. By doing so, they inadvertently eliminate the local compensatory margin needed for online error correction. Consequently, they optimize for efficiency at the severe expense of execution reliability, highlighting the fundamental need for a framework that inherently preserves compensation margins during long-horizon planning.

Moreover, these efficiency-driven planners tend to drive the manipulator close to its kinematic limits to maximize global efficiency. In this process, they unintentionally eliminate the compensation margin required for online error correction. While some approaches~\cite{haviland2022holistic} attempt to retain kinematic flexibility by maximizing classical manipulability indices~\cite{yoshikawa1985manipulability}, merely driving the manipulator away from singularities only guarantees instantaneous end-effector velocity generation. However, it fails to ensure the capability to compensate for finite end-effector errors, necessitating explicit consideration of both joint limits and configuration.

\subsection{Reliability-oriented Reactive Execution}

To address the inherent task-pose uncertainties in real-world environments, existing literature typically employs active perception and reactive control. Active perception strategies aim to acquire observations of the target first, then utilize perception algorithms to estimate the true target pose before task execution, thereby enabling successful manipulation of the object~\cite{reister2022combining,burgess2023architecture,burgess2024reactive,zhang2024gamma,jauhri2024active}. However, they often force the robot to come to a full stop to acquire stable observations, which severely degrades operational efficiency~\cite{reister2022combining,zhang2024gamma,jauhri2024active}. Conversely, methods that perform active perception during approaching do not require the robot to come to a halt, thus improving operational efficiency~\cite{burgess2023architecture,burgess2024reactive}. However, their efficient motion inherently compresses the time available for sensor data accumulation, often failing to provide sufficient observation windows for stable pose estimation. This is particularly problematic because modern 6D pose estimators~\cite{wen2024foundationpose,liang2025dynamicpose} rely on stable and sufficient observation windows to initialize and refine accurate predictions. Without such observation windows, these estimators cannot function properly, thereby compromising the overall reliability of task execution.

% To ensure reliable manipulation, the system must be able to compensate in real time for the error between the actual gripper pose and the desired pose derived from the latest target estimate.

% To achieve precise manipulation and thereby enhance manipulation reliability, it is necessary to compensate for the error between the actual gripper pose and the desired gripper pose derived from the latest target estimates. This error is dynamically time-varying, arising from perception, control, and task-pose errors. Consequently, equipping the system with the capability to adjust the mobile manipulator’s motions in real time to compensate for such errors is crucial for improving reliability. 

To compensate for end-effector error caused by coarse task priors, perception errors, and control errors, existing approaches rely on reactive controllers (typically operating above 20 Hz)~\cite{haviland2022holistic,du2023hierarchical,burgess2023architecture,wang2024arm,wang2025ehc,spahn2023dynamic,spahn2024demonstrating}. While highly reliable for reaching target end-effector poses, these controllers inherently operate over short horizons and generate greedy control behaviors that completely ignore global task efficiency. Although attempts~\cite{burgess2024reactive} have been made to integrate reactive long-horizon base controllers~\cite{missura2022fast} to improve efficiency, maintaining real-time performance over extended horizons forces these methods to plan base motions without considering the manipulator's kinematic feasibility, leading to task failures or collisions in complex 3D environments.

\subsection{Safe Interaction during Manipulation}
% Ensuring contact safety without sacrificing fluid motion is a major bottleneck in continuous mobile manipulation. During the critical pre-grasp approach and post-place retraction phases, minor deviations can cause the end-effector to collide with and topple the target. To mitigate this, many approaches~\cite{thakar2018towards,reister2022combining,ShenW-RSS-25,vosylius2024instant} enforce handcrafted motion sequences, deliberately halting the base to execute isolated arm insertions. While safe, these stops severely interrupt momentum and degrade efficiency. Furthermore, standard collision avoidance constraints struggle with intended-contact tasks. Treating the target rigidly as an obstacle~\cite{thakar2020manipulator} drastically narrows the feasible solution space, while attaching the object collision model to the robot~\cite{ichnowski2020deep} inevitably triggers false-positive collisions with the support surface upon placement.

Beyond reaching the target, achieving safe contact without compromising smooth motion remains a major bottleneck in sequential mobile manipulation. During the critical pre-grasp approach and post-placement retraction phases, failing to account for collisions between the gripper and the manipulated object can easily cause the end-effector to collide with and knock over the target~\citet{burgess2024reactive}. To mitigate this, many approaches enforce hand-crafted, rigid motion primitives, including deliberately halting to execute isolated end-effector approach and retraction motions~\cite{thakar2018towards,reister2022combining,ShenW-RSS-25,vosylius2024instant}. While these primitives ensure safety, the resulting stops disrupt the robot’s operational continuity and reduce overall efficiency.

Furthermore, standard collision-avoidance constraints struggle to handle tasks that require intentional contact. Treating the target object rigidly as an obstacle significantly narrows the feasible solution space and increases computation time when the robot needs to grasp the object~\cite{thakar2020manipulator}. Alternatively, attaching the object’s collision model to the robot’s kinematic chain during transport inherently triggers false-positive collisions with the support surface upon placement, preventing the end-effector from reaching the desired operational pose~\cite{ichnowski2020deep}. Therefore, an adaptive collision handling mechanism is required to flexibly manage intentional contacts and actual obstacles without relying on inefficient, hand-crafted stops.

\section{Task Model and Problem Statement}

\subsection{Task Model}
\begin{figure}[h]
\centering
\includegraphics[width=0.85\linewidth]{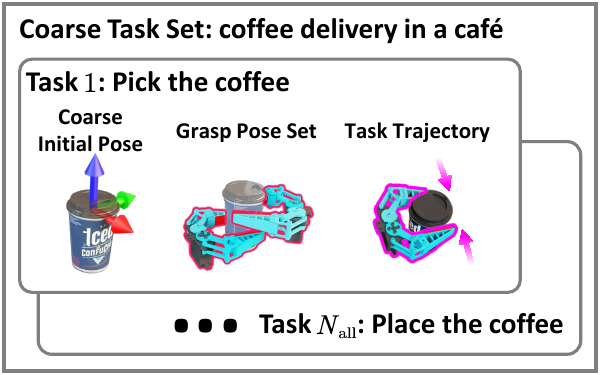}
\vspace{-0.1cm}
\caption{Illustration of Task Model.
}
\label{fig:task_model}
\vspace{-0.1cm}
\end{figure}

Denote $\boldsymbol{\mathcal{T}}_\mathrm{all}=\{\mathcal{T}_1,\dots,\mathcal{T}_{N_{\mathrm{all}}}\}$ as an ordered set of $N_{\mathrm{all}}$ coarse object-centric mobile manipulation tasks (Fig.~\ref{fig:task_model}). Each task $\mathcal{T}_i$ is specified by three components: (1) the coarse initial pose of the manipulated object $\boldsymbol{P}_{\mathrm{o}_i}\in\mathrm{SE}(3)$ (relative to the world frame); (2) a set of feasible end-effector grasp poses $\boldsymbol{\mathcal{C}}_i\subset\mathrm{SE}(3)$ (defined relative to the object frame); (3) a task trajectory, consisting of an object-centric pose trajectory $\,^{\mathrm{o}_i}\boldsymbol{P}_i(\tau)\in\mathrm{SE}(3)$ (expressed in the object’s initial frame), together with a binary end-effector command trajectory $s_i(\tau)\in\{0,1\}$ (0 for open, 1 for closed), both defined over a local task time horizon $\tau\in[0, T_{\mathcal{T},i}]$~\cite{huang2024rekep,hsu2025spot,pan2025omnimanip}. 

The coarse object pose $\boldsymbol{P}_{\mathrm{o}_i}$ serves as an initial reference, while the unknown but true object pose is denoted by $\boldsymbol{P}_{\mathrm{o}_i}^\star$. For tasks that require object pose estimation (e.g., grasping tasks), $\boldsymbol{P}_{\mathrm{o}_i}^\star$ must be estimated before execution to ensure reliable task completion. We assume bounded pose uncertainty, under which viewpoints planned from the coarse object pose remain valid for online estimation of the actual pose. During execution, the latest pose estimate is denoted by $\hat{\boldsymbol{P}}_{\mathrm{o}_i}$, with $\hat{\boldsymbol{P}}_{\mathrm{o}_i}=\boldsymbol{P}_{\mathrm{o}_i}$ at initialization. The corresponding task trajectory in the world frame is given by 
\begin{equation}
    \boldsymbol{P}_{\mathrm{t}, i}(\tau):=\hat{\boldsymbol{P}}_{\mathrm{o}_i}\,\,^{\mathrm{o}_i}\boldsymbol{P}_i(\tau), \tau\in[0,T_{\mathcal{T},i}].
\end{equation}
We assume that the resulting task trajectory $\boldsymbol{P}_{\mathrm{t},i}(\tau)$ is safe for the manipulated object. In particular, an instant pick or place task is modeled by setting $T_{\mathcal{T},i}=0$, which reduces the task trajectory to consist of one single task pose $\boldsymbol{P}_{\mathrm{t}, i}(0)=\hat{\boldsymbol{P}}_{\mathrm{o}_i}$ and one single gripper command $s_i(0)$. Given a feasible grasp ${}^{\mathrm{o}_i}\boldsymbol{P}_{\mathcal{C},i}\in\boldsymbol{\mathcal{C}}_i$, define $\bar{\boldsymbol{P}}_{\mathrm{t},i}(\tau):=\boldsymbol{P}_{\mathrm{t},i}(\tau)\,{}^{\mathrm{o}_i}\boldsymbol{P}_{\mathcal{C},i}$ as the task end-effector trajectory, which induces the desired object motion $\boldsymbol{P}_{\mathrm{t},i}(\tau)$.

\subsection{Problem Statement}
We consider a mobile manipulator robot capable of coupled locomotion, manipulation, and perception, operating in a 3D workspace with obstacles $\mathcal{M}_{\mathrm{env}}\subset\mathbb{R}^3$ (Fig.~\ref{fig:problem_overview}A). The mobile manipulator consists of a mobile base, a manipulator with an end-effector, and an onboard perception sensor (e.g., RGB-D camera). Its nonlinear dynamics are given by
\begin{equation*}
    \dot{\boldsymbol{x}}=f_{\mathrm{MM}}(\boldsymbol{x},\boldsymbol{u}),
\end{equation*}
where $\boldsymbol{x} \in\mathcal{X}$ and $\boldsymbol{u}\in\mathcal{U}$ are the whole body states and control inputs. $\mathcal{X}$ and $\mathcal{U}$ denote the state and input space. 
Define $\boldsymbol{x}_\mathrm{init}$ as the initial state. 
Let $f_{\mathrm{FK},\mathrm{ee}}:\mathcal{X}\rightarrow \mathrm{SE}(3)$ and $f_{\mathrm{FK},\mathrm{c}}:\mathcal{X}\rightarrow \mathrm{SE}(3)$ denote the forward-kinematics maps that return the poses of the end-effector and the onboard perception sensor in the world frame, respectively. 

\paragraph{Problem.}
Given the environment map $\mathcal{M}_{\mathrm{env}}$, the ordered task set $\boldsymbol{\mathcal{T}}_\mathrm{all}$, and the initial state $\boldsymbol{x}_{\mathrm{init}}$, our goal is to compute a whole-body state-input trajectory pair $(\boldsymbol{x}(t),\boldsymbol{u}(t)),t\in[0,T]$, such that all tasks are executed in the prescribed order while minimizing the overall mission duration.

Formally, the objective is to minimize $T$ subject to: 1) the robot dynamics $\dot{\boldsymbol{x}}=f_{\mathrm{MM}}(\boldsymbol{x},\boldsymbol{u})$ with $\boldsymbol{x}(0)=\boldsymbol{x}_{\mathrm{init}}$; 2) sequential execution of tasks without overlap; 3) feasible realization of each task through the prescribed task-relative end-effector motion; and 4) safety and feasibility constraints arising from robot kinematics, actuation limits, obstacle avoidance, and manipulated-object safety. For tasks whose reliable execution requires online estimation of the actual target pose, the trajectory must also provide sufficient object observability before task execution.

\section{System Overview}
% As illustrated in Fig.~\ref{fig:system_overview}, we propose a co-designed planning--execution framework operating at two complementary levels. Our framework starts from (i) a prior environmental point cloud map and (ii) a set of tightly arranged sequential tasks $\boldsymbol{\mathcal{T}}_\mathrm{all}$ with coarse initial poses of the manipulated objects. Rather than processing the entire task sequence at once, the framework operates on a rolling active subset of tasks, as is common in sequential mobile manipulation tasks~\cite{burgess2023architecture,du2023hierarchical}. Specifically, we define the active task set as $\boldsymbol{\mathcal{T}}_{\mathrm{act}}=\{\mathcal{T}_l,\ldots,\mathcal{T}_{l+N_\mathcal{T}-1}\}$, where $N_\mathcal{T}$ is the number of currently active tasks. Once the leading task $\mathcal{T}_l$ is completed, the active task set advances by one task.

\begin{figure}[h]
\centering
\includegraphics[width=0.99\linewidth]{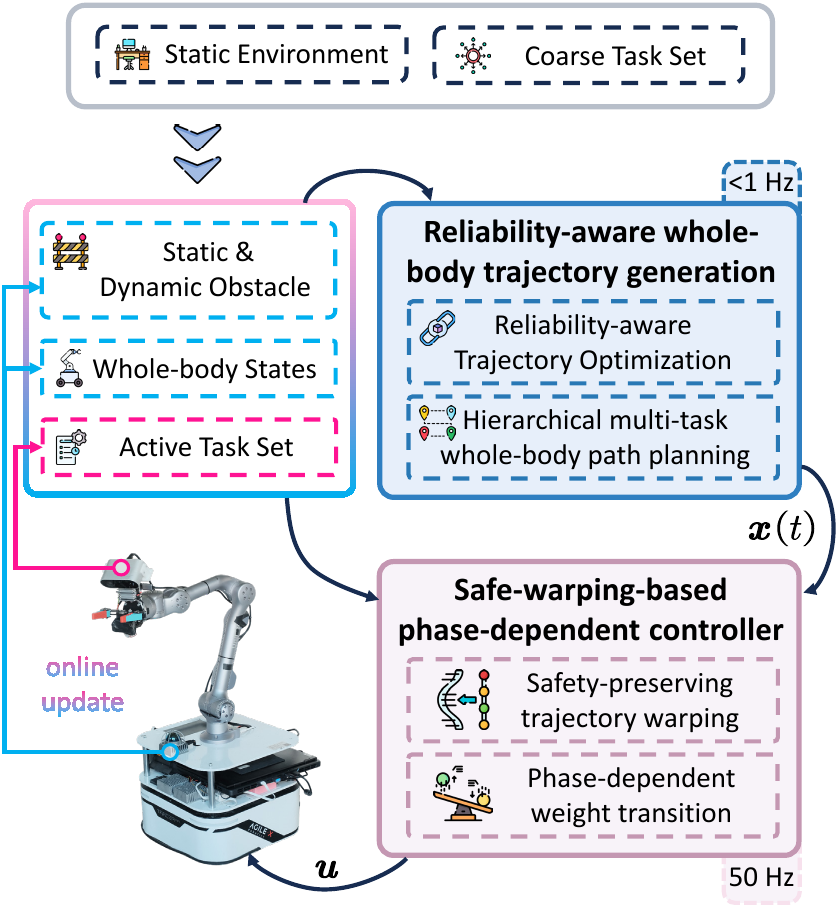}
\vspace{-0.1cm}
\caption{System overview. Starting from a coarse prior environment map and a coarse task set, the planner computes a time-efficient whole-body trajectory that is explicitly shaped to support reliable task execution. During execution, the planner continuously replans using the latest obstacle, robot state, and task information updated online via onboard perception. Based on both the planned trajectory and the latest online information, the controller then smoothly switches between global trajectory tracking and task-error compensation while preserving the safe interaction structure and the overall efficiency of the planned motion. This planning–control loop runs continuously until all tasks are completed.
}
\label{fig:system_overview}
\vspace{-0.1cm}
\end{figure}

As illustrated in Fig.~\ref{fig:system_overview}, we propose a co-designed planning--execution framework for sequential mobile manipulation that operates at two complementary levels. The system takes as input (i) a prior environmental point-cloud map and (ii) a set of tightly arranged sequential tasks $\boldsymbol{\mathcal{T}}_\mathrm{all}$ with coarse initial poses of the manipulated objects. The prior map serves only as a coarse global geometric reference for long-horizon planning. During execution, the robot relies exclusively on onboard perception to continuously update dynamic obstacles, together with the whole-body state and the actual pose estimate of the target object, without requiring external sensing infrastructure. Rather than processing the entire task sequence at once, the framework operates on a rolling active subset of tasks, as is common in sequential mobile manipulation tasks~\cite{burgess2023architecture,du2023hierarchical}. Specifically, we define the active task set as $\boldsymbol{\mathcal{T}}_{\mathrm{act}}=\{\mathcal{T}_l,\ldots,\mathcal{T}_{l+N_\mathcal{T}-1}\}$, where $N_\mathcal{T}$ is the number of currently active tasks. Once the leading task $\mathcal{T}_l$ is completed, the active task set advances by one task.

Based on the coarse references, the planner computes an efficient long-horizon whole-body trajectory $\boldsymbol{x}(t)$ that minimizes inter-task transition time while explicitly shaping the trajectory to promote reliable task completion under real-world uncertainties (Sec.~\nameref{sec:planner}). Especially during the robot's approach to the target vicinity for the task, the whole-body trajectory generates motions that allow the onboard perception sensor to observe the actual object to be manipulated. The perception module then uses these observations to estimate the object's actual pose, and the planner replans at each planning cycle using the latest estimates of the object's pose. In addition, the trajectory facilitates actual task execution without compromising efficiency, including leaving kinematic margin for the manipulator to compensate for task errors under perception, target pose, and control uncertainty, and ensures the safety of the manipulated objects. 

For execution, a model predictive controller (MPC) employs a phase-dependent smooth transition strategy between global trajectory tracking and task-error compensation, while preserving the safe interaction structure and overall efficiency of the planned trajectory (Sec.~\nameref{sec:controller}). In particular, it tracks the efficient motion of the global trajectory and the active perception motion before executing the task. Then, during task-critical phases, it performs task-error compensation to correct errors induced by the lag between replanning updates and the latest object pose estimates, ensuring successful task execution.

\begin{figure*}[t]
\centering
\includegraphics[width=0.95\linewidth]{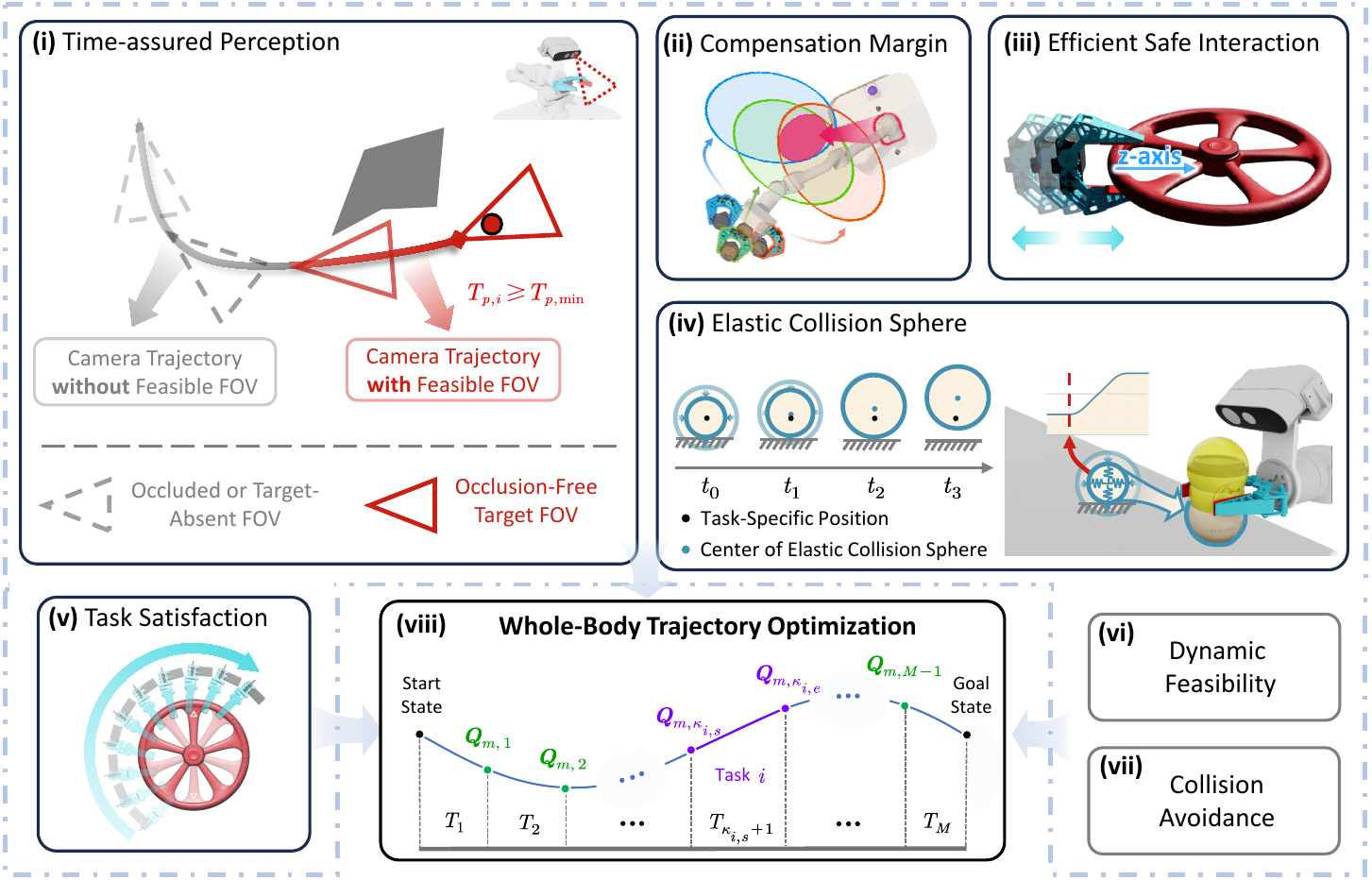}
\vspace{-0.1cm}
\caption{Overview of reliability-aware whole-body trajectory optimization. It generates a time-parameterized whole-body trajectory while enforcing (i) time-assured active perception, (ii) compensation margin zone constraints, (iii) efficient safe interaction constraints, (iv) elastic collision sphere safety constraints, (v) task satisfaction constraints, (vi) dynamic feasibility, and (vii) collision avoidance, yielding (viii) the optimized whole-body trajectory. 
}
\label{fig:trajectory_optimization}
\vspace{-0.4cm}
\end{figure*}

\section{Reliability-aware Whole-body Trajectory Generation}
\label{sec:planner}

At the planning level, we first formulate the whole-body trajectory generation problem as a spatial-temporal optimization that minimizes overall execution time while explicitly promoting reliable task execution (Sec.~\nameref{subsec:trajectory_opt}). 
We then warm-start this optimization with a whole-body path generated by a hierarchical multi-task planner  (Sec.~\nameref{subsec:path_planning}), which provides 
(i) a collision-free discrete whole-body path $\mathcal{P}=\{\boldsymbol{x}_0,\boldsymbol{x}_1,\dots,\boldsymbol{x}_M\}$,
(ii) one feasible grasp transform ${}^{\mathrm{o}_i}\boldsymbol{P}_{\mathcal{C},i}\in\boldsymbol{\mathcal{C}}_i$ per task, and 
(iii) task-phase indices $\mathcal{K}=\{(\kappa_{i,\mathrm{s}},\kappa_{i,\mathrm{e}})\}_{i=1}^{N_{\mathcal{T}}}$, where
$\kappa_{i,\mathrm{s}}$ and $\kappa_{i,\mathrm{e}}$ are the start and end waypoint indices of task $i$ on $\mathcal{P}$.

\subsection{Reliability-aware Trajectory Optimization Formulation}
\label{subsec:trajectory_opt}
% \subsubsection{Robot Model}
For the experimental studies, we use a representative mobile manipulator configuration \cite{burgess2024reactive} in this work, specifically a two-wheel differential-drive base paired with an $\mathcal{L}$-DOF manipulator with a two-finger gripper attached to the last link of the manipulator (Fig.~\ref{fig:system_overview}). The state of the mobile manipulator is denoted as $\boldsymbol{x}=[\boldsymbol{x}_\mathfrak{b}^\top,\boldsymbol{q}_\mathfrak{m}^\top]^\top$, where $\boldsymbol{x}_\mathfrak{b}=[\boldsymbol{q}_{\mathfrak{b}},\psi]^\top\in\mathrm{SE}(2)$ is the pose of the mobile base and $\boldsymbol{q}_\mathfrak{m}\in\mathbb{R}^{\mathcal{L}}$ is the joint angles of the manipulator. $\boldsymbol{q}_{\mathfrak{b}}=[q_x,q_y]^\top \in \mathbb{R}^2$ and $\psi$ represents the mobile base planar position and orientation, respectively.

% Denote $\omega_l$ and $\omega_r$ as the angular velocities of the left and right wheels of the mobile base, respectively. To empower efficient trajectory planning, we define $\boldsymbol{q} = [\boldsymbol{q}_{\mathfrak{b}}^\top, \boldsymbol{q}_{\mathfrak{m}}^\top]^\top\in\mathbb{R}^{2+\mathcal{L}}$ as generalized joint configuration space. $\psi =\mathrm{atan2}\left( \dot{q}_y, \dot{q}_x \right)$

\subsubsection{Trajectory Representation}
\label{subsubsec:traj_rep}
We plan the whole-body MM trajectory in generalized joint configuration space $\boldsymbol{q} = [\boldsymbol{q}_{\mathfrak{b}}^\top, \boldsymbol{q}_{\mathfrak{m}}^\top]^\top\in\mathbb{R}^{2+\mathcal{L}}$. 
% The state of the mobile manipulator, including the orientation of the base, as well as the velocities/accelerations of the wheel and the joint, can be analytically derived from $\boldsymbol{q}$ and its time derivatives \cite{wu2024real}. 
The whole-body trajectory $\boldsymbol{x}(t)=[q_x(t),q_y(t),\psi(t),\boldsymbol{q}_\mathfrak{m}^\top(t)]^\top$ can be easily obtained from $\boldsymbol{q}(t)$, with $\psi(t)=\mathrm{atan2}(\dot{q}_y(t),\dot{q}_x(t))$ being the orientation of the differential-driven mobile base.

We represent trajectory $\boldsymbol{q}(t)$ as a $M$-segment piecewise polynomial of degree $2s-1$:
\begin{equation}
\boldsymbol{q}(t) = 
\begin{cases}
    \boldsymbol{q}_{1}(t) = \boldsymbol{C}_{1}^{\top}\boldsymbol{\beta}(t - \bar{t}_{0}), & t \in [\bar{t}_{0}, \bar{t}_{1}] \\
    \quad \vdots & \quad \vdots \\
    \boldsymbol{q}_{M}(t) = \boldsymbol{C}_{M}^{\top}\boldsymbol{\beta}(t - \bar{t}_{M-1}), & t \in [\bar{t}_{M-1}, \bar{t}_{M}],
\end{cases}
\end{equation}
where $\boldsymbol{\beta}(t) = [1, t, \dots, t^{2s-1}]^\top$ is the time basis vector, $\boldsymbol{T} = [T_1, \dots, T_M]^\top$ is the vector of segment durations, $\bar{t}_i = \sum_{j=1}^i T_j$ is the cumulative time at the end of segment $i$, and $\boldsymbol{C} = [\boldsymbol{C}_1^\top, \dots, \boldsymbol{C}_M^\top]^\top$ is polynomial coefficient matrix, which can be obtained in linear complexity by a map $\boldsymbol{C}=\boldsymbol{C}(\boldsymbol{Q}_{m},\boldsymbol{T},\boldsymbol{q}_\text{0},\boldsymbol{q}_\text{f})$ \cite{WANG2022GCOPTER} that minimizes control effort of the trajectory. $\boldsymbol{q}_\text{0}$ and $\boldsymbol{q}_\text{f}$ are the start and final state of the trajectory. $\boldsymbol{Q}_{m}$ denotes the intermediate waypoints. We additionally introduce the perception time allocation $\boldsymbol{T}_{\mathrm{p}}=\{T_{\mathrm{p},i}\}_{i\in\mathcal{I}_{\text{perc}}}$, where $T_{\mathrm{p},i}$ represents the optimizable duration of the active perception phase preceding the start of task $i$, and $\mathcal{I}_{\text{perc}}$ denotes the index set of tasks that necessitate task pose estimation (e.g., grasping tasks). Treating $T_{\mathrm{p},i}$ as a decision variable enables the solver to adaptively optimize observation windows, leveraging available temporal slack to acquire more visual data for robust pose estimation without compromising overall efficiency.

\subsubsection{Trajectory Optimization}
We then formulate the whole-body trajectory generation problem as a spatial-temporal trajectory optimization (Fig.~\ref{fig:trajectory_optimization}). 
To make the non-convex problem tractable, we assume that the task-phase indices $\mathcal{K}$ and the discrete grasp transforms $\{\,^{\mathrm{o}_i}\boldsymbol{P}_{\mathcal{C},i}\}$ are provided by the front-end planner (detailed in the Sec.~\nameref{subsec:path_planning}) and remain fixed during optimization. We seek to find the optimal intermediate waypoint $\boldsymbol{Q}_{m}$, time allocation $\boldsymbol{T}$, final state $\boldsymbol{q}_\text{f}$ and perception time allocation $\boldsymbol{T}_{\mathrm{p}}$ that minimize a cost function balancing control effort, operation time, and perception duration:
\begin{equation}
\begin{aligned}
    \min_{\boldsymbol{Q}_m, \boldsymbol{T}, \boldsymbol{q}_\text{f},\boldsymbol{T}_{\mathrm{p}}} \quad &\int_{0}^{\bar{t}_M} \left\| \overset{(s)}{\boldsymbol{q}}(t) \right\|_2^2 dt + \omega_T \bar{t}_M-\omega_{T_{\mathrm{p}}}\|\boldsymbol{T}_{\mathrm{p}}\|_1 \\
    \text{s.t.}\quad &T_i > 0, \quad \forall i \in \{1, \dots, M\} \\
    &\mathcal{G}_g(\boldsymbol{q}^{[s]}(t),\boldsymbol{T},\boldsymbol{T}_{\mathrm{p}}) \le 0, \ \forall t \in \mathcal{S}_g, \forall g \in \mathcal{D}_\mathcal{G} \\
    &\mathcal{H}_h(\boldsymbol{q}^{[s]}(t),\boldsymbol{T},\boldsymbol{T}_{\mathrm{p}}) = 0, \ \forall t \in \mathcal{S}_h, \forall h \in \mathcal{D}_\mathcal{H}
\end{aligned}
\label{eq:opt_prob}
\end{equation}
This minimization is subject to:
\begin{itemize}
    \item \textbf{Time Allocation}: Segment durations must be positive: $T_i > 0, \forall i \in \{1, \dots, M\}$.
    \item \textbf{General Constraints}: The trajectory must adhere to various inequality constraints $\mathcal{G}_g$ and equality constraints $\mathcal{H}_h$ over specified time intervals $\mathcal{S}$ and constraint sets $\mathcal{D}_\mathcal{G}$ and $\mathcal{D}_\mathcal{H}$, which will be further discussed in the following sections.
\end{itemize}
In addition to the reliability-aware constraints described below, the optimizer enforces standard whole-body feasibility constraints, including task-satisfaction, dynamic feasibility, and safety constraints. Explicit formulations are provided in \nameref{app:constraints}.

\subsubsection{Time-assured Active Perception}
To address the insufficiency of observation windows for reliable task pose estimation (arising from efficiency-observability trade-offs in high-efficiency trajectories), we propose a simple yet effective Time-assured Active Perception (TAP) constraint (Fig.~\ref{fig:trajectory_optimization}(i)) to guarantee adequate observation time $T_{\mathrm{p},i}$ before manipulation:
\begin{align}
    T_{\mathrm{p},\min} \leq T_{\mathrm{p},i} \leq \sum_{j=\kappa_{i-1,\mathrm{e}}}^{\kappa_{i,\mathrm{s}}-1} T_j, \quad \forall i\in \mathcal{I}_{\text{perc}}.
\end{align}
Here, we define $\kappa_{0,\mathrm{e}} := 1$. $T_{\mathrm{p},\min}$ is the minimum observation time required to accommodate the latency of perception pipelines; and $\sum_{j=\kappa_{i-1,\mathrm{e}}}^{\kappa_{i,\mathrm{s}}-1} T_j$ corresponds to the time interval between the end of the $(i-1)$-th task and the start of the $i$-th task, i.e., the available time span for perception. 

Subsequently, visibility constraints are enforced during the perception interval preceding the start of task $i$ to ensure the onboard perception sensor can observe the target throughout $T_{\mathrm{p},i}$:
\begin{equation}
\begin{aligned}
    &\mathcal{G}_{g,i}(\boldsymbol{q}^{[s]}(t),\boldsymbol{T},\boldsymbol{T}_{\mathrm{p}}) \leq 0, \forall g\in\mathcal{D}_{\mathcal{G}_\mathfrak{v}}, \\ 
    &\ \forall t \in [\bar{t}_{\kappa_{i,\mathrm{s}}}-T_{\mathrm{p},i},\bar{t}_{\kappa_{i,\mathrm{s}}}), \ \forall i\in\mathcal{I}_{\text{perc}},
\end{aligned}
\end{equation}
where $\mathcal{D}_{\mathcal{G}_\mathfrak{v}} = \{\mathfrak{v}_a, \mathfrak{v}_d, \mathfrak{v}_o\}$ defines the set of visibility constraint indices, corresponding to the field-of-view constraint ($\mathfrak{v}_a$), the maximum sensing range constraint ($\mathfrak{v}_d$), and the occlusion-free constraint ($\mathfrak{v}_o$), respectively. Detailed formulations of these three constraints are provided in \nameref{app:visibility}.

By explicitly coupling a time guarantee with visibility constraints, TAP mitigates the efficiency-observability trade-off, preventing the optimizer from neglecting observation requirements in pursuit of efficiency while ensuring our system has sufficient tolerance for the perception pipeline’s latency to generate a reliable pose estimate, thus eliminating task failures caused by stale information.

\subsubsection{Compensation Margin Zone}
To preserve sufficient kinematic margin for online task-error compensation under time-efficient motions, we introduce the Compensation Margin Zone (CMZ) constraint (Fig.~\ref{fig:trajectory_optimization}(ii)). Instead of solely ensuring the nominal task end-effector pose is reachable, the CMZ explicitly restricts the manipulator base to zones from which a neighborhood of the nominal pose also remains reachable. In this way, the robot can still compensate for bounded task-pose errors during task-critical phases, rather than operating at configurations where even small target deviations would cause compensation failure.

Inspired by inverse reachability-based placement methods~\cite{xu2020planning,xu2021planning}, we formulate the CMZ as a continuous-time trajectory constraint to promote compensation capability during execution. Let $\mathcal{I}_{\mathrm{cmz}}$ denote the index set of tasks that require task-error compensation (e.g., grasping tasks), and let $\mathcal{S}_{\mathrm{cmz},i}$ represent the corresponding set of local timestamps during task $i$ where this compensation capability must be maintained (e.g., grasping instant).

For each task $i\in\mathcal{I}_{\mathrm{cmz}}$ and timestamp $\tau\in\mathcal{S}_{\mathrm{cmz},i}$, we construct a local neighborhood of the nominal task end-effector pose $\bar{\boldsymbol{P}}_{\mathrm{t},i}(\tau)$ by sampling a set of nearby poses
\begin{equation}
    \mathcal{P}_{\mathrm{cmz},i}(\tau)\subset \mathrm{SE}(3).
\end{equation}
These samples represent bounded task pose perturbations caused by real-world uncertainty in perception, control, and target pose. They provide a tractable discrete approximation of the local deviations that the online controller is expected to compensate for. In practice, $\mathcal{P}_{\mathrm{cmz},i}(\tau)$ is constructed by sampling both position and orientation perturbations around $\bar{\boldsymbol{P}}_{\mathrm{t},i}(\tau)$. We first sample positions on a sphere of radius $r_{\mathrm{cmz}}$ centered at the nominal end-effector position. For each sampled position, we then sample rotational perturbations around the nominal approach direction after a fixed tilt offset. Any sampled pose that is in collision is discarded. The resulting set provides a discrete approximation of the target-pose deviations that the controller is expected to compensate online. Details of the sampling strategy are provided in \nameref{app:cmz}.

For each sampled pose $\boldsymbol{P}\in\mathcal{P}_{\mathrm{cmz},i}(\tau)$, we use the inverse reachability map~\cite{vahrenkamp2013robot} to compute the set of planar manipulator-base positions from which $\boldsymbol{P}$ is reachable, denoted by $\mathcal{R}(\boldsymbol{P})\subset\mathbb{R}^2$. We then intersect these reachable sets across all sampled poses:
\begin{equation}
\mathcal{R}_{\mathrm{cmz},i}(\tau)
=
\bigcap_{\boldsymbol{P}\in\mathcal{P}_{\mathrm{cmz},i}(\tau)}
\mathcal{R}(\boldsymbol{P}).
\label{eq:cmz_intersection}
\end{equation}
The resulting intersection $\mathcal{R}_{\mathrm{cmz},i}(\tau)$ contains the manipulator-base positions from which all the sampled poses in the neighborhood remain reachable. Therefore, it defines the set of base positions from which bounded task-pose errors can still be compensated kinematically. If $\mathcal{R}_{\mathrm{cmz},i}(\tau)$ is empty, this implies that the requested compensation margin $r_{\mathrm{cmz}}$ is too large under current workspace constraints. In such cases, we iteratively reduce $r_{\mathrm{cmz}}$ and repeat the sampling and intersection procedure until a nonempty feasible compensation zone is obtained. Thus, the CMZ acts as a feasibility-aware robustness constraint, maximizing the allowable local compensation margin within the physical workspace limits.

To facilitate efficient trajectory optimization, we approximate $\mathcal{R}_{\mathrm{cmz},i}(\tau)$ with a 2D ellipse using Principal Component Analysis (PCA)~\cite{abdi2010principal}, denoted as $\mathcal{E}_{\mathrm{c},i}(\tau)\subset\mathbb{R}^2$:
\begin{equation}
\begin{aligned}
\mathcal{E}_{\mathrm{c},i}(\tau)
=
\Bigl\{
&
\boldsymbol{o}_{\mathrm{c},i}(\tau)
+
\boldsymbol{R}_{\mathrm{c},i}(\tau)
\boldsymbol{Q}_{\mathrm{c},i}(\tau)
\boldsymbol{v}
\\
&\;\Big|\;
\boldsymbol{v}\in\mathbb{R}^2,\;
\|\boldsymbol{v}\|_2 \le 1
\Bigr\}.
\end{aligned}
\end{equation}
where $\boldsymbol{o}_{\mathrm{c},i}(\tau) \in \mathbb{R}^2$, $\boldsymbol{R}_{\mathrm{c},i}(\tau) \in \mathrm{SO}(2)$ and $\boldsymbol{Q}_{\mathrm{c},i}(\tau) = \mathrm{diag}(a_i(\tau), b_i(\tau))$ denote the center, orientation, and semi-axes matrix of $\mathcal{E}_{\mathrm{c},i}(\tau)$, respectively.
Consequently, we constrain the manipulator base $\boldsymbol{p}_{\mathfrak{m},\mathrm{b},i}(\tau):=\big[f_{\mathrm{FK,b}}(\boldsymbol{x}(\bar{t}_{\kappa_{i,\mathrm{s}}}+\tau))\big]_{\mathrm{xy}}$ to reside within $\mathcal{E}_{\mathrm{c},i}(\tau)$ at the compensation timestamps:
\begin{equation}
\begin{aligned}
\|\boldsymbol{Q}_{\mathrm{c},i}^{-1}(\tau)\boldsymbol{R}_{\mathrm{c},i}^{\top}(\tau)(\boldsymbol{p}_{\mathfrak{m},\mathrm{b},i}(\tau)-\boldsymbol{o}_{\mathrm{c},i}(\tau))\|_2
\leq
1,
\\
\forall\tau\in\mathcal{S}_{\mathrm{cmz},i},
\forall i\in\mathcal{I}_{\mathrm{cmz}}.
\end{aligned}
\label{eq:cmz_constraint}
\end{equation}
% \begin{equation}
% \begin{aligned}
% \big[f_{\mathrm{FK,b}}(\boldsymbol{x}(\bar{t}_{\kappa_{i,\mathrm{s}}}+\tau))\big]_{\mathrm{xy}}
% \in
% \mathcal{E}_{\mathrm{cmz},i}(\tau),
% \\
% \tau\in\mathcal{S}_{\mathrm{cmz},i},
% i\in\mathcal{I}_{\mathrm{cmz}}.
% \end{aligned}
% \label{eq:cmz_constraint}
% \end{equation}
Here, $[\cdot]_{\mathrm{xy}}:\mathrm{SE}(3)\rightarrow\mathbb{R}^2$ extracts the planar position, and $f_{\mathrm{FK,b}}(\boldsymbol{x})\in\mathrm{SE}(3)$ denotes the pose of the manipulator base at state $\boldsymbol{x}$.

By explicitly enforcing Eq.~\eqref{eq:cmz_constraint}, the CMZ prevents the optimizer from forcing the manipulator into kinematic-limit configurations in pursuit of time-efficient motions, while ensuring that the online controller retains sufficient kinematic redundancy to compensate for end-effector errors arising during execution under real-world uncertainty, thus improving reliability.

\subsubsection{Efficient Safe Interaction Motion}
To avoid failure modes caused by gripper--object collisions during the pre-grasp approach and post-placement retraction (e.g., grasp failure caused by pushing the object away during approach), we introduce the Efficient Safe Interaction (ESI) motion constraints (Fig.~\ref{fig:trajectory_optimization}(iii)). The key idea is that the grasp/place pose of the end-effector is chosen to be collision-free by construction; thus, we restrict approach/retraction to a 1D manifold by (i) keeping the end-effector orientation close to the grasp/place orientation and (ii) translating along the approach ray defined at that pose (Fig.~\ref{fig:trajectory_optimization}(iii)). Crucially, ESI is imposed over a time window whose length scales with $T_{\kappa_{i,\mathrm{s}}}$ (pre-grasp) and $T_{\kappa_{i,\mathrm{e}}+1}$ (post-placement), which are optimized jointly with the trajectory. Therefore, the solver can adjust the timing of approach/retraction without prescribing a fixed duration. 

Let $\mathcal{I}_{\text{grasp}}$ denote the index set of tasks that start with grasping an object and $\mathcal{I}_{\text{place}}$ denote the index set of tasks that end with placing an object. Define $\mathcal{S}_{\mathrm{pre},i}:=\bigl(\bar{t}_{\kappa_{i,\mathrm{s}}}-\alpha_m T_{\kappa_{i,\mathrm{s}}},\,\bar{t}_{\kappa_{i,\mathrm{s}}}\bigr)$ and $\mathcal{S}_{\mathrm{post},i}:=(\bar{t}_{\kappa_{i,\mathrm{e}}},\,\bar{t}_{\kappa_{i,\mathrm{e}}}+\alpha_m T_{\kappa_{i,\mathrm{e}}+1})$ as the pre-grasp and post-place time window, respectively, where $\alpha_m\in(0,1)$ determines the relative window length. The optimization of $T_{\kappa_{i,\mathrm{s}}}$ and $T_{\kappa_{i,\mathrm{e}}+1}$ will adjust the time window $\mathcal{S}_{\mathrm{pre},i}$ and $\mathcal{S}_{\mathrm{post},i}$ thus determining the best duration for the safe pre-grasp and post-place motion.

\begin{figure}[t]
\centering
\includegraphics[width=0.99\linewidth]{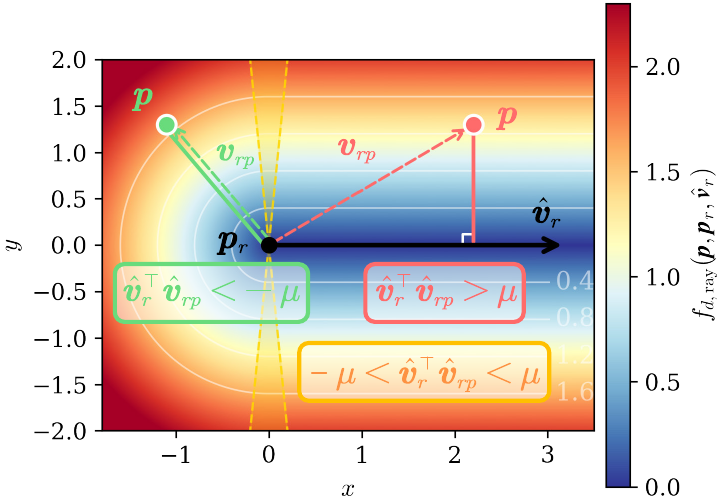}
\vspace{-0.1cm}
\caption{Illustration of $f_{d,\mathrm{ray}}$: smooth transition from perpendicular distance to Euclidean distance.
}
\label{fig:fdray}
\vspace{-0.1cm}
\end{figure}

Then we define ESI motion constraints. Denote $[\cdot]_{\boldsymbol{p}}$, $[\cdot]_{\boldsymbol{z}}$, and $[\cdot]_{\boldsymbol{R}}$ as the operators that extract the position, the $z$-axis of the rotation matrix, and the rotation matrix from a homogeneous transform, respectively. To simplify notation, let $\bar{\boldsymbol{P}}_{\mathrm{t,b},i}:=\bar{\boldsymbol{P}}_{\mathrm{t},i}(0)$, $\bar{\boldsymbol{P}}_{\mathrm{t,e},i}:=\bar{\boldsymbol{P}}_{\mathrm{t},i}(T_{\mathcal{T},i})$, and $\boldsymbol{p}_{\mathrm{ee}}(t):=\bigl[f_{\mathrm{FK},\mathrm{ee}}(\boldsymbol{x}(t))\bigr]_{\boldsymbol{p}}$. For each $i\in\mathcal{I}_{\text{grasp}}$ and   $t\in\mathcal{S}_{\mathrm{pre},i}$, we enforce
\begin{equation}
    f_{d,\mathrm{ray}}\!\left(
  \boldsymbol{p}_{\mathrm{ee}}(t),\,
  [\bar{\boldsymbol{P}}_{\mathrm{t,b},i}]_{\boldsymbol{p}},\,
  -[\bar{\boldsymbol{P}}_{\mathrm{t,b},i}]_{\boldsymbol{z}}
\right) \le d_{\mathrm{pos}},
\end{equation}
\begin{equation}
    f_{d,\boldsymbol{R}}\!\left(
  \bigl[f_{\mathrm{FK},\mathrm{ee}}(\boldsymbol{x}(t))\bigr]_{\boldsymbol{R}},\,
  [\bar{\boldsymbol{P}}_{\mathrm{t,b},i}]_{\boldsymbol{R}}
\right) \le d_o,
\end{equation}
% \begin{subequations}
% \begin{align}
% f_{d,\mathrm{ray}}\!\left(
%   \boldsymbol{p}_{\mathrm{ee}}(t),\,
%   [\bar{\boldsymbol{P}}_{\mathrm{t,b},i}]_{\boldsymbol{p}},\,
%   -[\bar{\boldsymbol{P}}_{\mathrm{t,b},i}]_{\boldsymbol{z}}
% \right) &\le d_{\mathrm{pos}},\\
% f_{d,\boldsymbol{R}}\!\left(
%   \bigl[f_{\mathrm{FK},\mathrm{ee}}(\boldsymbol{x}(t))\bigr]_{\boldsymbol{R}},\,
%   [\bar{\boldsymbol{P}}_{\mathrm{t,b},i}]_{\boldsymbol{R}}
% \right) &\le d_o,
% \end{align}
% \end{subequations}
where $d_{\mathrm{pos}}$ and $d_o$ are position and orientation tolerances. $f_{d,\mathrm{ray}}(\boldsymbol{p},\boldsymbol{p}_r,\hat{\boldsymbol{v}}_r)$ is the smooth distance from point $\boldsymbol{p}$ to the ray that originates at $\boldsymbol{p}_r$ and points along $\hat{\boldsymbol{v}}_r$. To ensure differentiability when the point projects behind the ray origin, we blend the perpendicular distance with the Euclidean distance to the origin (Fig.~\ref{fig:fdray}) using a smooth step function:
\begin{equation}
\label{eq:supp_ray_dist}
\begin{aligned}
&f_{d,\mathrm{ray}}\left( \boldsymbol{p},\boldsymbol{p}_{r},\hat{\boldsymbol{v}}_{r} \right)
\\&=
\Bigl( 1-f_{\log}\bigl( \hat{\boldsymbol{v}}_{r}^{\top}\hat{\boldsymbol{v}}_{rp},-\mu,\mu \bigr) \Bigr)\left\| \boldsymbol{p}-\boldsymbol{p}_{r} \right\| _2\\
&\ +f_{\log}\bigl( \hat{\boldsymbol{v}}_{r}^{\top}\hat{\boldsymbol{v}}_{rp},-\mu,\mu \bigr)
\left\| \bigl( \boldsymbol{I}_3-\hat{\boldsymbol{v}}_{r}\hat{\boldsymbol{v}}_{r}^{\top} \bigr) \left( \boldsymbol{p}-\boldsymbol{p}_{r} \right) \right\| _2,
\end{aligned}
\end{equation}
where $\hat{\boldsymbol{v}}_{rp}=(\boldsymbol{p}-\boldsymbol{p}_{r})/\left\| \boldsymbol{p}-\boldsymbol{p}_{r} \right\| _2$ is the unit vector pointing to $\boldsymbol{p}$, and $\mu$ is a smoothing parameter. The blending function $f_{\log}(x,a,b)$ is a $C^2$-continuous scalar function defined as:
\begin{align}
f_{\log}(x,a,b)=\begin{cases}
	0,&		\bar{x}\le -\mu,\\
	\frac{(\bar{x}+\mu )^3(\mu -\bar{x})}{2\mu ^4},&		-\mu <\bar{x}\le 0,\\
	\frac{(\bar{x}+\mu )(\bar{x}-\mu )^3}{2\mu ^4}+1,&		0<\bar{x}\le \mu,\\
	1,&		\bar{x}>\mu,
\end{cases}
\label{eq:supp_smooth_log}
\end{align}
where $\bar{x}=x-0.5(a+b)$ and $\mu=0.5(b-a)$.
\begin{equation}
f_{d,\boldsymbol{R}}(\boldsymbol{R}_1,\boldsymbol{R}_2) = \mathrm{arccos} \left( \frac{1}{2}\,\mathrm{tr}\left( \boldsymbol{R}_1^\top \boldsymbol{R}_2 \right) - \frac{1}{2} \right)
\end{equation}
is the orientation error between $\boldsymbol{R}_1$ and $\boldsymbol{R}_2$. We define the local $z$-axis of the end-effector to point along the direction of the gripper fingers (Fig.~\ref{fig:trajectory_optimization}(iii)). 

To avoid starting the ESI window excessively close to the object, we additionally constrain the gripper position at the window start to lie on the same ray, at a prescribed offset distance $d_m$ from the grasp pose:
\begin{equation}
\label{eq:esi_grasp_anchor}
\begin{aligned}
&\left\|
 \boldsymbol{p}_{\mathrm{ee}}(t)
 - \Bigl([\bar{\boldsymbol{P}}_{\mathrm{t,b},i}]_{\boldsymbol{p}}
 - d_m[\bar{\boldsymbol{P}}_{\mathrm{t,b},i}]_{\boldsymbol{z}}\Bigr)
\right\|_2 \le d_{\mathrm{pos}},\\
&\quad t = \bar{t}_{\kappa_{i,\mathrm{s}}}-\alpha_m T_{\kappa_{i,\mathrm{s}}},\quad \forall\, i\in \mathcal{I}_{\text{grasp}}.
\end{aligned}
\end{equation}

Post-place ESI constraints are defined analogously for $i\in\mathcal{I}_{\text{place}}$, using the final task end-effector pose $\bar{\boldsymbol{P}}_{\mathrm{t,e},i}$: 
\begin{equation}
\begin{aligned}
    f_{d,\mathrm{ray}}\!\left(
  \boldsymbol{p}_{\mathrm{ee}}(t),\,
  [\bar{\boldsymbol{P}}_{\mathrm{t,e},i}]_{\boldsymbol{p}},\,
  -[\bar{\boldsymbol{P}}_{\mathrm{t,e},i}]_{\boldsymbol{z}}
\right) \le d_{\mathrm{pos}},\\t\in\mathcal{S}_{\mathrm{post},i},
\end{aligned}
\end{equation}
\begin{equation}
    f_{d,\boldsymbol{R}}\!\left(
  \bigl[f_{\mathrm{FK},\mathrm{ee}}(\boldsymbol{x}(t))\bigr]_{\boldsymbol{R}},\,
  [\bar{\boldsymbol{P}}_{\mathrm{t,e},i}]_{\boldsymbol{R}}
\right) \le d_o,t\in\mathcal{S}_{\mathrm{post},i},
\end{equation}
% \begin{subequations}
% \label{eq:esi_place_window}
% \begin{align}
% f_{d,\mathrm{ray}}\!\left(
%   \boldsymbol{p}_{\mathrm{ee}}(t),\,
%   [\bar{\boldsymbol{P}}_{\mathrm{t,e},i}]_{\boldsymbol{p}},\,
%   -[\bar{\boldsymbol{P}}_{\mathrm{t,e},i}]_{\boldsymbol{z}}
% \right) &\le d_{\mathrm{pos}},t\in\mathcal{S}_{\mathrm{post}}\\
% f_{d,\boldsymbol{R}}\!\left(
%   \bigl[f_{\mathrm{FK},\mathrm{ee}}(\boldsymbol{x}(t))\bigr]_{\boldsymbol{R}},\,
%   [\bar{\boldsymbol{P}}_{\mathrm{t,e},i}]_{\boldsymbol{R}}
% \right) &\le d_o.
% \end{align}
% \end{subequations}
\begin{equation}
\label{eq:esi_place_anchor}
\begin{aligned}
\left\|
 \boldsymbol{p}_{\mathrm{ee}}(t)
 - \Bigl([\bar{\boldsymbol{P}}_{\mathrm{t,e},i}]_{\boldsymbol{p}}
 - d_m[\bar{\boldsymbol{P}}_{\mathrm{t,e},i}]_{\boldsymbol{z}}\Bigr)
\right\|_2 \le d_{\mathrm{pos}},\\
\quad t = \bar{t}_{\kappa_{i,\mathrm{e}}}+\alpha_m T_{\kappa_{i,\mathrm{e}}+1}.
\end{aligned}
\end{equation}

Overall, ESI provides an explicit safety envelope for approach and retraction, while the optimizability of $T_{\kappa_{i,\mathrm{s}}}$ and $T_{\kappa_{i,\mathrm{e}}+1}$ preserves temporal flexibility and thus supports time-efficient planning.

\subsubsection{Elastic Collision Spheres for Intended-Contact Manipulation}

Manipulation tasks that involve intended contact with the environment, e.g., placing an object onto a support surface, require the held object to be allowed to approach the environment at task poses, while maintaining a strict safety margin during transport to avoid slipping from the end-effector. A fixed collision model is therefore prone to being overly conservative at intended-contact poses or unsafe in non-contact phases. We address this with Elastic Collision Spheres (ECS), which modulate sphere radius as a smooth function of the sphere's displacement from its task-specific target.

We model the environment using a Euclidean Signed Distance Field (ESDF) map \cite{zhou2019robust} $D_{\text{ESDF}}:\mathbb{R}^3\!\rightarrow\!\mathbb{R}$, where a collision sphere centered at $\boldsymbol{p}$ with radius $r$ is nominally collision-free if $D_{\text{ESDF}}(\boldsymbol{p}) \ge r$. However, in practice, a positive safety margin $d_\mathrm{s}$ is incorporated to provide a robust buffer against tracking errors during motion and numerical optimization tolerances, resulting in the robust safety condition 
\begin{equation}
\label{eq:esdf_sphere}
    D_{\text{ESDF}}(\boldsymbol{p}) \ge r+d_\mathrm{s}.
\end{equation}
For a held object $\mathrm{o}$, to simplify collision checking while maintaining safety, its geometry is approximated by elastic spheres indexed by $m\in\{1,\dots,m_o\}$ with centers ${}^{\mathrm{o}}\boldsymbol{p}_{\mathrm{o},m}$ in the object frame and conservative radius $r_{\mathrm{o},m}$. Let $\boldsymbol{p}_{\mathrm{t},\mathrm{o},m}$ denote the task-specific target location in the world frame of sphere $m$ at the relevant task pose. For example, if the task requires to place the object $\mathrm{o}$ at pose $\boldsymbol{P}_\mathrm{t}$, the task-specific target location of sphere $m$ is $\boldsymbol{p}_{\mathrm{t},\mathrm{o},m}=\boldsymbol{P}_\mathrm{t}\,{}^{\mathrm{o}}\boldsymbol{p}_{\mathrm{o},m}$.

\paragraph{Elastic radius.}
Here, we simplify the notion for clearance by omitting the object and sphere indices $(\mathrm{o},m)$. 
The elastic radius is designed to dynamically adjust with the sphere center $\boldsymbol{p}$ relative to the task-specific target position $\boldsymbol{p}_{\mathrm{t}}$ (Fig.~\ref{fig:trajectory_optimization}(iv)). 
At the intended contact pose $\boldsymbol{p}_{\mathrm{t}}$, the signed distance to the environment $D_{\text{ESDF}}(\boldsymbol{p}_{\mathrm{t}})$ is inherently small. Consequently, the standard safety condition Eq.~\eqref{eq:esdf_sphere} is often impossible to satisfy with the initial conservative radius $r$. However, we assume that the target pose is physically feasible and safe for the object. This apparent violation of the safety constraint typically arises from the use of an overly conservative bounding radius $r$ and the inherent discretization in the ESDF map, rather than from actual physical penetration.
To make the constraint satisfied, the maximum permissible radius becomes $r_{\mathrm{min}}=D_{\text{ESDF}}(\boldsymbol{p}_{\mathrm{t}}) - d_\mathrm{s}$. We therefore define the required shrinkage for initial conservative radius $r$ at $\boldsymbol{p}_{t}$ as 
\begin{equation}
    f_{d_v}(\boldsymbol{p}_{t},r)=\max\!\left(0,\, r+d_\mathrm{s}-D_{\text{ESDF}}(\boldsymbol{p}_{t})\right),
\end{equation}
which represents how much the initial radius $r$ exceeds the safety limit at $\boldsymbol{p}_{\mathrm{t}}$. 

Based on the required shrinkage $f_{d_v}(\boldsymbol{p}_{t},r)$, we design an elastic radius that equals $r_{\mathrm{min}}$ at $\boldsymbol{p}_{t}$ to satisfy Eq.~\eqref{eq:esdf_sphere}, and smoothly recovers toward the conservative radius $r$ as the sphere moves away from $\boldsymbol{p}_{t}$. This recovery intentionally biases departures motion toward increasing clearance to prevent unintended contact between the object and the support surface (Fig.~\ref{fig:trajectory_optimization}(iv)).
Therefore, we define the elastic radius for a sphere centered at $\boldsymbol{p}$ as
\begin{equation}
\begin{aligned}
r_{\mathrm{elastic}}(\boldsymbol{p}_{t},\boldsymbol{p},r)
&= r - f_{d_v}(\boldsymbol{p}_{t},r)\\
&+ f_s\!\left(\|\boldsymbol{p}_{t}-\boldsymbol{p}\|_2,\ f_{d_v}(\boldsymbol{p}_{t},r)\right),
\end{aligned}
\label{eq:elastic_radius}
\end{equation}
where $f_s(x,d_v)$ serves as a smooth recovery function (with $x=\|\boldsymbol{p}_{t}-\boldsymbol{p}\|_2$ and $d_v=f_{d_v}(\boldsymbol{p}_{t},r)$). It is constructed to increase monotonically from $0$ (at $x=0$) to $d_v$, allowing the elastic radius to gradually recover back to the conservative radius $r$ as the sphere moves away from $\boldsymbol{p}_{t}$. The formulation is defined as:
\begin{equation}
f_s(x, d_v) =
\begin{cases}
    \left( \mu - 0.5x \right) \left( \dfrac{x}{\mu} \right)^3, & 0 \le x < \mu,\\[8pt]
    x - 0.5\mu, & \mu \le x < d_v,\\[4pt]
    d_v - \left( \mu - 0.5\bar{x} \right) \left( \dfrac{\bar{x}}{\mu} \right)^3, & d_v \le x < d_v+\mu,\\[8pt]
    d_v, & x \ge d_v+\mu,
\end{cases}
\label{eq:fs_def}
\end{equation}
where $\bar{x} = d_v + \mu - x$, and $0<\mu < d_v$ is a small smoothing parameter. 

\begin{figure}[t]
\centering
\includegraphics[width=0.6\linewidth]{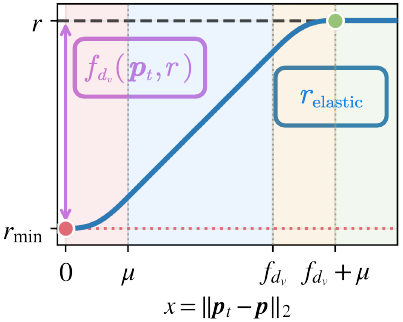}
\vspace{-0.1cm}
\caption{Illustration of the elastic radius $r_\mathrm{elastic}$ as a function of the distance $x = |\boldsymbol{p}_t - \boldsymbol{p}|_2$ from the target position.
}
\label{fig:relastic}
\vspace{-0.1cm}
\end{figure}

The elastic radius consists of four phases (Fig.~\ref{fig:relastic}):
\begin{enumerate}
    \item[(i)] Smooth Start ($0 \le x < \mu$): The radius starts at $r_\mathrm{min}$ with zero derivative at $x=0$, ensuring that the gradient of the elastic radius with respect to position vanishes at the exact intended contact point. It smoothly accelerates to a slope of 1 at $x=\mu$.
    \item[(ii)] Linear Recovery ($\mu \le x < d_v$): In this region, the elastic radius expands at a 1:1 ratio with the distance from the target, biasing the departure trajectory away from the support surface.
    \item[(iii)] Smooth Saturation ($d_v \le x < d_v+\mu$): The radius decelerates smoothly, transitioning the slope from 1 back to 0.
    \item[(iv)] Full Recovery ($x \ge d_v+\mu$): The radius saturates at $r$. 
\end{enumerate}

\paragraph{Object safety constraint.}
% During holding intervals, we enforce for each task $i\in\mathcal{I}_{\mathrm{hold}}$ and each sphere $m$,
Based on the elastic radius, we define the object safety constraint.
Let $\mathcal{I}_{\mathrm{hold}}$ denote the set of indices for tasks that result in the robot holding an object upon completion, such as a picking task. 
For each task $i\in\mathcal{I}_{\mathrm{hold}}$, the held object must remain safe from the end of the task $i$ to the start of the task $i+1$ (i.e., $\mathcal{S}_{\mathrm{hold},i}=[\bar{t}_{\kappa_{i,\mathrm{e}}},\bar{t}_{\kappa_{i+1,\mathrm{s}}}]$). 
During $\mathcal{S}_{\mathrm{hold},i}$, for each elastic collision sphere $m$ of the held object $\mathrm{o}_i$, we enforce:
\begin{equation}
\begin{aligned}
r_{\mathrm{elastic}}\bigl(\boldsymbol{p}_{\mathrm{t},i,\mathrm{o}_i,m},\boldsymbol{p}_{\mathrm{o}_i,m}(t),r_{\mathrm{o}_i,m}\bigr)+d_\mathrm{s}\\\leq
D_{\text{ESDF}}\bigl(\boldsymbol{p}_{\mathrm{o}_i,m}(t)\bigr)+d_{\mathrm{ela}},
\end{aligned}
\end{equation}
\begin{equation}
\begin{aligned}
r_{\mathrm{elastic}}\bigl(\boldsymbol{p}_{\mathrm{t},i+1,\mathrm{o}_i,m},\boldsymbol{p}_{\mathrm{o}_i,m}(t),r_{\mathrm{o}_i,m}\bigr)+d_\mathrm{s}\\\leq
D_{\text{ESDF}}\bigl(\boldsymbol{p}_{\mathrm{o}_i,m}(t)\bigr)+d_{\mathrm{ela}},
\end{aligned}
\end{equation}
where $\boldsymbol{p}_{\mathrm{o}_i,m}(t)=f_{\mathrm{FK,ee}}(\boldsymbol{x}(t))\,^{\mathrm{o}_i}\boldsymbol{P}_{\mathcal{C},i}^{-1}\,{}^{\mathrm{o}_i}\boldsymbol{p}_{\mathrm{o}_i,m}$ is the position of the $m$-th elastic collision sphere of the object $\mathrm{o}_i$ in the world frame when held by the MM with grasp pose $\,^{\mathrm{o}_i}\boldsymbol{P}_{\mathcal{C},i}$. The terms $\boldsymbol{p}_{\mathrm{t},i,\mathrm{o}_i,m}$ and $\boldsymbol{p}_{\mathrm{t},i+1,\mathrm{o}_i,m}$ are the task-specific sphere position for the end of the task $i$ and the start of the task $i+1$, respectively. $d_{\mathrm{ela}}$ is a small constraint tolerance introduced to account for finite optimization accuracy near intended-contact poses.

In summary, ECS resolves the contact--safety conflict by state-dependent radius modulation, which (i) preserves the reachability of intended-contact poses and (ii) biases departures toward increasing clearance to prevent unintended contacts through radius recovery.

\begin{figure*}[t]
\centering
\includegraphics[width=0.95\linewidth]{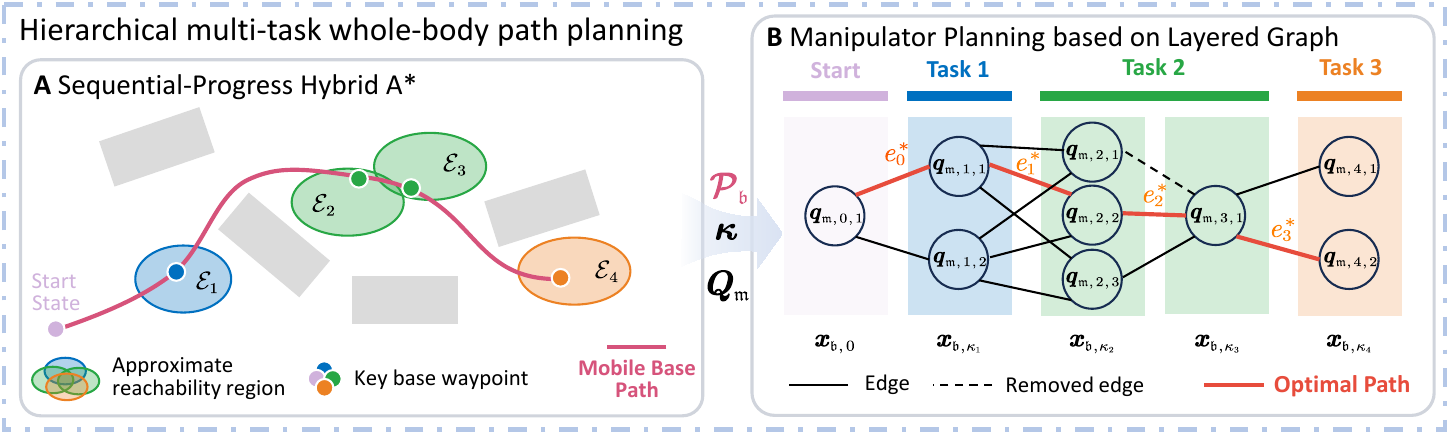}
\vspace{-0.1cm}
\caption{Overview of hierarchical multi-task whole-body path planner consisting of (\textbf{A}) Sequential-Progress Hybrid A* for the mobile base, and (\textbf{B}) a layered-graph shortest-path search for the manipulator.
}
\label{fig:path_planning}
\vspace{-0.1cm}
\end{figure*}

\subsection{Hierarchical Multi-task Whole-body Path Planning}
\label{subsec:path_planning}
The continuous-time optimization in Eq.~\eqref{eq:opt_prob} is computationally intensive due to the long planning horizon, the high dimensionality of the mobile manipulator, and the complex constraints coupled through non-linear forward kinematics (Fig.~\ref{fig:trajectory_optimization}). For online solving, it is crucial to provide the optimizer with a high-quality initialization. 

To achieve this efficiently, we propose a hierarchical path planning framework (Fig.~\ref{fig:path_planning}). First, we discretize the continuous task trajectories into a sequence of reachable keypoints. Next, a Sequential-Progress Hybrid A* algorithm searches for a collision-free mobile base path from which the end-effector can sequentially visit these keypoints (Sec.~\nameref{subsubsec:seq_astar}). Finally, we construct a layered graph to compute an optimal manipulator path conditioned on this base path (Sec.~\nameref{subsubsec:mani_path}). Seamlessly synchronizing the base and manipulator paths yields the complete collision-free whole-body path $\boldsymbol{\mathcal{P}}$, while simultaneously determining the task phase indices $\mathcal{K}$ and selecting a feasible, consistent grasp $\,^{\mathrm{o}_i}\boldsymbol{P}_{\mathcal{C},i}\in\boldsymbol{\mathcal{C}}_i$ for each task.

\subsubsection{Keypoint Sequence}
To enable discrete path search over continuous multi-task trajectories, we discretize each task trajectory into an ordered sequence of representative poses. Specifically, for each task trajectory $\boldsymbol{P}_{\mathrm{t},i}(\tau)$, $i\in\{1,\dots,N_{\mathcal{T}}\}$, we uniformly sample poses in time and concatenate them in task order to obtain a global keypoint sequence
$\boldsymbol{\mathfrak{p}}=\{\mathfrak{p}_1,\dots,\mathfrak{p}_{N_{\mathfrak{p}}}\}$, where each $\mathfrak{p}_k=\{\boldsymbol{P}_{\mathfrak{p},k},i_{\mathfrak{p},k}\}$ consists of a task pose $\boldsymbol{P}_{\mathfrak{p},k}\in\mathrm{SE}(3)$ sampled on $\boldsymbol{P}_{\mathrm{t},i}(\tau)$ and a task index $i_{\mathfrak{p},k}\in\{1,\cdots,N_{\mathcal{T}}\}$. The resulting keypoints serve as the ordered discrete targets for both the subsequent mobile base and manipulator path planning.

A keypoint $\mathfrak{p}_k$ is reachable from a base pose $\boldsymbol{x}_{\mathfrak{b}}$ if there exists a collision-free manipulator configuration $\boldsymbol{q}_{\mathfrak{m}}$ such that the end-effector reaches the task pose with one admissible grasp $\,^{\mathrm{o}_{i_{\mathfrak{p},k}}}\boldsymbol{P}_{\mathcal{C}}\in\boldsymbol{\mathcal{C}}_{i_{\mathfrak{p},k}}$ for that task:
\begin{equation}
\begin{aligned}
&\exists\,\boldsymbol{q}_{\mathfrak{m}},\,^{\mathrm{o}_{i_{\mathfrak{p},k}}}\boldsymbol{P}_{\mathcal{C}}\in\boldsymbol{\mathcal{C}}_{i_{\mathfrak{p},k}},\\ 
&\text{s.t.}\ \ f_{\mathrm{FK},\mathrm{ee}}\big([\boldsymbol{x}_{\mathfrak{b}}^\top,\boldsymbol{q}_{\mathfrak{m}}^\top]^\top\big)=\boldsymbol{P}_{\mathfrak{p},k}\,^{\mathrm{o}_{i_{\mathfrak{p},k}}}\boldsymbol{P}_{\mathcal{C}}.
\end{aligned}
\label{eq:keypoint_reachability}
\end{equation}

\begin{algorithm}[t]
\caption{Sequential-Progress Hybrid A*}
\label{alg:progress_astar}
\begin{algorithmic}[1]
\Require Initial base pose $\boldsymbol{x}_{\mathfrak{b},0}$, initial manipulator state $\boldsymbol{q}_{\mathfrak{m},0}$, initial grasp pose set $\boldsymbol{\mathcal{C}}_{\mathrm{init}}$, keypoints $\{\mathfrak{p}_k\}_{k=1}^{N_{\mathfrak{p}}}$, reachability ellipses $\{\mathcal{E}_k\}_{k=1}^{N_{\mathfrak{p}}}$
\Ensure Base path $\boldsymbol{\mathcal{P}}_{\mathfrak{b}}$, key base waypoint indices $\boldsymbol{\kappa}$, manipulator configs $\boldsymbol{Q}_{\mathfrak{m}}$
\State  Create node $\mathcal{N}_0$ with $\mathcal{N}_0.\boldsymbol{x}_{\mathfrak{b}}\gets\boldsymbol{x}_{\mathfrak{b},0}$, $\mathcal{N}_0.\mathfrak{n}\gets1$, $\mathcal{N}_0.g\gets0$, $\mathcal{N}_0.h\gets$ ProgressAwareHeuristic$(\boldsymbol{x}_{\mathfrak{b},0}, 1,N_{\mathfrak{p}},\{\mathcal{E}_k\}_{k=1}^{N_{\mathfrak{p}}})$, $\mathcal{N}_0.\boldsymbol{Q}_{\mathrm{valid}} \gets \{\boldsymbol{q}_{\mathfrak{m},0}\}$, $\mathcal{N}_0.\boldsymbol{\mathcal{C}}_{\mathrm{valid}} \gets \boldsymbol{\mathcal{C}}_{\mathrm{init}}$
\State Initialize open set $\mathcal{O} \gets \{ \mathcal{N}_0 \}$ and close set $\mathcal{C} \gets \emptyset$
\While{$\mathcal{O} \neq \emptyset$}
    \State Pop node $\mathcal{N}$ with minimum $g + h$ from $\mathcal{O}$
    \If{$\mathcal{N}.\mathfrak{n} = N_{\mathfrak{p}} + 1$}
        \State \Return $\boldsymbol{\mathcal{P}}_{\mathfrak{b}}, \boldsymbol{\kappa}, \boldsymbol{Q}_{\mathfrak{m}}$ by backtracking
        % \Comment{Search terminates successfully}
    \EndIf
    \ForAll{each successor base pose $\boldsymbol{x}'_{\mathfrak{b}}$ of $\mathcal{N}.\boldsymbol{x}_{\mathfrak{b}}$}
        \State ${g}' \gets \mathcal{N}.g+\Delta g'$, $\mathfrak{n}' \gets \mathcal{N}.\mathfrak{n}$
        \State $\boldsymbol{Q}'_{\mathrm{valid}},\boldsymbol{\mathcal{C}}'_{\mathrm{valid}}\gets$ReachabilityCheck($\boldsymbol{x}'_{\mathfrak{b}},\mathcal{N},\mathfrak{p}_\mathfrak{n},\mathcal{E}_{\mathfrak{n}}$)
        \If{$\boldsymbol{Q}'_{\mathrm{valid}}\neq \emptyset$}
            \State $\mathfrak{n}' \gets \mathcal{N}.\mathfrak{n} + 1$
        \EndIf
        \State $h'\gets$ProgressAwareHeuristic($\boldsymbol{x}'_{\mathfrak{b}}, \mathfrak{n}',N_{\mathfrak{p}},\{\mathcal{E}_k\}_{k=1}^{N_{\mathfrak{p}}}$)
        \State $\mathcal{N}'\gets$Node($\boldsymbol{x}'_{\mathfrak{b}}, \mathfrak{n}',g',h',\boldsymbol{Q}'_{\mathrm{valid}},\boldsymbol{\mathcal{C}}'_{\mathrm{valid}},\mathcal{N}$)
        \If{$\mathcal{N}'\notin\mathcal{C}$}
            \State $\mathcal{O}$.Update($\mathcal{N}'$) 
            % \Comment{Push $\mathcal{N}'$ or update node in $\mathcal{O}$}
        \EndIf
    \EndFor
    \State $\mathcal{C}$.Push$(\mathcal{N})$
\EndWhile
\end{algorithmic}
\end{algorithm}

\subsubsection{Sequential-Progress Hybrid A*}
\label{subsubsec:seq_astar}
\paragraph{Overview}
We first plan a collision-free mobile-base path that visits the keypoints in sequence (Fig.~\ref{fig:path_planning}A). The overall planner is outlined in Algo.~\ref{alg:progress_astar}. To encode sequential progress, we extend Hybrid A* \cite{dolgov2010path} by augmenting the search state from the base pose $\boldsymbol{x}_{\mathfrak{b}}$ to $(\boldsymbol{x}_{\mathfrak{b}},\mathfrak{n})$, where $\mathfrak{n}$ indexes the next target keypoint. The progress index $\mathfrak{n}$ advances by one whenever the current keypoint $\mathfrak{p}_{\mathfrak{n}}$ is reachable from the evaluated base pose $\boldsymbol{x}_{\mathfrak{b}}$. 

The search terminates successfully when a node with $\mathfrak{n}=N_{\mathfrak{p}}+1$ is popped from the open set. At this point, backtracking yields a short base path $\boldsymbol{\mathcal{P}}_{\mathfrak{b}}=\{\boldsymbol{x}_{\mathfrak{b},0},\dots,\boldsymbol{x}_{\mathfrak{b},M}\}$ alongside key base waypoint indices $\boldsymbol{\kappa}=\{\kappa_1,\dots,\kappa_{N_{\mathfrak{p}}}\}$, where $\boldsymbol{x}_{\mathfrak{b},\kappa_k}$ denotes the specific waypoint from which $\mathfrak{p}_k$ is reachable. Additionally, the planner outputs the collected manipulator configurations $\boldsymbol{Q}_{\mathfrak{m}}=\{\boldsymbol{Q}_{1},\dots,\boldsymbol{Q}_{N_{\mathfrak{p}}}\}$, where each $\boldsymbol{Q}_{k}=\{\boldsymbol{q}_{\mathfrak{m},k,i}\}_i$ corresponds to the valid joint set $\boldsymbol{Q}_{\mathrm{valid}}$ stored in the node associated with $\boldsymbol{x}_{\mathfrak{b},\kappa_k}$.

To evaluate the state transitions and guide the search efficiently, the main algorithm relies on two core sub-routines: a two-stage reachability verification (Algo.~\ref{alg:reachability}) and a progress-aware heuristic (Algo.~\ref{alg:heuristic}), detailed as follows.

\paragraph{Reachability verification}
To efficiently evaluate the reachability condition in Eq.~\eqref{eq:keypoint_reachability}, we adopt a two-stage verification procedure (Algo.~\ref{alg:reachability}). Given a candidate base pose $\boldsymbol{x}_{\mathfrak{b}}$, we first apply a lightweight geometric filter to prune poses that are unlikely to reach $\mathfrak{p}_{\mathfrak{n}}$. For the remaining candidates, we perform an exact kinematic check based on inverse kinematics (IK). This second stage explicitly enforces grasp consistency and task-trajectory constraints across continuous manipulation sequences.

\begin{algorithm}[t]
\caption{Reachability Verification}
\label{alg:reachability}
\begin{algorithmic}[1]
\Require Base pose $\boldsymbol{x}_{\mathfrak{b}}$, current node $\mathcal{N}$, keypoint $\mathfrak{p}_\mathfrak{n}$, reachability ellipse $\mathcal{E}_\mathfrak{n}$
\Ensure valid joint configurations set $\boldsymbol{Q}_{\mathrm{valid}}$, valid grasp poses set $\boldsymbol{\mathcal{C}}_{\mathrm{valid}}$
\State $\boldsymbol{Q}_{\mathrm{valid}} \gets \emptyset$, $\boldsymbol{\mathcal{C}}_{\mathrm{valid}}\gets \emptyset$, $\boldsymbol{q}_{\mathfrak{b}}\gets$ GetPosition($\boldsymbol{x}_{\mathfrak{b}}$)
\If{$\boldsymbol{q}_{\mathfrak{b}}\in\mathcal{E}_{\mathfrak{n}}$} \Comment{Fast geometric filter}
\Statex \hspace*{-\algorithmicindent}\textbf{Step A: Grasp-consistency enforcement}
    \If{IsContinuousOperation($\mathfrak{p}_\mathfrak{n}$)}
        \State $\mathcal{N}_\mathrm{key}=\mathcal{N}$
        \While{$\mathcal{N}_\mathrm{key}.\boldsymbol{\mathcal{C}}_{\mathrm{valid}}=\emptyset$} \Comment{Find last key node}
            \State $\mathcal{N}_\mathrm{key}=\mathcal{N}_\mathrm{key}.$Parent
        \EndWhile
        \State $\boldsymbol{\mathcal{C}}_{\mathrm{cand}} \gets \mathcal{N}_\mathrm{key}.\boldsymbol{\mathcal{C}}_{\mathrm{valid}}$
        % \State $\boldsymbol{Q}_{\mathrm{cand}} \gets \mathcal{N}_\mathrm{key}.\boldsymbol{Q}_{\mathrm{valid}}$
    \Else
        \State $\boldsymbol{\mathcal{C}}_{\mathrm{cand}} \gets \boldsymbol{\mathcal{C}}_{i_{\mathfrak{p},\mathfrak{n}}}$
        % \State $\boldsymbol{Q}_{\mathrm{cand}} \gets \emptyset$
    \EndIf
\Statex \hspace*{-\algorithmicindent}\textbf{Step B: Feasibility verification}
    \ForAll{$\,^{\mathrm{o}}\boldsymbol{P}_{\mathcal{C}}\in \boldsymbol{\mathcal{C}}_{\mathrm{cand}}$}
        \State $Q_{\mathrm{IK}}\gets$ ComputeIK$(\boldsymbol{x}_{\mathfrak{b}},\boldsymbol{P}_{\mathfrak{p},\mathfrak{n}},\,^{\mathrm{o}}\boldsymbol{P}_{\mathcal{C}})$
        \ForAll{$\boldsymbol{q}_{\mathfrak{m}}\in Q_{\mathrm{IK}}$}
        \If{FeasibilityCheck($\boldsymbol{q}_{\mathfrak{m}},\mathcal{N}$)}
                % \Comment{Progress index increases}
                \State $\boldsymbol{Q}_{\mathrm{valid}}$.Push ($\boldsymbol{q}_{\mathfrak{m}}$)  
                \State $\boldsymbol{\mathcal{C}}_{\mathrm{valid}}$.Push ($\,^{\mathrm{o}}\boldsymbol{P}_{\mathcal{C}}$)
                % \Comment{Store valid configurations}
            \EndIf
        \EndFor
    \EndFor
\EndIf
\State \Return $\boldsymbol{Q}_{\mathrm{valid}}$, $\boldsymbol{\mathcal{C}}_{\mathrm{valid}}$
\end{algorithmic}
\end{algorithm}

\textit{Stage 1: Fast Geometric Filter.}
For each keypoint $\mathfrak{p}_i$, we pre-compute a 2D reachability region $\mathcal{E}_i \subset \mathbb{R}^2$ to approximate the set of valid base placements (Fig.~\ref{fig:path_planning}A). Derived from the inverse reachability map \cite{vahrenkamp2013robot}, this region is parameterized as an ellipse:
\begin{equation}
    \mathcal{E}_{i} = \left\{ \boldsymbol{o}_{\mathcal{E},i} + \boldsymbol{R}_{\mathcal{E},i} \boldsymbol{Q}_{\mathcal{E},i} \boldsymbol{v} \;\middle|\; \boldsymbol{v}\in\mathbb{R}^2, \, \|\boldsymbol{v}\|_2 \le 1 \right\}.
\end{equation}
% where $\boldsymbol{o}_{\mathcal{E},i} \in \mathbb{R}^2$, $\boldsymbol{R}_{\mathcal{E},i} \in \mathrm{SO}(2)$ and $\boldsymbol{Q}_{\mathcal{E},i} = \mathrm{diag}(a_k, b_k)$ denote the center, orientation, and semi-axes matrix of $\mathcal{E}_i$, respectively. 
A successor base pose $\boldsymbol{x}'_{\mathfrak{b}}$ is considered a valid candidate for $\mathfrak{p}_{\mathfrak{n}}$ only if its position component $\boldsymbol{q}'_{\mathfrak{b}}$ lies within this region (i.e. $\boldsymbol{q}'_{\mathfrak{b}}\in\mathcal{E}_{\mathfrak{n}}$).

Since $\mathcal{E}_i$ serves only as a coarse geometric filter, inclusion within this region does not guarantee a feasible and collision-free manipulator configuration. Therefore, candidates that pass this filter must undergo an exact IK-based verification, which addresses two critical requirements: grasp consistency (Step A of Algo.~\ref{alg:reachability}) and kinematic feasibility (Step B of Algo.~\ref{alg:reachability}).

\textit{Stage 2, Step A: Grasp-Consistency Enforcement.}
Successful manipulation often requires the end-effector to maintain an invariant grasp pose relative to the object over a continuous operation sequence, which applies both to multiple keypoints within a single task (e.g., maintaining the same grasp while closing a drawer) and to semantically coupled tasks (e.g., a placing task following a picking task). 

To preserve this consistency, each search node maintains a valid grasp set, $\boldsymbol{\mathcal{C}}_{\mathrm{valid}}$. We define the key nodes as search nodes that successfully reach the keypoint. Only key nodes have a nonempty $\boldsymbol{\mathcal{C}}_{\mathrm{valid}}$, which stores the subset of grasp poses for which feasible manipulator configurations exist to reach the corresponding keypoint. When evaluating reachability for $\mathfrak{p}_{\mathfrak{n}}$, we determine a candidate grasp pose set $\boldsymbol{\mathcal{C}}_{\mathrm{cand}}$ according to whether $\mathfrak{p}_{\mathfrak{n}}$ belongs to a continuous operation. If $\mathfrak{p}_{\mathfrak{n}}$ is an independent operation or the first keypoint of a continuous operation, the robot can select any task-compatible grasp pose (i.e. $\boldsymbol{\mathcal{C}}_{\mathrm{cand}} = \boldsymbol{\mathcal{C}}_{i_{\mathfrak{p},\mathfrak{n}}}$). Otherwise, we backtrack along the search tree to the last key node $\mathcal{N}_{\mathrm{key}}$ and inherit its valid grasp pose
$\boldsymbol{\mathcal{C}}_{\mathrm{cand}}=\mathcal{N}_{\mathrm{key}}.\boldsymbol{\mathcal{C}}_{\mathrm{valid}}$. The IK solver is subsequently evaluated only over $\boldsymbol{\mathcal{C}}_{\mathrm{cand}}$, ensuring grasp consistency throughout the continuous manipulation segment.

\textit{Stage 2, Step B: Feasibility verification.}
For each candidate grasp pose in $\boldsymbol{\mathcal{C}}_{\mathrm{cand}}$, we compute IK solutions~\cite{diankov2010automated} to reach $\boldsymbol{P}_{\mathfrak{p},\mathfrak{n}}$ from the base pose $\boldsymbol{x}_{\mathfrak{b}}$. Each solution is then subjected to a feasibility check. In addition to standard safety and joint limit validations, we verify task-trajectory consistency for consecutive keypoints within the same task, which is necessary for operations with explicit geometric execution constraints, such as the straight-line end-effector motion required during drawer opening. Specifically, we reconstruct the synchronized whole-body motion by combining the searched base path with linearly interpolated arm joint configurations, traced back to the previous keypoint node. We then compare the resulting end-effector trajectory against the desired task trajectory, ensuring the deviation remains below a predefined threshold (detailed in \nameref{app:front_end}). Only IK solutions satisfying all criteria are retained, with their corresponding joint configurations and grasp poses stored in $\boldsymbol{Q}_{\mathrm{valid}}$ and $\boldsymbol{\mathcal{C}}_{\mathrm{valid}}$, respectively. 

\begin{algorithm}[t]
\caption{Progress-Aware Heuristic}
\label{alg:heuristic}
\begin{algorithmic}[1]
\Require Base state $\boldsymbol{x}_{\mathfrak{b}}$, progress index $\mathfrak{n}$
, total keypoints $N_{\mathfrak{p}}$, reachability ellipses $\{\mathcal{E}_k\}_{k=1}^{N_{\mathfrak{p}}}$
\Ensure Heuristic value $h$
\State $\boldsymbol{q}_{\mathfrak{b}}\gets$ GetPosition$(\boldsymbol{x}_{\mathfrak{b}})$
% \State $k_{\mathrm{effect}},\boldsymbol{p}_{\text{start}} \gets$ Effective$(\boldsymbol{q}_{\mathfrak{b}},\mathfrak{n},N_{\mathfrak{p}},\{\mathcal{E}_k\}_{k=1}^{N_{\mathfrak{p}}})$
\State $k_{\mathrm{effect}} \gets \mathfrak{n}$
\State $h\gets0$
\Statex \hspace*{-\algorithmicindent}\textbf{Step 1: Determine effective start index and position}
\While{$\boldsymbol{q}_{\mathfrak{b}} \in \mathcal{E}_{k_{\mathrm{effect}}}$ \textbf{and} $k_{\mathrm{effect}} \le N_{\mathfrak{p}}$} 
    \State $k_{\mathrm{effect}} \gets k_{\mathrm{effect}} + 1$
\EndWhile
\If{$k_{\mathrm{effect}}=N_{\mathfrak{p}}+1$}
    \State \Return h
\EndIf
% \Statex \hspace*{-\algorithmicindent}\textbf{Step 2: Determine starting position} $\boldsymbol{p}_{\text{start}}$
\If{$k_{\mathrm{effect}} = \mathfrak{n}$} 
    \State $\boldsymbol{p}_{\text{start}} \gets \boldsymbol{q}_{\mathfrak{b}}$ 
    % \Comment{No regions skipped}
\Else
    \State $\boldsymbol{p}_{\text{start}} \gets \mathop{\arg\min}\limits_{\boldsymbol{p} \in \partial\mathcal{E}_{\mathfrak{n}}} d_{\boldsymbol{p}\mathcal{E}}(\boldsymbol{p}, \mathcal{E}_{k_{\mathrm{effect}}})$ 
    % \Comment{Boundary point of $\mathcal{E}_{\mathfrak{n}}$ closest to $\mathcal{E}_{k_{\mathrm{effect}}}$}
\EndIf
\Statex \hspace*{-\algorithmicindent}\textbf{Step 2: Calculate heuristic value}
\State $\boldsymbol{c}_{\mathcal{E},\mathrm{last}}\gets\boldsymbol{p}_{\mathrm{start}}$
\For{$k = k_{\mathrm{effect}}$ \textbf{to} $N_{\mathfrak{p}}-1$}
    \State $\boldsymbol{c}_{\mathcal{E}} \gets \mathop{\arg\min}\limits_{\boldsymbol{p} \in \partial \mathcal{E}_k} \|\boldsymbol{c}_{\mathcal{E},\mathrm{last}} - \boldsymbol{p}\|_2 + d_{\boldsymbol{p}\mathcal{E}}(\boldsymbol{p}, \mathcal{E}_{k+1})$
    \State $h \gets h + \left\| \boldsymbol{c}_{\mathcal{E},\mathrm{last}}-\boldsymbol{c}_{\mathcal{E}} \right\|_2$
    \State $\boldsymbol{c}_{\mathcal{E},\mathrm{last}}\gets\boldsymbol{c}_{\mathcal{E}}$
\EndFor
% \If{$k_{\mathrm{effect}} \le N_{\mathfrak{p}}-1$}
    \State $h \gets h + d_{\boldsymbol{p}\mathcal{E}}(\boldsymbol{c}_{\mathcal{E},\mathrm{last}},\mathcal{E}_{N_{\mathfrak{p}}})$
% \EndIf
\State \Return h
\end{algorithmic}
\end{algorithm}

\paragraph{Progress-aware heuristic}
To accelerate the search by guiding the planner through the sequence of remaining reachability regions, we propose a progress-aware heuristic $h(\boldsymbol{x}_{\mathfrak{b}}, \mathfrak{n})$. This heuristic estimates the remaining travel distance required to visit the remaining reachability ellipses $\{\mathcal{E}_k\}_{k=\mathfrak{n}}^{N_{\mathfrak{p}}}$ sequentially from the current state $\boldsymbol{x}_{\mathfrak{b}}$. The computation consists of two main steps (Algo.~\ref{alg:heuristic}).

\textit{Step 1: Determine effective start index and position.}
To prevent the heuristic from guiding the robot back toward the boundary of the region it has already satisfied, we first determine the effective start index $k_{\mathrm{effect}}$ ($k_{\mathrm{effect}} \ge \mathfrak{n}$). If the current base position $\boldsymbol{q}_{\mathfrak{b}}$ already resides within $\mathcal{E}_{k_{\mathrm{effect}}}$ with initially $k_{\mathrm{effect}}=\mathfrak{n}$, we iteratively increment $k_{\mathrm{effect}}$ until $\boldsymbol{q}_{\mathfrak{b}} \notin \mathcal{E}_{k_{\mathrm{effect}}}$. 

Next, we determine the starting position of the heuristic path, denoted as $\boldsymbol{p}_{\mathrm{start}}$. If no regions are skipped ($k_{\mathrm{effect}} = \mathfrak{n}$), we simply set $\boldsymbol{p}_{\text{start}} = \boldsymbol{q}_{\mathfrak{b}}$. However, if regions are skipped ($k_{\mathrm{effect}} > \mathfrak{n}$), $\boldsymbol{p}_{\text{start}}$ is selected as the point on the boundary of the current target region ($\partial\mathcal{E}_{\mathfrak{n}}$) that minimizes the distance to the next effective region $\mathcal{E}_{k_{\mathrm{effect}}}$:
\begin{equation}
\label{eq:p_two_eli}
\boldsymbol{p}_{\mathrm{start}}=\mathop{\arg\min}\limits_{\boldsymbol{p} \in \partial\mathcal{E}_{\mathfrak{n}}}\ d_{\boldsymbol{p}\mathcal{E}}(\boldsymbol{p}, \mathcal{E}_{k_{\mathrm{effect}}}).
\end{equation}
The function $d_{\boldsymbol{p}\mathcal{E}}(\boldsymbol{p}, \mathcal{E})$ calculates the distance from a query point $\boldsymbol{p} \in \mathbb{R}^2$ to an ellipse $\mathcal{E}$:
\begin{equation}
\begin{aligned}
    d_{\boldsymbol{p}\mathcal{E}}&(\boldsymbol{p}, \mathcal{E}) =\\
    &\begin{cases}
        0, & \text{if } \| \tilde{\boldsymbol{p}} \|_2 \le 1, \\
        \left\| \boldsymbol{p} - \left(\boldsymbol{o}_{\mathcal{E}} + \boldsymbol{R}_{\mathcal{E}}\boldsymbol{Q}_{\mathcal{E}} \frac{\tilde{\boldsymbol{p}}}{\| \tilde{\boldsymbol{p}} \|_2}\right) \right\|_2, & \text{otherwise},
    \end{cases}
    \label{eq:dist_p_E}
\end{aligned}
\end{equation}
where $\tilde{\boldsymbol{p}} = \boldsymbol{Q}_{\mathcal{E}}^{-1}\boldsymbol{R}_{\mathcal{E}}^{\top}(\boldsymbol{p} - \boldsymbol{o}_{\mathcal{E}})$ transforms the query point into the ellipse's canonical frame. In practice, we solve Eq.~\eqref{eq:p_two_eli} efficiently by uniformly sampling points along the boundary $\partial\mathcal{E}_{\mathfrak{n}}$.

\textit{Step 2: Calculate heuristic value.}
Computing the exact shortest path that goes through the remaining regions $\{\mathcal{E}_k\}_{k=k_{\mathrm{effect}}}^{N_{\mathfrak{p}}}$ is computationally expensive. Therefore, we approximate the remaining travel distance using a greedy piecewise-linear path. Starting from $\boldsymbol{c}_{\mathcal{E},\mathrm{last}} = \boldsymbol{p}_{\mathrm{start}}$, the algorithm incrementally accumulates the lengths of line segments connecting representative boundary points. These points are selected sequentially via a one-step lookahead rule. For each region $\mathcal{E}_k$, we select the boundary point $\boldsymbol{c}_{\mathcal{E}} \in \partial \mathcal{E}_k$ that minimizes the sum of the distance from the previous point $\boldsymbol{c}_{\mathcal{E},\mathrm{last}}$ and the shortest distance to the subsequent region $\mathcal{E}_{k+1}$:
\begin{equation}
\begin{aligned}
    \boldsymbol{c}_{\mathcal{E}} = \mathop{\mathrm{arg\,min}}_{\boldsymbol{p}} \
    & \|\boldsymbol{c}_{\mathcal{E},\mathrm{last}} - \boldsymbol{p}\|_2 + d_{\boldsymbol{p}\mathcal{E}}(\boldsymbol{p}, \mathcal{E}_{k+1}), \\
    \text{s.t.}\ & \boldsymbol{p} \in \partial \mathcal{E}_k.
\end{aligned}
\label{eq:heu_lookahead}
\end{equation}
Similar to Eq.~\eqref{eq:p_two_eli}, we solve Eq.~\eqref{eq:heu_lookahead} efficiently in practice by uniformly sampling candidate points along the boundary $\partial \mathcal{E}_k$. The term $d_{\boldsymbol{p}\mathcal{E}}(\boldsymbol{p}, \mathcal{E}_{k+1})$ acts as a potential field, biasing the selection of $\boldsymbol{c}_{\mathcal{E}}$ toward the entrance of the next region. This aligns the piecewise linear path with the global task direction. The final heuristic $h$ is the total length of this path, concluding with the distance to the final region $\mathcal{E}_{N_{\mathfrak{p}}}$.

\subsubsection{Manipulator Path Planning based on Layered Graph}
\label{subsubsec:mani_path}
Given the base path $\boldsymbol{\mathcal{P}}_{\mathfrak{b}}$, waypoint indices $\boldsymbol{\kappa}$, and valid manipulator configurations $\boldsymbol{Q}_{\mathfrak{m}}$, we plan manipulator path between successive key base waypoints. We formulate this as a shortest-path problem on a layered graph (Fig.~\ref{fig:path_planning}B). The first layer contains the starting configuration $\boldsymbol{q}_{\mathfrak{m},0}$ at base pose $\boldsymbol{x}_{\mathfrak{b},0}$. Each subsequent $k$-th layer corresponds to the keypoint $\mathfrak{p}_k$, where nodes represent the feasible manipulator configurations stored in $\boldsymbol{Q}_{k}$ at base waypoint $\boldsymbol{x}_{\mathfrak{b},\kappa_k}$.

Directed edges connect nodes in adjacent layers. The existence of an edge is determined by successfully generating a local, collision-free manipulator path~\cite{wu2024real}, constrained by the concurrent base motion along the corresponding segment of $\boldsymbol{\mathcal{P}}_{\mathfrak{b}}$. The cost of the valid edge is defined as the length of the generated path, and the path itself is temporarily cached within the edge. To ensure valid execution, we explicitly prune edges (e.g., the dashed line in Fig.~\ref{fig:path_planning}B) that (i) violate the grasp consistency constraints or (ii) deviate excessively from the task trajectory between consecutive keypoints that belong to the same task.  

We then apply Dijkstra's algorithm~\cite{dijkstra2022note} to find the shortest path through the graph (red edges $e_k^*$ in Fig.~\ref{fig:path_planning}B). By concatenating the cached manipulator paths along this optimal sequence and synchronizing them with base path $\boldsymbol{\mathcal{P}}_{\mathfrak{b}}$, we reconstruct a discrete whole-body path $\boldsymbol{\mathcal{P}}=\{\boldsymbol{x}_0,\dots,\boldsymbol{x}_M\}$ that sequentially visits all keypoints. Furthermore, this process yields (i) a selected grasp pose $\,^{\mathrm{o}_i}\boldsymbol{P}_{\mathcal{C},i}\in\boldsymbol{\mathcal{C}}_i$ for each task, and (ii) execution segments $\mathcal{K}=\{(\kappa_{i,\mathrm{s}},\kappa_{i,\mathrm{e}})\}_{i=1}^{N_{\mathcal{T}}}$ for each task mapped onto $\boldsymbol{\mathcal{P}}$.\\

Finally, $(\boldsymbol{\mathcal{P}},\{\,^{\mathrm{o}_i}\boldsymbol{P}_{\mathcal{C},i}\},\mathcal{K})$ is used to initialize the spatial-temporal trajectory optimization in Eq.~\eqref{eq:opt_prob}. And it is solved using the Powell-Hestenes-Rockafellar Augmented Lagrangian Method (PHR-ALM)~\cite{rockafellar1974augmented}, which refines a time-parameterized whole-body trajectory $\boldsymbol{x}(t),t\in[0,\bar{t}_M]$.

\section{Safe-warping-based Phase-dependent Controller}
\label{sec:controller}

\begin{figure*}[t]
\centering
\includegraphics[width=0.9\linewidth]{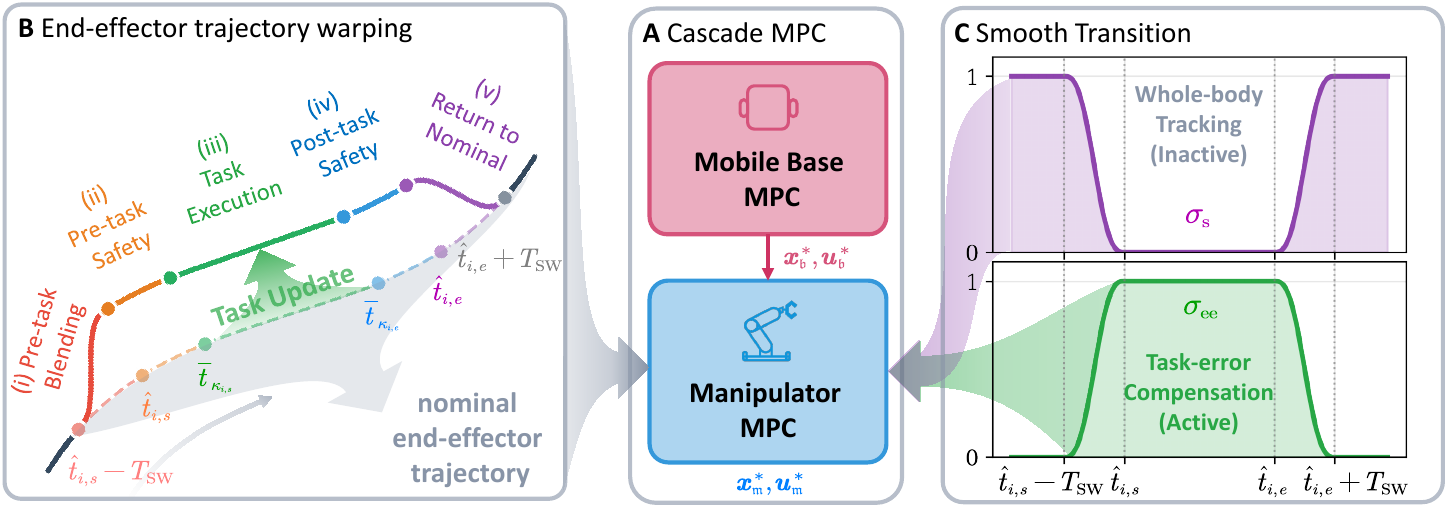}
\vspace{-0.1cm}
\caption{Safe-warping-based phase-dependent controller.
(\textbf{A}) a cascaded model predictive controller, (\textbf{B}) end-effector trajectory warping incorporates online task pose estimation into a safe compensation motion, and (\textbf{C}) smooth switching weights transition between whole-body trajectory tracking and task-error compensation.
}
\label{fig:controller}
\vspace{-0.4cm}
\end{figure*}

The solution to Eq.~\eqref{eq:opt_prob} provides a time-efficient and reliability-aware whole-body reference trajectory $\boldsymbol{x}(t)$. 
However, during execution, directly tracking $\boldsymbol{x}(t)$ can still lead to task failure due to uncertainty in the true manipulated object pose $\boldsymbol{P}^\star_{\mathrm{o}_i}$. 
In particular, the latest pose estimate $\hat{\boldsymbol{P}}_{\mathrm{o}_i}$ may become available after the final planning cycle preceding the execution of task $\mathcal{T}_i$, creating a mismatch between the planned and actual task conditions.

To improve robustness to pose uncertainty while maintaining the efficiency and reliability of the planned global trajectory, we develop a safe-warping-based phase-dependent controller (Fig.~\ref{fig:controller}). 
It smoothly switches between (i) global trajectory tracking during task-noncritical phases, prioritizing time efficiency, and (ii) real-time task error compensation during task-critical phases, enhancing reliability by compensating state and pose errors induced by real-world uncertainties in real time using the warped end-effector reference (Fig.~\ref{fig:controller}B) together with smoothly scheduled cost weights (Fig.~\ref{fig:controller}C).

\subsection{Controller Overview}
\label{subsec:controller_overview}
The controller is implemented using a cascaded MPC architecture (Fig.~\ref{fig:controller}A): at each control cycle, a mobile-base MPC first plans a short-horizon reactive base motion for trajectory tracking and collision avoidance; conditioned on this predicted base motion, a manipulator MPC optimizes joint commands to track the global trajectory and compensate task errors during the task-critical phase using smoothly scheduled cost weights (Sec.~\nameref{subsec:smoo_switch}). 
Crucially, our safety-preserving trajectory warping (Sec.~\nameref{subsec:traj_warp}) modifies the end-effector reference so that online task pose updates can be incorporated without destroying the planned pre-/post-task safe-interaction structure, thereby maintaining safety among the end-effector, the manipulated object, and the environment during real-time compensation. 
% Moreover, dynamic obstacles are handled online by embedding proximity penalties (and, if applicable, constraints) into the MPC objectives (summarized in Sec.~\nameref{subsec:cascaded_mpc_summary} and detailed in Appendix~\ref{app:mpc_formulation}).

% Let $N$ denote the prediction horizon and $\Delta t$ the time step. Denote $t_{\mathrm{ref}}$ as the current time on the reference trajectory $\boldsymbol{x}(t)$. We define the prediction time steps as $t_k = t_{\mathrm{ref}} + k\Delta t$ for $k=0, \dots, N$. 

\subsection{Safety-preserving Trajectory Warping}
\label{subsec:traj_warp}
During execution, there may be a mismatch between the nominal end-effector reference $\boldsymbol{P}_{\mathrm{ee}}^{\mathrm{ref}}(t)=f_{\mathrm{FK,ee}}(\boldsymbol{x}(t)),t\in[\bar{t}_{\kappa_{i,\mathrm{s}}}, \bar{t}_{\kappa_{i,\mathrm{e}}}]$ and the latest task end-effector trajectory $\bar{\boldsymbol{P}}_{\mathrm{t},i}(\tau),\tau\in[0,T_{\mathcal{T},i}]$ (i.e., $\boldsymbol{P}_{\mathrm{ee}}^{\mathrm{ref}}(\bar{t}_{\kappa_{i,\mathrm{s}}}+\tau)\neq\bar{\boldsymbol{P}}_{\mathrm{t},i}(\tau)$) due to the task pose estimate update.
Naively replacing $\boldsymbol{P}_{\mathrm{ee}}^{\mathrm{ref}}(t)$ with $\bar{\boldsymbol{P}}_{\mathrm{t},i}(\tau)$ is infeasible, because the planned safe interaction motions around task are crucial for reliable and safe task execution. 
Direct replacement would destroy these motions and introduce discontinuities.
We therefore propose a safety-preserving trajectory warping strategy that retargets the task-related segments to the latest estimated task trajectory while preserving the nominal relative safe motions before and after the task-critical phase (Fig.~\ref{fig:controller}B).

\subsubsection{Task-critical Phase and Warping Window}
For each task $i$, we define a task-critical phase
\begin{equation}
    \mathcal{S}_{\mathrm{mani},i} = [\hat{t}_{i,\mathrm{s}}, \hat{t}_{i,\mathrm{e}}],
\end{equation}
where $\hat{t}_{i,\mathrm{s}} = \bar{t}_{\kappa_{i,\mathrm{s}}} - T_{\mathrm{s},i}$ and $\hat{t}_{i,\mathrm{e}} = \bar{t}_{\kappa_{i,\mathrm{e}}} + T_{\mathrm{e},i}$. $\mathcal{S}_{\mathrm{mani},i}$ extends the nominal task interval $[\bar{t}_{\kappa_{i,\mathrm{s}}}, \bar{t}_{\kappa_{i,\mathrm{e}}}]$ by pre-/post-task safe-interaction durations $T_{\mathrm{s},i}$ and $T_{\mathrm{e},i}$, which covers the aforementioned pre- and post-task safe interaction motions generated by the trajectory planner. To ensure seamless transitions, we introduce a switching duration $T_{\mathrm{sw}}$ and define the warping window
\begin{equation}
    \mathcal{S}^{+}_{\mathrm{mani},i} := [\hat{t}_{i,\mathrm{s}}-T_{\mathrm{sw}},\, \hat{t}_{i,\mathrm{e}}+T_{\mathrm{sw}}).
\end{equation}
In our setting, we assume the windows $\{\mathcal{S}^{+}_{\mathrm{mani},i}\}$ do not overlap for all tasks, so at any time there exists at most one active task index $i$.

\subsubsection{Preserving Nominal Relative Safe Motions}
The key idea is to preserve the nominal relative end-effector motion around the task, but expressed in the latest task frame. Specifically, we define the nominal relative end-effector motions with respect to the task start and end end-effector poses, covering the pre- and post-task segments together with the switching intervals, as
\begin{equation*}
\begin{aligned}
    \boldsymbol{P}^{\mathrm{rel}}_{i,\mathrm{s}}(t) &= \boldsymbol{P}_{\mathrm{ee}}^{\mathrm{ref}}(\bar{t}_{\kappa_{i,\mathrm{s}}})^{-1} \boldsymbol{P}_{\mathrm{ee}}^{\mathrm{ref}}(t),t\in[\hat{t}_{i,\mathrm{s}}-T_{\mathrm{sw}},\, \bar{t}_{\kappa_{i,\mathrm{s}}}),\\
    \boldsymbol{P}^{\mathrm{rel}}_{i,\mathrm{e}}(t) &= \boldsymbol{P}_{\mathrm{ee}}^{\mathrm{ref}}(\bar{t}_{\kappa_{i,\mathrm{e}}})^{-1} \boldsymbol{P}_{\mathrm{ee}}^{\mathrm{ref}}(t), t\in [\bar{t}_{\kappa_{i,\mathrm{e}}},\, \hat{t}_{i,\mathrm{e}}+T_{\mathrm{sw}}).
\end{aligned}
\end{equation*}
Then, we rigidly retarget these relative motions to the latest task trajectory during the pre- and post- manipulation phases
\begin{equation*}
\begin{aligned}
    &\bar{\boldsymbol{P}}_{\mathrm{t},i}(0)\,\boldsymbol{P}^{\mathrm{rel}}_{i,\mathrm{s}}(t),t\in[\hat{t}_{i,\mathrm{s}},\, \bar{t}_{\kappa_{i,\mathrm{s}}}),\\
    &\bar{\boldsymbol{P}}_{\mathrm{t},i}(T_{\mathcal{T},i})\,\boldsymbol{P}^{\mathrm{rel}}_{i,\mathrm{e}}(t), t\in [\bar{t}_{\kappa_{i,\mathrm{e}}},\, \hat{t}_{i,\mathrm{e}}),
\end{aligned}
\end{equation*}
which are the safe interaction motions for the latest task pose estimates.

\subsubsection{Smooth Blending via $\mathrm{SE}(3)$ Interpolation}
To avoid control discontinuities, we blend between the nominal and warped references over the entry/exit switching intervals using an $\mathrm{SE}(3)$ interpolation operator $\boldsymbol{f}_{\mathrm{interp}}(\boldsymbol{P}_A,\boldsymbol{P}_B,\alpha)$ (SLERP for rotation and linear interpolation for translation). The interpolation factors are scheduled as
\begin{equation*}
\begin{aligned}
\alpha_{\mathrm{in}}(t)&=\alpha_{\mathrm{poly}}\!\left(1-\frac{\hat{t}_{i,\mathrm{s}}-t}{T_{\mathrm{sw}}}\right),t\in[\hat{t}_{i,\mathrm{s}}-T_{\mathrm{sw}},\, \hat{t}_{i,\mathrm{s}}),\\
\alpha_{\mathrm{out}}(t)&=\alpha_{\mathrm{poly}}\!\left(1-\frac{t-\hat{t}_{i,\mathrm{e}}}{T_{\mathrm{sw}}}\right),t\in[\hat{t}_{i,\mathrm{e}},\, \hat{t}_{i,\mathrm{e}}+T_{\mathrm{sw}}),
\end{aligned}
\end{equation*}
where $\alpha_{\mathrm{poly}}(\tau)=6\tau^5-15\tau^4+10\tau^3$ for $\tau\in[0,1]$. Accordingly, we define
\begin{equation}
\begin{aligned}
f_{\mathrm{in},i}(t)=\boldsymbol{f}_{\mathrm{interp}}\!\Big(
\boldsymbol{P}_{\mathrm{ee}}^{\mathrm{ref}}(\bar{t}_{\kappa_{i,\mathrm{s}}}),\,
\bar{\boldsymbol{P}}_{\mathrm{t},i}(0),\,
\alpha_{\mathrm{in}}(t)
\Big),\\
t\in[\hat{t}_{i,\mathrm{s}}-T_{\mathrm{sw}},\, \hat{t}_{i,\mathrm{s}}),
\end{aligned}
\end{equation}
\begin{equation}
\begin{aligned}
f_{\mathrm{out},i}(t)=\boldsymbol{f}_{\mathrm{interp}}\!\Big(
\boldsymbol{P}_{\mathrm{ee}}^{\mathrm{ref}}(\bar{t}_{\kappa_{i,\mathrm{e}}}),\,
\bar{\boldsymbol{P}}_{\mathrm{t},i}(T_{\mathcal{T},i}),\,
\alpha_{\mathrm{out}}(t)
\Big),\\
t\in[\hat{t}_{i,\mathrm{e}},\, \hat{t}_{i,\mathrm{e}}+T_{\mathrm{sw}})
\end{aligned}
\end{equation}
which act as transition transforms that smoothly interpolate between the nominal reference pose and the latest task trajectory. In particular, they satisfy $f_{\mathrm{in},i}(\hat{t}_{i,\mathrm{s}}-T_{\mathrm{sw}})=\boldsymbol{P}_{\mathrm{ee}}^{\mathrm{ref}}(\bar{t}_{\kappa_{i,\mathrm{s}}})$ and $f_{\mathrm{in},i}(\hat{t}_{i,\mathrm{s}})=\bar{\boldsymbol{P}}_{\mathrm{t},i}(0)$, and similarly $f_{\mathrm{out},i}(\hat{t}_{i,\mathrm{e}})=\bar{\boldsymbol{P}}_{\mathrm{t},i}(T_{\mathcal{T},i})$ and $f_{\mathrm{out},i}(\hat{t}_{i,\mathrm{e}}+T_{\mathrm{sw}})=\boldsymbol{P}_{\mathrm{ee}}^{\mathrm{ref}}(\bar{t}_{\kappa_{i,\mathrm{e}}})$, thus avoiding pose and control discontinuities.

\subsubsection{Piecewise Construction of the Warped Trajectory}
We compose the warped trajectory (Fig.~\ref{fig:controller}B) by (i) blending into the warped trajectory, (ii) retargeted approach, (iii) task tracking, (iv) retargeted retreat, and (v) blending back to the nominal reference trajectory. The warped end-effector trajectory is constructed piecewise as
\begin{equation}
\begin{aligned}
&\boldsymbol{P}^{\mathrm{warp}}_{\mathrm{ee}}(t) =\\
&\begin{cases}
    f_{\mathrm{in},i}(t)\,\boldsymbol{P}^{\mathrm{rel}}_{i,\mathrm{s}}(t),
    & \exists\, i:\ t \in [\hat{t}_{i,\mathrm{s}}-T_{\mathrm{sw}},\, \hat{t}_{i,\mathrm{s}}), \\[4pt]
    \bar{\boldsymbol{P}}_{\mathrm{t},i}(0)\,\boldsymbol{P}^{\mathrm{rel}}_{i,\mathrm{s}}(t),
    & \exists\, i:\ t \in [\hat{t}_{i,\mathrm{s}},\, \bar{t}_{\kappa_{i,\mathrm{s}}}), \\[4pt]
    \bar{\boldsymbol{P}}_{\mathrm{t},i}(t - \bar{t}_{\kappa_{i,\mathrm{s}}}),
    & \exists\, i:\ t \in [\bar{t}_{\kappa_{i,\mathrm{s}}},\, \bar{t}_{\kappa_{i,\mathrm{e}}}), \\[4pt]
    \bar{\boldsymbol{P}}_{\mathrm{t},i}(T_{\mathcal{T},i})\,\boldsymbol{P}^{\mathrm{rel}}_{i,\mathrm{e}}(t),
    & \exists\, i:\ t \in [\bar{t}_{\kappa_{i,\mathrm{e}}},\, \hat{t}_{i,\mathrm{e}}), \\[4pt]
    f_{\mathrm{out},i}(t)\,\boldsymbol{P}^{\mathrm{rel}}_{i,\mathrm{e}}(t),
    & \exists\, i:\ t \in [\hat{t}_{i,\mathrm{e}},\, \hat{t}_{i,\mathrm{e}}+T_{\mathrm{sw}}), \\[4pt]
    \boldsymbol{P}_{\mathrm{ee}}^{\mathrm{ref}}(t),
    & t \notin \bigcup_{i} \mathcal{S}^{+}_{\mathrm{mani},i}.
\end{cases}
\end{aligned}
\label{eq:warping_operator}
\end{equation}
This construction preserves the nominal pre-/post-task motion up to a rigid transformation in the updated task frame, thereby maintaining the planned collision-clearance and interaction structure, while allowing the end-effector to follow the latest estimated task trajectory within $[\bar{t}_{\kappa_{i,\mathrm{s}}},\,\bar{t}_{\kappa_{i,\mathrm{e}}}]$.

\subsection{Phase-dependent Weight Transition}
\label{subsec:smoo_switch}
With the safe warped trajectory as the reference, the complementary switching weights $\sigma_{\mathrm{s}}$ and $\sigma_{\mathrm{ee}}$ are tuned to prioritize task error compensation within $\mathcal{S}_{\mathrm{mani},i}$ (i.e., $\sigma_{\mathrm{s}} \approx 0$ and $\sigma_{\mathrm{ee}} \approx 1$) and revert to whole-body trajectory tracking outside this phase (Fig.~\ref{fig:controller}C). 
To avoid abrupt changes in the MPC objective that could induce oscillatory behavior, the weight transition is synchronized with the trajectory warping over the same duration $T_{\mathrm{sw}}$. 
\begin{equation}
\begin{aligned}
\sigma_{\mathrm{s},i}(t) &=1 - (  f_{\log}(t- \hat{t}_{i,\mathrm{s}}, -T_{\mathrm{sw}}, 0) \\
&\qquad \qquad - f_{\log}(t- \hat{t}_{i,\mathrm{e}}, 0, T_{\mathrm{sw}}) ),\\
\sigma_{\mathrm{s}}(t) &= \prod_{i}^{}\sigma_{\mathrm{s},i}(t), \\
\sigma_{\mathrm{ee}}(t) &= 1-\sigma_{\mathrm{s}}(t).
\end{aligned}
\label{eq:smoo_switch}
\end{equation}
This ensures $\sigma_{\mathrm{ee}} \approx 1$ (and $\sigma_{\mathrm{s}} \approx 0$) within the task-critical phase $\mathcal{S}_{\mathrm{mani},i}$ to enforce task compliance, while reverting to global trajectory tracking outside.

\subsection{Cascaded MPC Integration}
\label{subsec:cascaded_mpc_integration}
Let $N$ denote the prediction horizon and $\Delta t$ the time step. Denote $t_{\mathrm{ref}}$ as the current time on the reference trajectory $\boldsymbol{x}(t)$. We define the prediction time steps as $t_k = t_{\mathrm{ref}} + k\Delta t$ for $k=0, \dots, N$. 
At each MPC update, the base MPC continues to track the global reference and avoid dynamic obstacles, while the manipulator MPC uses $\boldsymbol{P}^{\mathrm{warp}}_{\mathrm{ee},k}=\boldsymbol{P}^{\mathrm{warp}}_{\mathrm{ee}}(t_k)$ as the end-effector reference and $\sigma_{\mathrm{s}},\sigma_{\mathrm{ee}}$ to smoothly reweight its objective between tracking the whole-body and compensation for task-errors.

Specifically, conditioned on the predicted base trajectory $\bar{\boldsymbol{X}}_{\mathfrak{b}}^{*}$ computed by the base MPC, the manipulator MPC solves a receding-horizon Optimal Control Problem (OCP) whose stage cost is composed as
\begin{equation}
\label{eq:exec_cost_compact}
l_{\mathfrak{m},k}
=
\sigma_{\mathrm{s}}(t_k)\,l_{\mathrm{track},k}
+
\sigma_{\mathrm{ee}}(t_k)\,l_{\mathrm{task},k}
+
l_{\mathfrak{m},\mathrm{dy},k},
\end{equation}
where $l_{\mathrm{track},k}$ penalizes deviations from the global reference $\boldsymbol{x}(t_k)$, $l_{\mathfrak{m},\mathrm{dy},k}$ penalizes proximity to obstacles, and the task error compensation term is defined w.r.t.\ the warped reference:
\begin{equation}
\label{eq:task_cost_warp_ref}
l_{\mathrm{task},k}
=
\big\| [\boldsymbol{e}_{\mathrm{ee},k}]_{\boldsymbol{p}} \big\|_{\boldsymbol{Q}_{\mathrm{p}}}
-\omega_{R}\cdot \mathrm{tr} \big([\boldsymbol{e}_{\mathrm{ee},k}]_{\boldsymbol{R}}\big).
\end{equation}
Here $\boldsymbol{e}_{\mathrm{ee},k}:=\boldsymbol{P}_{\mathrm{ee},k}^{-1}\boldsymbol{P}^{\mathrm{warp}}_{\mathrm{ee},k}$ represents the pose error, $\boldsymbol{P}_{\mathrm{ee},k}=f_{\mathrm{FK,ee}}(\bar{\boldsymbol{x}}_{\mathfrak{b},k}^{*},\bar{\boldsymbol{x}}_{\mathfrak{m},k})$ is the predicted end-effector pose along the horizon. $\boldsymbol{Q}_{\mathrm{p}}\succeq 0$ and $\omega_{R}\geq0$ are position and orientation weight.
Because $\sigma_{\mathrm{s}}(t)$ and $\sigma_{\mathrm{ee}}(t)$ are synchronized with the warping window $\mathcal{S}^{+}_{\mathrm{mani},i}$, the objective transitions smoothly: outside task-critical phases, $\sigma_{\mathrm{s}}\approx 1$ emphasizes global tracking, while inside task-critical phases, $\sigma_{\mathrm{ee}}\approx 1$ emphasizes task error compensation relative to $\boldsymbol{P}^{\mathrm{warp}}_{\mathrm{ee},k}$. The complete OCP formulations, dynamic extensions, and obstacle modeling details are provided in \nameref{app:cascaded_mpc}. The resulting OCPs are solved online in a real-time receding-horizon manner using Acados~\cite{Verschueren2021} with HPIPM~\cite{frison2020hpipm} as the underlying solver.

% Both MPC modules operate in a receding-horizon manner, and only the first optimized inputs are executed at each cycle. 

In summary, the proposed controller integrates a safety-preserving end-effector trajectory warping and synchronized smooth weight switching into a cascaded MPC execution layer. 
This design enables online incorporation of task pose updates for reliable error compensation during task-critical phases, while retaining the planned safe-interaction structure and reverting to efficient global trajectory tracking outside these phases.

\section{Experiments}
In this section, we evaluate the proposed framework through a set of real-world and simulation experiments, designed to answer five questions: 
\begin{itemize}
  \item[1)] Can the framework efficiently execute long-horizon continuous mobile manipulation in constrained and dynamic real-world environments?
  \item[2)] Can it maintain reliable and efficient execution under substantial task-pose uncertainty in long-horizon continuous tasks?
  \item[3)] Can the framework extend to complex mobile manipulation tasks with diverse end-effector constraints that require continuous base--arm coordination?
  \item[4)] How does the proposed framework compare with the SOTA methods in terms of both efficiency and reliability?
  \item[5)] How does each component of the proposed framework contribute to the overall improvements in system performance?
\end{itemize}

The first three questions are addressed through a diverse set of real-world experiments in Sec.~\nameref{subsec:real_world}, including long-horizon mobile manipulation in constrained and dynamic environments, reliability validation under persistent task-pose uncertainty, and complex tasks with diverse end-effector constraints. Sec.~\nameref{subsec:benchmark} then compares the proposed framework with SOTA methods on simulation benchmarks, demonstrating superior performance in both efficiency and reliability. Finally, the ablation studies in Sec.~\nameref{subsec:ablation} examine the contribution of each key component to task-error compensation, as well as to safe and precise motion generation.

\subsection{Implementation Details}
As shown in Fig.~\ref{fig:problem_overview}(A), we adopt a differential-drive mobile manipulator, a simple yet sufficient platform to validate the problems considered. It comprises a two-wheel differential-drive base~\cite{agilextracerminiurl}, a 6-DOF manipulator~\cite{unitreez1url} with a two-finger gripper, and an onboard computer, integrating the mobility and manipulation capabilities required for implementing and validating our framework. For real-time perception, the robot relies solely on onboard sensors with no external devices. A base-mounted LiDAR–IMU unit~\cite{mid360url} enables the estimation of the mobile base state through LiDAR–inertial odometry~\cite{xu2022fast} and the detection of dynamic obstacles from LiDAR measurements. The joint states of the manipulator are obtained from internal joint encoders. An eye-in-hand RGB-D camera~\cite{femtobolturl} on the manipulator’s last link supports real-time estimation of target object states~\cite{liang2025dynamicpose}, critical for task-error compensation. 

In the real-world experiments, the prior environmental point cloud map is built using Fast-LIO~\cite{xu2022fast}, and online registration to the map is performed using small\_gicp~\cite{small_gicp}. Task trajectories are specified either from human-demonstrated target object motions or by manual design. For online planning, the size of the active task set is set to $N_{\mathcal{T}}=2$, meaning that at most two upcoming tasks are considered at each time step. For each updated active task set, the first planning call is allocated 2.5~s after the previous leading task is completed, and subsequent replanning is performed every 1.4~s until the current leading task is completed. The controller runs at 50~Hz to maintain responsiveness to real-world uncertainties. For computation, the real-world system runs entirely on the onboard computer equipped with an Intel i9-14900HX CPU and an NVIDIA RTX 4060 GPU. Simulation experiments are conducted on a computer equipped with an Intel i7-13700F CPU and an NVIDIA RTX 4090 GPU.

% sphere-set-approximation~\cite{wang2006variational}

\begin{figure*}[h]
\centering
\includegraphics[width=0.9\linewidth]{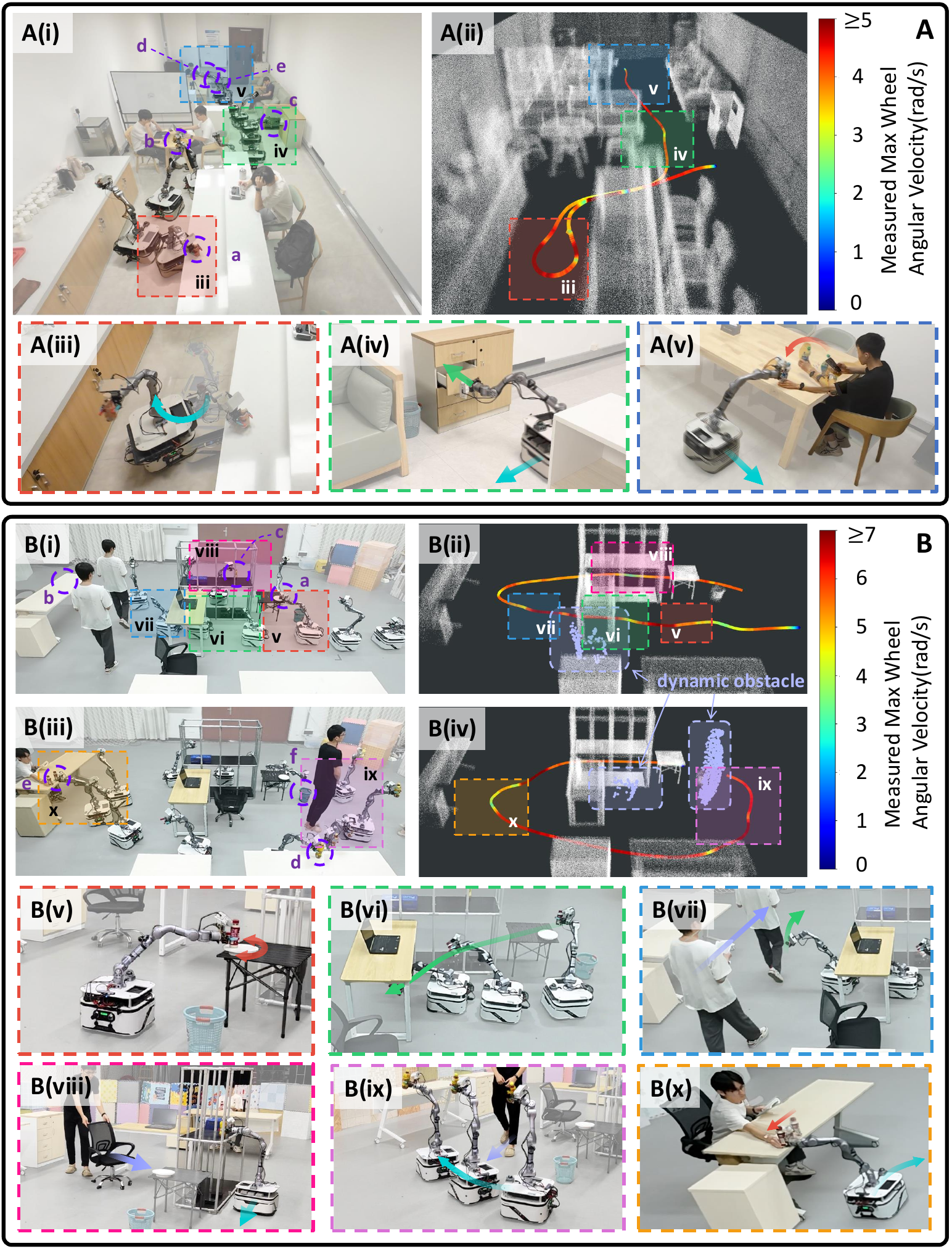}
\vspace{-0.1cm}
\caption{Mobile manipulation in constrained and dynamic environments.
(\textbf{A}) Constrained indoor environment. 
(i) Overview of the office lounge with the executed trajectory of the mobile manipulator and five task locations (a-e). 
(ii) Point cloud of the environment with the executed mobile base trajectory color-coded by the measured maximum wheel angular velocity of the two driven wheels. 
(iii)–(v) Representative motion segments: (iii) grasping a bottle while the mobile base executes a tight U-turn, (iv) closing an open drawer while moving toward the opposite side of the lounge, and (v) grasping a bottle randomly placed on the table. 
(\textbf{B}) Dynamic environment. 
(i) and (iii) Overviews of the environment and the mobile manipulator trajectories during two phases with six task locations (a-f).
(ii) and (iv) Point clouds of the environment and dynamic obstacles with mobile base trajectories, color-coded by the measured maximum wheel angular velocity during two phases. 
(v)–(x) Representative events: (v) grasping a bottle on a reciprocating turntable, (vi) lowering the manipulator to pass under a table, (vii) raising the manipulator to avoid a pedestrian, (viii) a person moving a chair to block the path under the table, (ix) the mobile base detouring to avoid a pedestrian, and (x) grasping a bottle that is displaced by a person just before grasping.
}
\label{fig:experiment_complex}
\vspace{-0.4cm}
\end{figure*}

\subsection{Real-world Experiments}
\label{subsec:real_world}
\subsubsection{Constrained and Dynamic Real-World Environments}
\label{subsubsec:exp_real_complex}
To answer the first question, two long-horizon real-world experiments were conducted in a constrained office lounge and a dynamic environment, validating the framework’s ability to generate continuous whole-body motion across tightly arranged tasks, manipulation-on-the-move performance, and robustness to dynamic disturbances. Visual experimental results are provided in Extension~2.

\paragraph{Scenario 1: Constrained indoor environment.}
In this scenario (Fig.~\ref{fig:head}A and Fig.~\ref{fig:experiment_complex}A), set in an office lounge, the MM coordinated navigation and manipulation and sequentially completed five tasks within a confined space. It started by traversing a narrow corridor before approaching a target coffee bottle. To ensure stable perception, the MM actively adjusted the camera viewpoint via our time-assured active perception strategy. In a particularly demanding maneuver, the robot simultaneously grasped the bottle while its mobile base executed a tight U-turn (Fig.~\ref{fig:experiment_complex}A(iii)), demonstrating reliable manipulation-on-the-move in a constrained space. After delivering the coffee to a nearby table, the robot smoothly closed an open drawer along its path (Fig.~\ref{fig:experiment_complex}A(iv)) before proceeding to grasp a randomly placed bottle (Fig.~\ref{fig:experiment_complex}A(v)) on the opposite side of the lounge, then seamlessly discarded the bottle in a nearby trash can. The entire sequence was executed as a single continuous motion without unnecessary stops.

\paragraph{Scenario 2: Dynamic environment.}
The second scenario, consisting of six consecutive tasks (Fig.~\ref{fig:experiment_complex}B), evaluated the system's performance under dynamic disturbances, including moving targets and obstacles. The sequence began with grasping a bottle from a reciprocating turntable (Fig.~\ref{fig:experiment_complex}B(v)), which required continuous motion adjustments enabled by our real-time task-error compensation controller. The robot then traversed a blocked aisle, navigating beneath a table (Fig.~\ref{fig:experiment_complex}B(vi)) through coordinated base–arm motion from whole-body trajectory generation. Next, before the placing task, a pedestrian suddenly crossed the robot’s path (Fig.~\ref{fig:experiment_complex}B(vii)). The controller reacted in real-time to generate collision-avoidance motions, demonstrating robustness to unexpected obstacles. Then, after grasping a bottle from the shelf, the under-table passage became obstructed by a chair (Fig.~\ref{fig:experiment_complex}B(viii)). The robot replanned an efficient alternative route through the main aisle, adapting to the updated environment. Before placing the bottle, a pedestrian moved randomly (Fig.~\ref{fig:experiment_complex}B(ix)). The system replanned a trajectory online again to maintain clearance. Finally, when the target object was displaced before grasping (Fig.~\ref{fig:experiment_complex}B(x)), task-error compensation adapted the motion to the latest target pose, enabling a successful grasp. The experiment ended with the bottle being disposed of successfully.

Collectively, these two long-horizon experiments validate that our framework enables the efficient and reliable execution of tightly arranged mobile manipulation tasks in both constrained and dynamic environments. Throughout the complete sequence of tasks, the robot maintained a continuous base movement (Fig.~\ref{fig:experiment_complex}A(ii), B(ii), and B(iv), where wheel angular velocities were calculated by Eq.~\eqref{eq:base_wheel_omega} via the measured linear and angular velocities of the mobile base), while generating smooth whole-body motions that integrate navigation and manipulation with safe interaction motions, maintaining object stability during approach, manipulation, and retraction to support reliable task completion. 

\subsubsection{Reliability Validation in Long-Horizon Continuous Tasks}
\label{subsubsec:exp_real_reliability}

\begin{figure*}[t]
\centering
\includegraphics[width=0.8\linewidth]{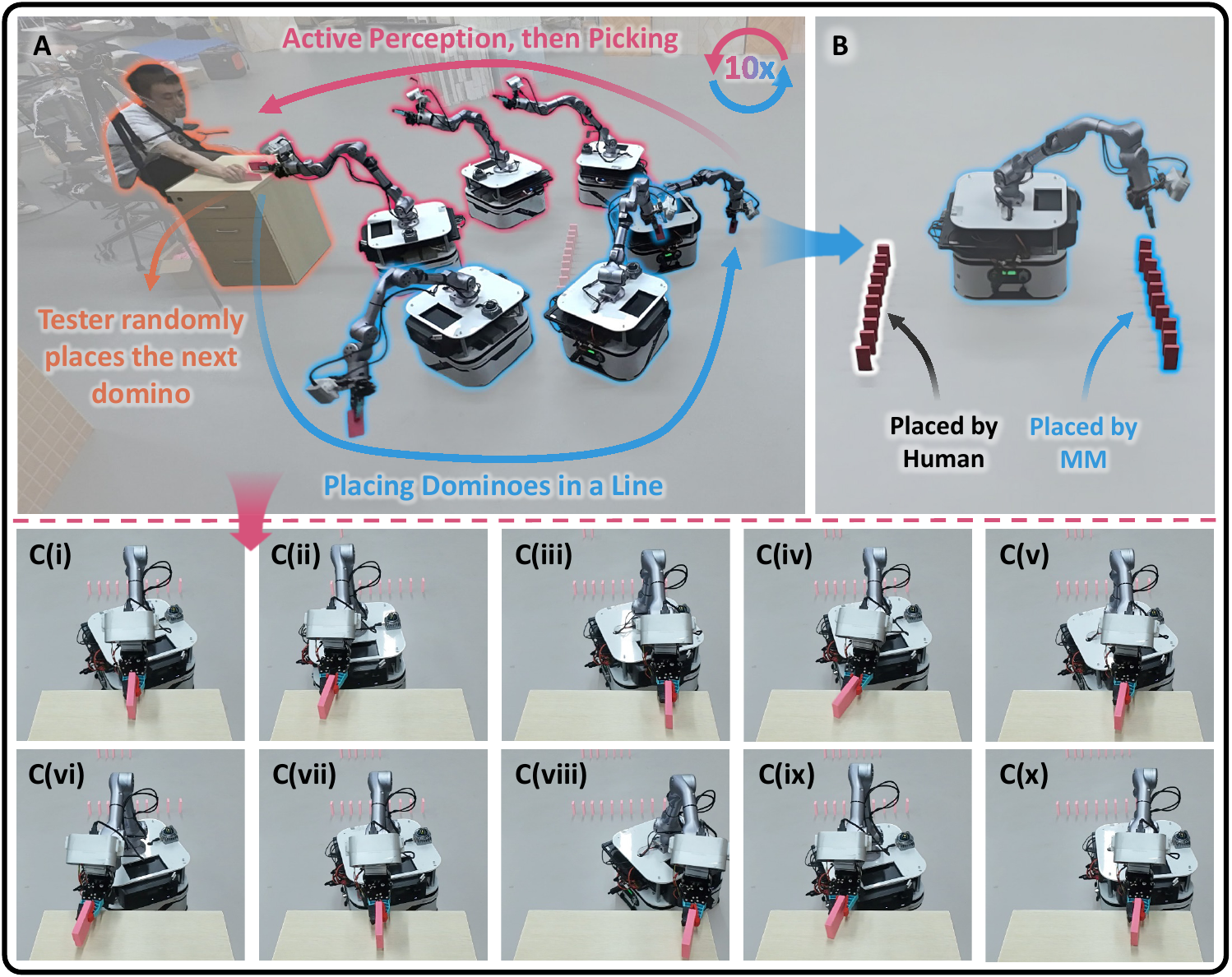}
\vspace{-0.1cm}
\caption{Long-horizon tasks under task pose uncertainty.
(\textbf{A}) Tasks overview: the MM repeatedly ($10\times$) employs active perception to estimate domino poses, then grasps and arranges them into a straight line.
(\textbf{B}) Final placement comparison: the line of dominoes arranged by a human (left) versus that produced by the MM (right).
(\textbf{C}) Snapshots captured at the grasping instant for the ten consecutive grasping tasks (i)--(x).
}
\label{fig:experiment_reliability_task}
\vspace{-0.4cm}
\end{figure*}

% \begin{figure*}[t]
% \centering
% \includegraphics[width=0.9\linewidth]{fig/experiment_reliability.pdf}
% \vspace{-0.1cm}
% \caption{Long-horizon tasks under pose uncertainty.
% (\textbf{A}) Tasks overview: the MM repeatedly ($10\times$) employs active perception to estimate domino poses, then grasps and arranges them into a straight line.
% (\textbf{B}) Final placement comparison: the line of dominoes arranged by a human (left) versus that produced by the MM (right).
% (\textbf{C}) Snapshots captured at the grasping instant for the first six of ten consecutive grasping tasks (i)--(vi).
% (\textbf{D}) Top-down comparison of the coarse initial pose and the final refined estimate (last perception estimate before grasping) of the domino for the ten grasping tasks.
% (\textbf{E}) Time evolution of pose estimation deviation (mean and min-max range over the ten trials) relative to the final refined estimate.
% (\textbf{F}) Comparison of computation time and trajectory duration.
% (\textbf{G}) Distribution of pose errors between the planned and target manipulation poses.
% (\textbf{H}) Position and orientation errors with respect to the final refined estimate, including trajectory error (deviation of the planned trajectory) and post-compensation error (final deviation after task-error compensation).
% }
% \label{fig:experiment_reliability}
% \vspace{-0.4cm}
% \end{figure*}

To assess the reliability of our framework in real-world long-horizon tasks, we designed a challenging experiment in which a mobile manipulator sequentially picked up 10 dominoes from a table (Fig.~\ref{fig:experiment_reliability_task}A) and placed them in a straight line on the opposite side (Fig.~\ref{fig:experiment_reliability_task}B), totaling 20 tasks. As illustrated in Fig.~\ref{fig:experiment_reliability_task}A, only coarse initial poses of dominoes were provided across trials, and the system had to rely on its eye-in-hand camera to estimate the actual poses and adjust its motion during execution. After each placement, a tester deliberately introduced target pose uncertainty by randomly repositioning the next domino. Throughout the experiment, the mobile base and the manipulator of the MM maintained continuous coordinated motion, and the system completed all 20 tasks, producing a straight domino line comparable to that arranged by a human (Fig.~\ref{fig:experiment_reliability_task}B, and Extension~3).

\begin{figure}[h]
\centering
\includegraphics[width=0.9\linewidth]{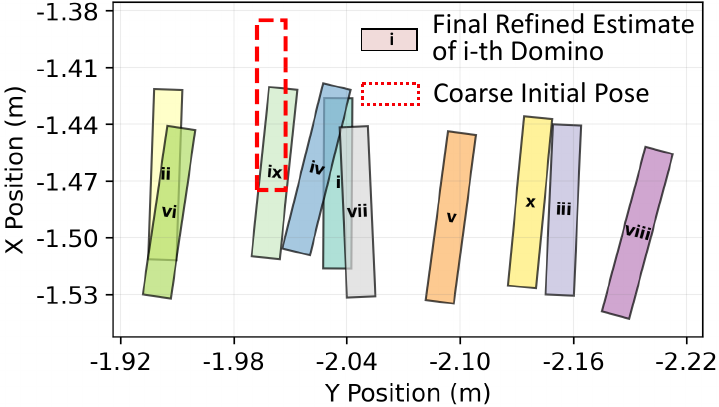}
\vspace{-0.1cm}
\caption{Top-down comparison of the coarse initial pose and the final refined estimate (last perception estimate before grasping) of the domino for the ten grasping tasks.
}
\label{fig:experiment_reliability_final_pose}
\vspace{-0.2cm}
\end{figure}

\begin{figure}[h]
\centering
\includegraphics[width=0.9\linewidth]{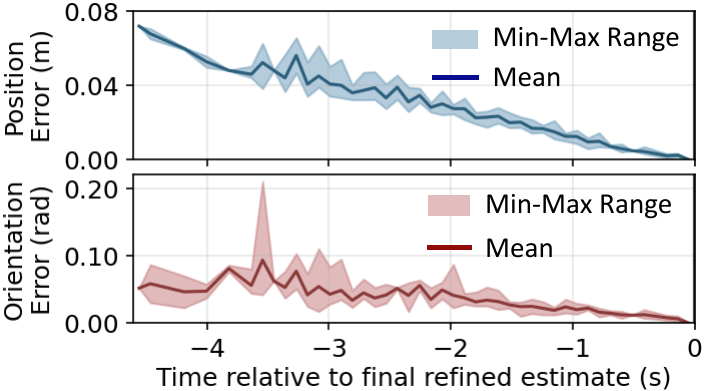}
\vspace{-0.1cm}
\caption{Time evolution of pose estimation deviation (mean and min-max range over the ten trials) relative to the final refined estimate.
}
\label{fig:experiment_reliability_evolution}
\vspace{-0.2cm}
\end{figure}

In this experiment, there are two key challenges: target pose uncertainty from randomized domino placements and estimation deviations. Fig.~\ref{fig:experiment_reliability_task}C shows snapshots of the grasping moment from the ten picking tasks, with the MM grasping the dominoes randomly positioned on the table. Fig.~\ref{fig:experiment_reliability_final_pose} compares the coarse initial pose with the final refined estimate (last domino pose estimate before grasping) for each picking task, revealing discrepancies of up to 0.2~m. Here, because external ground truth is unavailable, we use the final refined estimate as a reference. Fig.~\ref{fig:experiment_reliability_evolution} then summarizes the deviation of earlier estimates from this reference over time across ten picking tasks. Specifically, the deviation at each time step is computed as the difference between the current estimate and the final refined estimate. As shown in Fig.~\ref{fig:experiment_reliability_evolution}, the persistence of these deviations highlights the necessity for the system to adjust its motion in real-time using the latest pose estimate to ensure a successful grasp.

\begin{figure}[h]
\centering
\includegraphics[width=0.9\linewidth]{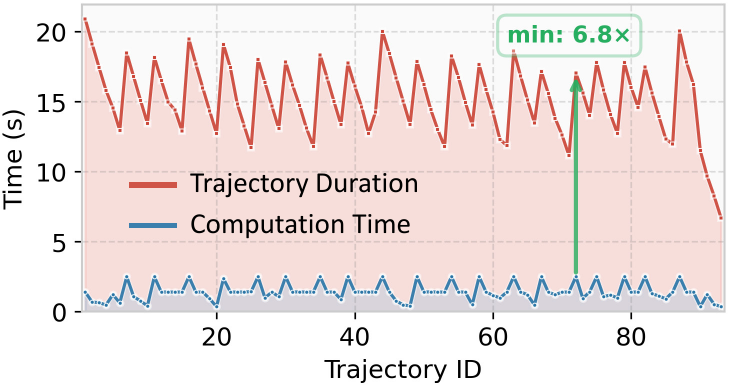}
\vspace{-0.1cm}
\caption{Comparison of computation time and trajectory duration.
}
\label{fig:experiment_reliability_computation}
\vspace{-0.4cm}
\end{figure}

\begin{figure}[h]
\centering
\includegraphics[width=0.9\linewidth]{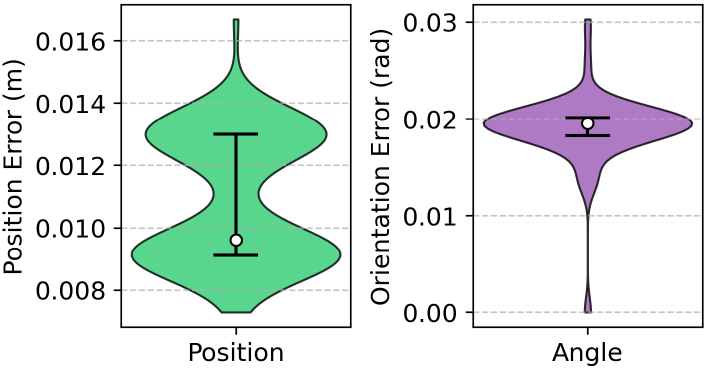}
\vspace{-0.1cm}
\caption{Distribution of pose errors between the planned and target manipulation poses.
}
\label{fig:experiment_reliability_precision}
\vspace{-0.2cm}
\end{figure}

We evaluate whether our planner can effectively replan trajectories online to retarget the end-effector to the latest pose estimate. This capability is crucial for efficient and reliable mobile manipulation. It allows the robot to maintain continuous motion by ensuring that each new trajectory is computed before the current trajectory completes execution. It also preserves reliability by keeping the replanned manipulation pose sufficiently accurate to meet the desired end-effector pose. To quantify this capability, we use three metrics: (i) the trajectory duration, (ii) the computation time (Fig.~\ref{fig:experiment_reliability_computation}), and (iii) the position and orientation error between the planned and target manipulation pose (Fig.~\ref{fig:experiment_reliability_precision}). Here, the target manipulation pose is defined as the desired end-effector pose derived from the most recent pose estimate available at the planning instance. As shown in Fig.~\ref{fig:experiment_reliability_computation}, the computation time remained consistently below the corresponding trajectory duration across replanning cycles. Even in the most demanding cycle, the planned trajectory lasted about 6.8 times longer than the time needed to compute it, leaving sufficient time to generate the next trajectory before execution finished and avoiding pauses for replanning. As summarized in Fig.~\ref{fig:experiment_reliability_precision}, the planner effectively converged to the target manipulation pose, yielding a median position error of 0.0096~m (max 0.0167~m) and an orientation error of 0.0195~rad (max 0.03~rad). These results confirm that the planner can replan online to align the end-effector with the latest pose estimate during continuous execution.

\begin{figure}[h]
\centering
\includegraphics[width=0.9\linewidth]{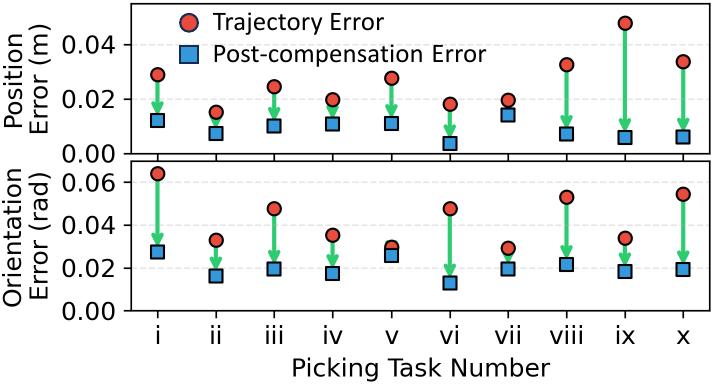}
\vspace{-0.1cm}
\caption{Position and orientation errors with respect to the final refined estimate, including trajectory error (deviation of the planned trajectory) and post-compensation error (final deviation after task-error compensation).
}
\label{fig:experiment_reliability_compensation}
\vspace{-0.4cm}
\end{figure}

Finally, we evaluate the real-time task-error compensation ability of our system. Despite the satisfactory precision of the planner (Fig.~\ref{fig:experiment_reliability_precision}), whole-body replanning is computationally intensive and executed at a much lower rate than the perception module (above 5 Hz). Consequently, the final refined estimate may become available after the last replanning update (Fig.~\ref{fig:experiment_reliability_evolution}). This latency results in a discrepancy between the planned manipulation pose and the target grasp pose derived from the final refined estimate (labeled as Trajectory Error in Fig.~\ref{fig:experiment_reliability_compensation}). To mitigate this, we employ a safe-warping-based phase-dependent controller (50~Hz) that smoothly switches between global trajectory tracking and task-error compensation during task-critical phases while maintaining safe interaction. As shown in Fig.~\ref{fig:experiment_reliability_compensation} (Post-compensation Error), this controller significantly reduces the final execution error to less than 0.014~m (median 0.0086~m) in position and 0.027~rad (median 0.0193~rad) in orientation.

Overall, these results demonstrate that our proposed unified framework enables the reliable execution of long-horizon manipulation sequences under perception-induced domino pose uncertainty. By synergizing reliability-aware long-horizon planning with a high-frequency safe-warping-based phase-dependent controller, our system sustains precise manipulation across repeated trials even with large (0.2~m) discrepancies between coarse and final refined domino poses.

\subsubsection{Base-Arm Coordination for Complex Tasks with Diverse End-effector Constraints}
\label{subsubsec:exp_real_extensibility}
\begin{figure*}[t]
\centering
\includegraphics[width=0.9\linewidth]{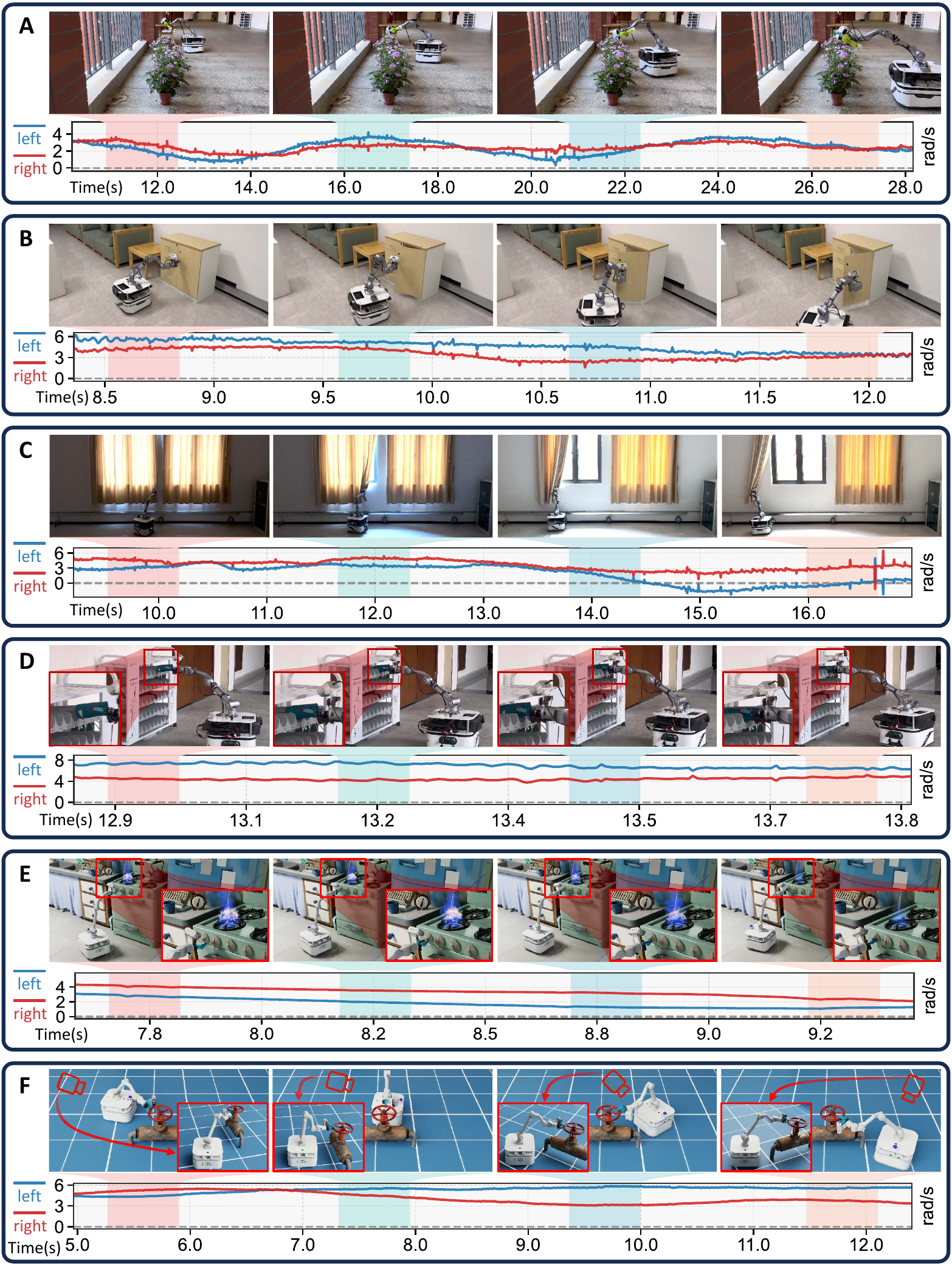}
\vspace{-0.1cm}
\caption{Base-arm coordination for complex tasks.
For each of six mobile manipulation tasks, the top row shows representative snapshots and the bottom row plots the left and right wheel angular velocities over time. In each plot, four shaded time intervals indicate the trajectory segments corresponding to the four snapshots above and highlight how the wheel angular velocities evolve across representative phases of execution. The tasks include (\textbf{A}) watering a row of flowers, (\textbf{B}) opening a cabinet door along a circular arc, (\textbf{C}) pulling a curtain open along a straight line, (\textbf{D}) inserting an umbrella into an umbrella stand, (\textbf{E}) turning off a stove knob, and (\textbf{F}) turning a large valve while moving around it. Across all tasks, the wheel velocities remain non-zero and are automatically modulated to accommodate the end-effector–constrained phases while maintaining continuous whole-body motion.
}
\label{fig:experiment_extensibility}
\vspace{-0.4cm}
\end{figure*}

To demonstrate the extensibility of our framework to complex tasks beyond pick-and-place, specifically those with diverse end-effector trajectory constraints, we conducted several experiments (Fig.~\ref{fig:experiment_extensibility}). Successful completion of these tasks requires the end-effector to follow task-imposed geometric constraints while the mobile base and manipulator coordinate continuously. Importantly, in contact- and constraint-rich behaviors (e.g., door opening and valve turning), task success requires more than just end-effector trajectory tracking; it often involves complex whole-body constraints. Specifically, the mobile base must retract to ensure smooth door opening or maintain an appropriate distance during valve turning. Through these experiments, we demonstrate that our method enables reliable task completion via whole-body coordination, which is more challenging than tracking the end-effector trajectory in open free space. Visual results are provided in Extension~4.

For flower watering (Fig.~\ref{fig:experiment_extensibility}A), the end-effector followed a complex 6D motion: the robot first grasped the watering can while moving swiftly and kept it level when approaching the flowers. It then smoothly tilted and translated the can along a cubic B-spline, maintaining a consistent height and passing directly above each flower to water the row, before finally returning to a level pose. Opening a cabinet (Fig.~\ref{fig:experiment_extensibility}B) and pulling a curtain (Fig.~\ref{fig:experiment_extensibility}C) required long geometric paths—a circular arc and a straight line, respectively—whose spatial extent exceeded the standalone workspace of the manipulator. In both cases, the mobile base had to continuously reposition to keep the end-effector within a feasible manipulation region while the manipulator performed the constrained motion.

Inserting an umbrella into a stand hole (Fig.~\ref{fig:head}B and Fig.~\ref{fig:experiment_extensibility}D) and turning off a stove knob (Fig.~\ref{fig:experiment_extensibility}E) instead involved constrained phases of varying durations in which the end-effector had to remain within a small region for a finite time. For umbrella insertion, the constrained phase lasted about 0.8 s; nevertheless, our method enabled precise insertion without noticeable slowdown of the mobile base. In contrast, turning off the stove required an in-place rotational motion of the end-effector, confining the end-effector to an even smaller region for a longer duration (about 1.6 s). In this case, the mobile base autonomously slowed down to ensure the successful completion of the task. Finally, turning a large valve (Fig.~\ref{fig:experiment_extensibility}F) required the mobile base to move around the valve while the end-effector executed circular motion around the valve axis. At the beginning of this task, the manipulator retracted to allow the mobile base to move along a smaller radius around the valve, thereby shortening its path and improving efficiency; during the turning phase, it extended to maintain clearance between the mobile base and the valve pedestal, illustrating efficient and safe whole-body coordination of base and manipulator under task-imposed end-effector constraints.

Attributed to our whole-body long-horizon planning, the proposed framework successfully extends to complex tasks with diverse end-effector constraints. The flower-watering, cabinet-opening, and curtain-pulling tasks demonstrate our framework’s ability to execute long and diverse constrained end-effector motions by continuously repositioning the base. In contrast, the umbrella-insertion, stove-knob, and valve-turning tasks highlight its ability to autonomously coordinate the mobile base and manipulator to remain efficient while ensuring task completion. Across all six tasks, the system maintains continuous whole-body motion, avoiding unnecessary stops during both the approach and the constrained phases. As shown in Fig.~\ref{fig:experiment_extensibility}, the mobile base wheel velocities remain non-zero throughout execution and are automatically modulated in response to the evolving task constraints.

\subsection{Evaluation of Efficiency and Reliability}
\label{subsec:benchmark}

\subsubsection{Simulated Scenarios and Experimental Setup}

\begin{figure}[h]
\centering
\includegraphics[width=0.99\linewidth]{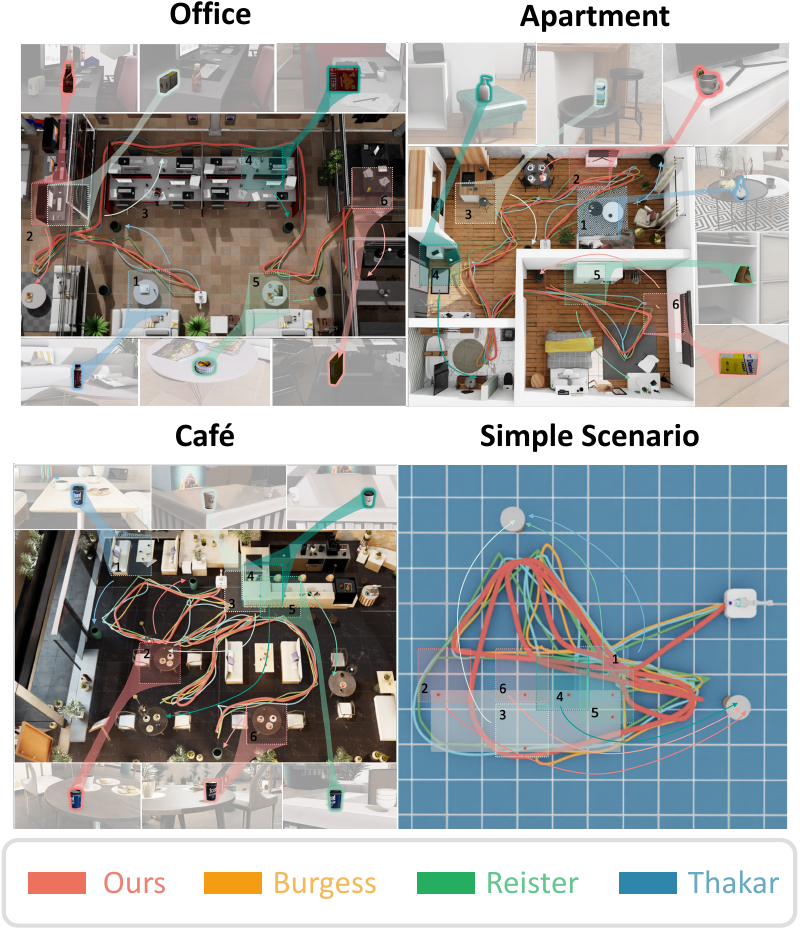}
\vspace{-0.2cm}
\caption{Overview of four benchmarking scenarios (office, apartment, café, and simple scenario) and mobile base trajectory for our method versus baselines. Each scenario comprises six pick-and-place/drop missions with object displacements of 0, 0.05, or 0.10\,m.
}
\label{fig:benchmark_scenarios}
\vspace{-0.2cm}
\end{figure}

To evaluate the efficiency and reliability of the proposed method, we conducted benchmark experiments in Isaac Sim~\cite{isaacsimurl} across four scenarios of increasing complexity (Fig.~\ref{fig:benchmark_scenarios}): an office ($8\,\mathrm{m}\times15\,\mathrm{m}$), an apartment ($9\,\mathrm{m}\times9\,\mathrm{m}$), a café ($9\,\mathrm{m}\times11\,\mathrm{m}$), and a simple scenario (Simple) ($6\,\mathrm{m}\times5\,\mathrm{m}$) adapted from~\cite{burgess2024reactive}. In each scenario, the mobile manipulator was required to execute six \emph{pick-and-place} or \emph{pick-and-drop} missions. Each mission comprised a pick task followed by a place or drop task. These missions were designed to require coordinated whole-body motion and safe interaction with both the manipulated objects and the surrounding environment, while also enabling large-scale quantitative evaluation. In all missions, the target objects were initially placed on support surfaces of different heights (Fig.~\ref{fig:benchmark_scenarios}), and the robot was required to pick them up and transport them to a target region for placement or dropping. For each mission, the coarse initial pose of the object to be picked, as well as the target pose for placement or dropping, was specified a priori.

To evaluate robustness to discrepancies between the coarse and true object poses, we introduced perturbations to the ground-truth poses of the objects to be picked. Specifically, for each mission, the coarse planar position $(x,y)$ of the target object was generated by perturbing its ground-truth position $(x_{\mathrm{gt}}, y_{\mathrm{gt}})$ with a displacement of fixed magnitude $d_\sigma \in \{0, 0.05, 0.10\}\,\mathrm{m}$ in a random direction:
\begin{equation*}
(x, y) = (x_{\mathrm{gt}}, y_{\mathrm{gt}}) + d_\sigma[\cos\theta, \sin\theta],
\end{equation*}
where $\theta$ was sampled uniformly from $[0, 2\pi)$, and samples that resulted in collision were discarded and resampled. These displacement magnitudes were chosen such that the resulting offsets were large enough to require noticeable motion correction during picking, while still ensuring that the true object remained within the camera field of view when the onboard sensor was oriented toward the coarse initial pose. During execution, the algorithms had no access to ground-truth object poses. Instead, in all experiments, the position and orientation of the target objects were continuously estimated from simulated RGB-D observations using the same perception module~\cite{liang2025dynamicpose}.

We compared our method against three state-of-the-art baselines: Burgess et al.’s reactive base controller for on-the-move manipulation \cite{burgess2024reactive}, Reister et al.’s stop-then-manipulate method \cite{reister2022combining}, and Thakar et al.’s planning-based method \cite{thakar2018towards}. Each method was evaluated 10 times in each scenario for each noise setting (a total of $4\times3\times10\times6=720$ missions per method). The experiment terminated at any of the three conditions: (i) the algorithm alternately sent six cycles of gripper closing and opening commands, which the algorithm considered as the completion of six missions; (ii) a collision occurred between the MM and the environment, or a self-collision occurred to the MM; (iii) the robot was stuck, i.e., the distance the mobile base moved within 30 seconds was less than 0.15~m.

\begin{table*}[t]
\footnotesize\sf\centering
\caption{Simulation benchmark results across four scenarios under different levels of task-pose displacement $d_\sigma$. For each scenario, we report the mission success rate, scenario success weighted by cycle time (SSCT), and average base linear velocity (Avg Vel) over 60 missions; for SSCT, both the mean and standard deviation (Std) are provided.}
\label{tab:sim_benchmark}
\setlength{\tabcolsep}{5.2pt}
\renewcommand{\arraystretch}{1.1}
\begin{tabular}{c|c|ccc|cc|cc|cc|c}
\toprule
\multirow{3}{*}{\textbf{Scenarios}} & \multirow{3}{*}{\textbf{Methods}}
& \multicolumn{3}{c|}{\textbf{Mission Success Rate (\%)}$\,\uparrow$}
& \multicolumn{6}{c|}{\textbf{SSCT}$\,\uparrow$}
& \multirow{3}{*}{\textbf{Avg Vel (m/s)}$\,\uparrow$} \\
& 
& \multirow{2}{*}{$d_\sigma=0.0$}
& \multirow{2}{*}{$d_\sigma=0.05$}
& \multirow{2}{*}{$d_\sigma=0.1$}
& \multicolumn{2}{c|}{$d_\sigma=0.0$}
& \multicolumn{2}{c|}{$d_\sigma=0.05$}
& \multicolumn{2}{c|}{$d_\sigma=0.1$}
& \\
& & & &
& \textbf{Mean} & \textbf{Std}
& \textbf{Mean} & \textbf{Std}
& \textbf{Mean} & \textbf{Std}
& \\
\midrule

\multirow{4}{*}{Office}
& Thakar  & 30.00 & 15.00 & 3.33
    & 0.0852 & 0.0561 
    & 0.0374 & 0.0385 
    & 0.0171 & 0.0343
& 0.1335 \\
& Reister & 93.33 & 68.33 & 65.00
    & 0.2628 & 0.0151 
    & 0.1905 & 0.0801 
    & 0.1979 & 0.0802
& 0.1118 \\
& Burgess & 26.67 & 18.33 & 21.67
    & 0.1674 & 0.0712 
    & 0.1001 & 0.0845 
    & 0.1217 & 0.0859
& 0.2362 \\
\rowcolor{gray!15}
\cellcolor{white} & Ours
& \textbf{100.0} & \textbf{100.0} & \textbf{100.0}
    & \textbf{0.6125} & 0.0168 
    & \textbf{0.6143} & 0.0118 
    & \textbf{0.6230} & 0.0193
& \textbf{0.2462} \\
\midrule

\multirow{4}{*}{Apartment}
& Thakar  & 0.00 & 3.33 & 0.00
    & 0.0000 & 0.0000 
    & 0.0097 & 0.0200 
    & 0.0000 & 0.0000
& 0.1347 \\
& Reister & 50.00 & 25.00 & 38.33
    & 0.1150 & 0.0000 
    & 0.0638 & 0.0277 
    & 0.1030 & 0.0402
& 0.1088 \\
& Burgess & 0.00 & 8.33 & 13.33
    & 0.0000 & 0.0000 
    & 0.0473 & 0.0876 
    & 0.0767 & 0.0907
& 0.1846 \\
\rowcolor{gray!15}
\cellcolor{white} & Ours
& \textbf{100.0} & \textbf{100.0} & \textbf{98.33}
    & \textbf{0.7006} & 0.0154 
    & \textbf{0.6919} & 0.0186 
    & \textbf{0.6754} & 0.0401
& \textbf{0.2459} \\
\midrule

\multirow{4}{*}{Café}
& Thakar  & 56.67 & 8.33 & 0.00
    & 0.1957 & 0.0758 
    & 0.0274 & 0.0475 
    & 0.0000 & 0.0000
& 0.1287 \\
& Reister & 100.00 & 100.00 & 100.00
    & 0.3062 & 0.0095 
    & 0.3076 & 0.0024 
    & 0.3087 & 0.0017
& 0.1145 \\
& Burgess & 46.67 & 51.67 & 48.33
    & 0.3647 & 0.0699 
    & 0.3995 & 0.1069 
    & 0.3856 & 0.1262
& \textbf{0.2465} \\
\rowcolor{gray!15}
\cellcolor{white} & Ours
& \textbf{100.0} & \textbf{100.0} & \textbf{100.0}
    & \textbf{0.7436} & 0.0213 
    & \textbf{0.7397} & 0.0150 
    & \textbf{0.7304} & 0.0182
& 0.2459 \\
\midrule

\multirow{4}{*}{Simple}
& Thakar  & 86.67 & 15.00 & 0.00
    & 0.2396 & 0.0655 
    & 0.0376 & 0.0243 
    & 0.0000 & 0.0000
& 0.1172 \\
& Reister & 100.0 & 96.67 & 41.67
    & 0.2206 & 0.0001 
    & 0.2115 & 0.0142 
    & 0.0943 & 0.0733
& 0.1047 \\
& Burgess & 90.00 & 76.67 & 88.33
    & 0.4701 & 0.0595 
    & 0.3978 & 0.0739 
    & 0.4483 & 0.0662
& 0.2186 \\
\rowcolor{gray!15}
\cellcolor{white} & Ours
& \textbf{100.0} & \textbf{100.0} & \textbf{100.0}
    & \textbf{0.5749} & 0.0174 
    & \textbf{0.5724} & 0.0160 
    & \textbf{0.5783} & 0.0096
& \textbf{0.2421} \\
\bottomrule
\end{tabular}
\vspace{-0.4cm}
\end{table*}

\subsubsection{Evaluation Metrics}
\paragraph{Mission Success.}
We categorize each trial into two mission types: (i) pick-and-place, where success is defined as completing a full pick-and-place cycle—i.e., the robot (a) successfully grasps the object (verified by the gripper not being fully closed), (b) transports it to the designated target, and (c) releases it, with the released object resting within the target region and meeting the pose tolerance (position error below $0.15\,\mathrm{m}$ and tilt (roll/pitch) less than $45^\circ$; and (ii) pick-and-drop, where success is defined as successfully releasing the object and the final planar position of the object lying within $0.15\,\mathrm{m}$ of the target location inside the specified target region.

\paragraph{Mission Cycle Time.}
Mission efficiency is quantified by the mission cycle time, defined as the elapsed time between consecutive task milestones. We designate the timestamp of a ``gripper-open'' event as the completion marker for each mission, since opening the gripper to release the object signifies the end of the mission. Consequently, for a sequence of $N$ missions, the cycle time for any subsequent mission is the interval between two consecutive gripper-open events. For the first mission, where there is no preceding gripper action, the cycle time is measured from the global execution trigger (issued to initiate the entire $N$ missions) to the first gripper-open event. Lower cycle time indicates higher execution efficiency.

\paragraph{Mission Success weighted by Cycle Time (MSCT).}
Mission performance is quantified by the Mission Success-weighted Cycle Time (MSCT) \cite{yokoyama2021success}, which jointly evaluates mission success and efficiency. For each mission, MSCT is computed as:
\begin{equation}
    \text{MSCT} = s \cdot \frac{T}{\max(C, T)}
\end{equation}
where $s \in \{0, 1\}$ is the success indicator, $T$ is the idealized lower-bound time for the mission, and $C$ is the actual mission cycle time. The idealized time $T$ is calculated as the time required for the mobile base to traverse the shortest collision-free path for the mission at maximum speed, accounting for the maximum reach of the manipulator. The MSCT metric ranges from 0 to 1, where $\text{MSCT} = 0$ indicates mission failure, and $0 < \text{MSCT} \leq 1$ reflects successful completion scaled by execution efficiency. Higher values indicate better performance.

\paragraph{Scenario Success weighted by Cycle Time (SSCT).}
For a scenario consisting of $N$ missions, we report the Scenario Success weighted by Cycle Time (SSCT) as the mean MSCT over all missions in that scenario:
\begin{equation}
\text{SSCT}=\frac{1}{N}\sum_{i=1}^{N}\text{MSCT}_i,
% \qquad
% \text{MSCT}_i = s_i\cdot \frac{T_i}{\max(C_i,T_i)}.
\end{equation}
SSCT also ranges from 0 to 1. Higher values indicate better overall scenario performance, jointly reflecting the scenario-level success rate and time efficiency across missions.

\subsubsection{Benchmark Results and Analysis}

We evaluated reliability using the mission success rate in Tab.~\ref{tab:sim_benchmark}, defined as the fraction of missions that completed the prescribed \emph{pick-and-place} and \emph{pick-and-drop} missions across all benchmark scenarios and perturbation settings. Over 720 missions spanning all object displacements, our method achieved a 99.86\% success rate, indicating consistently robust execution even under substantial target pose discrepancies. Burgess et al.~\cite{burgess2024reactive} achieved a success rate of 40.83\%, with failures primarily attributable to four systemic limitations. First, the base placement is selected from a predefined candidate set solely based on transition efficiency, without considering the feasibility of subsequent manipulation. Consequently, the chosen placement may render the target unreachable or require unsafe manipulation motions in complex 3D environments. Second, in terms of motion generation, the decoupled control of the mobile base and manipulator lacks unified whole-body collision awareness. Specifically, the base navigates without considering the manipulator's safety, leading to collisions with obstacles in cluttered environments (e.g., Office and Apartment). Third, the base motion generator \cite{missura2022fast} in Burgess et al.~\cite{burgess2024reactive} is prone to getting stuck near obstacles due to the limited computational time. This issue is exacerbated by the manipulator's limited workspace, which necessitates close-proximity navigation to obstacles (e.g., tables) for reachability. Finally, the method does not explicitly account for end-effector--object interaction during the approach and retraction phase. This lack of sensitivity often leads to unintended contact, knocking the object down before grasping or after placing. Reister et al.~\cite{reister2022combining} achieved a success rate of 73.19\%, with failures primarily attributed to the limitations of stop-then-manipulate decomposition. While a fixed base placement can ensure reachability for a coarse initial pose, it guarantees neither a collision-free pre-grasp configuration nor robustness to target displacement. Consequently, this results in either pushing the object away before grasping or failing to reach the actual target pose. Thakar et al.~\cite{thakar2018towards} achieved 18.19\% success. This method assumes a perfect task specification and does not incorporate online task pose estimation to correct motion. Moreover, because its base navigation and manipulation are decoupled, the base trajectory may fail to position the robot in a configuration that admits a valid pre-grasp pose, causing the robot to topple the object during the approach and resulting in grasp failure. Overall, the failures of the baseline methods can be summarized into three categories: (i) decoupled base--manipulator motion generation that leads to infeasible manipulation motion under workspace limits or target displacement; (ii) insufficient treatment of object collision, which can disturb the target and invalidate the grasp; and (iii) lack of closed-loop correction to reconcile execution with task pose estimation. In contrast, our method achieves near-perfect performance across all scenarios by combining reliability-aware whole-body trajectory generation with safety-aware task-error compensation, thereby sustaining reliability throughout long-horizon mobile manipulation. 

\begin{figure*}[t]
\vspace{0.3cm}
\centering
\includegraphics[width=0.9\linewidth]{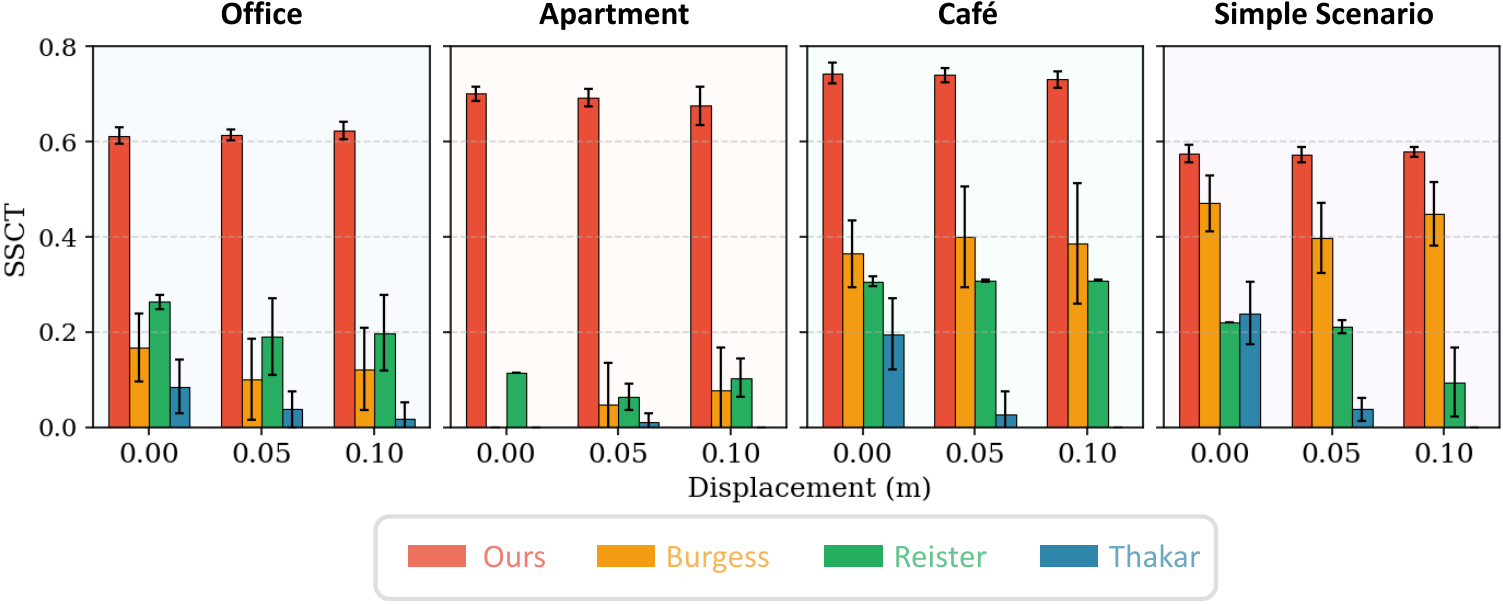}
% \vspace{-0.1cm}
\caption{Scenario Success-weighted Cycle Time (SSCT) for the benchmarked methods; bar heights denote the mean, and error bars the standard deviation, across 10 trials for each scenario–displacement combination.
}
\label{fig:benchmark_ssct}
\vspace{-0.3cm}
\end{figure*}

\begin{figure}[h]
\centering
\includegraphics[width=0.99\linewidth]{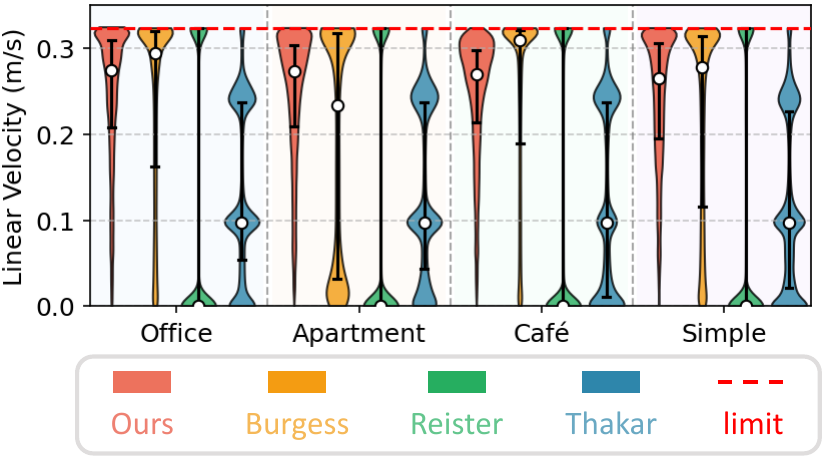}
\vspace{-0.2cm}
\caption{Distributions of mobile base linear velocity before collision or getting stuck for the four methods in each scenario, computed from 30 trials per scenario (10 trials for each displacement). White circles and capped vertical bars represent the median and interquartile range (IQR), respectively.
}
\label{fig:benchmark_linearvel}
\vspace{-0.3cm}
\end{figure}

Beyond reliability, the practical utility of a mobile manipulator also depends on how efficiently it executes the task sequence. However, efficiency is rendered futile without successful mission completion. Therefore, we evaluated efficiency using two complementary metrics: (i) the Scenario Success weighted by Cycle Time (SSCT) score (Fig.~\ref{fig:benchmark_ssct} and Tab.~\ref{tab:sim_benchmark}), which provides a single scenario-level score that jointly reflects both the success rate and the time efficiency, and (ii) the distribution of the linear velocity of the mobile base before the MM completes all the missions, gets stuck, or collides (Fig.~\ref{fig:benchmark_linearvel} and Tab.~\ref{tab:sim_benchmark}). Our method achieves the highest average base velocity across most scenarios, except for the Café scenario, where Burgess et al. exhibit a slightly higher mean velocity (0.2465 vs. our 0.2459 m/s), and attains the best SSCT across all four scenarios and displacement levels, indicating that it preserves fast mission execution while remaining robust to target pose perturbations. Burgess et al.~\cite{burgess2024reactive} exhibit a comparable average base speed, reflecting its aggressive on-the-move strategy; however, its overall efficiency is ultimately limited by the failure modes discussed above, which reduce effective throughput when failures are accounted for in SSCT. In contrast, Reister et al.~\cite{reister2022combining}, and Thakar et al.~\cite{thakar2018towards} are penalized both by lower success rates and by slower motion: the stop-then-manipulate decomposition in~\cite{reister2022combining} introduces frequent halts and re-alignments, while the separated plan--execute-replan structure in~\cite{thakar2018towards} induces additional waiting for replanning, both of which reduce efficiency. Overall, our framework improves mission throughput by jointly optimizing whole-body motion for feasibility and robustness, while using online, safety-aware task-error compensation to maintain high-speed execution under target displacement.

\subsection{Ablation Studies}
\label{subsec:ablation}

To answer the fifth question, we perform ablation studies to isolate the contributions of the main components in the proposed framework. We organize the analysis around two aspects that are central to overall system performance: (i) compensation ability under task-pose uncertainty (Sec.~\nameref{subsubsec:ablation_compensation}), and (ii) safe and precise motion generation during task execution (Sec.~\nameref{subsubsec:ablation_safe}). Across these studies, we use mission success rate to reflect reliability, and operation time to reflect efficiency. Specifically, operation time is defined as the sum of mission cycle time within a scenario:
\begin{equation}
\label{eq:operation_time}
    T_\mathrm{op} = \sum_{i=1}^{N} C_i,
\end{equation}
where $N$ is the number of missions in the scenario and $C_i$ is the mission cycle time of mission $i$. Lower $T_\mathrm{op}$ indicates higher efficiency.

\subsubsection{Ablation on Compensation Ability}
\label{subsubsec:ablation_compensation}
\begin{figure}[t]
\centering
\includegraphics[width=0.95\linewidth]{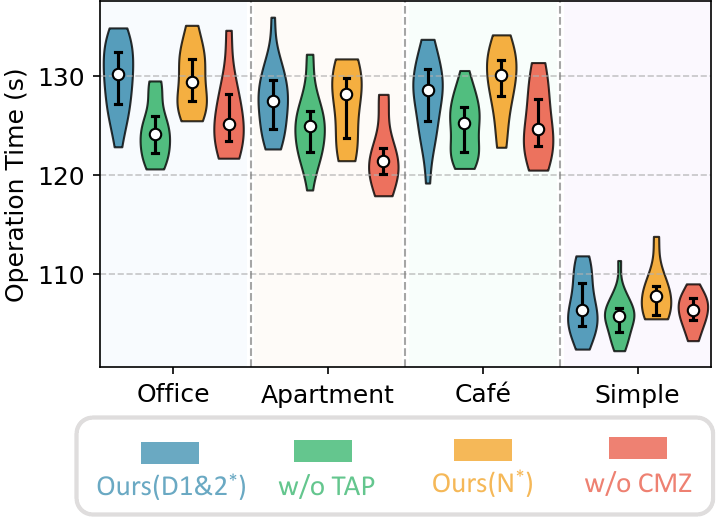}
\vspace{-0.2cm}
\caption{Distribution of operation time across four scenarios for our method and two ablations.
}
\label{fig:ablation_compensation_operation}
\vspace{-0.2cm}
\end{figure}

We ablate two key components that contribute to task-error compensation during trajectory generation, including (i) Time-assured Active Perception (TAP), and (ii) Compensation Margin Zone (CMZ). The experimental setting follows Sec.~\nameref{subsec:benchmark}, except that we inject an initial target displacement of $d=0.1\,\mathrm{m}$ to elicit compensation behavior. For each variant, we report mission success rate and operation time as shown in Fig.~\ref{fig:ablation_compensation_operation} and Tab.~\ref{tab:ablation_compensation}.

\begin{table}[t]
\footnotesize\sf\centering
\caption{Ablation results on compensation ability. For each variant, we report the mission success rate (SR) across 240 missions and the mean operation time averaged over 40 runs (10 per scenario).}
\label{tab:ablation_compensation}
\setlength{\tabcolsep}{7pt}
\renewcommand{\arraystretch}{1.2}
\begin{tabular}{c|c|cc}
\toprule
\textbf{Variants} & \textbf{SR (\%)} & \textbf{Operation Time (s)} \\
\midrule
w/o TAP(D1$\,^*$) & 90.42 & \textbf{119.87} \\
\rowcolor{gray!15} 
Ours(D1$\,^*$) & \textbf{99.58} & 123.24 \\
\midrule
w/o TAP(D2$\,^*$)    & 76.67 & \textbf{120.23} \\
\rowcolor{gray!15} 
Ours(D2$\,^*$)    & \textbf{99.17} & 122.91 \\
\midrule
w/o CMZ     & 88.75 & \textbf{119.97} \\
\rowcolor{gray!15} 
Ours(N$\,^*$)     & \textbf{100.0} & 123.56 \\
\bottomrule
\end{tabular}
\par\vspace{2pt}
\raggedright\footnotesize
{$\,^*$D1 and D2 denote perception-module initialization delay of 1 s and 2 s, respectively. N denotes keeping the trajectory generator conditioned on the coarse initial pose.}
\vspace{-0.1cm}
\end{table}

\begin{figure}[h]
\centering
\includegraphics[width=0.99\linewidth]{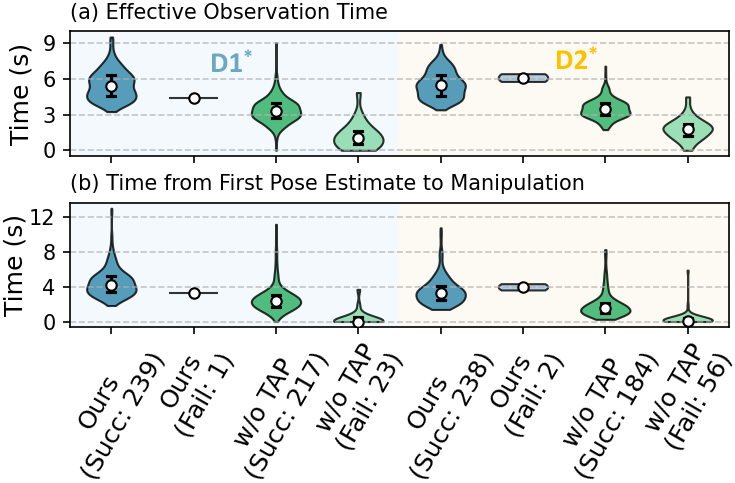}
\vspace{-0.1cm}
\caption{Distribution of (a) effective observation time and (b) time from first pose estimate to manipulation, comparing our method and the variant without time-assured active perception (TAP), under perception initialization delays of 1\,s (D1) and 2\,s (D2). D1 and D2 denote the perception-module initialization delay of 1 s and 2 s, respectively. 
}
\label{fig:ablation_compensation_observation}
\vspace{-0.3cm}
\end{figure}

\paragraph{Impact of Time-assured Active Perception (TAP).}
TAP explicitly enforces a guaranteed observation window before manipulation, ensuring that the perception pipeline has sufficient time to produce a pose estimate. We compare against a variant without TAP (\emph{w/o TAP}), which enforces an active-perception constraint only within a prescribed distance range (following Gao \emph{et al.}~\cite{gao2023adaptive}). To stress-test practical perception latency~\cite{wen2024foundationpose}, we emulate perception initialization delays of 1\,s (D1) and 2\,s (D2) for a new target. To better characterize the observation conditions for online pose estimation and the remaining time budget for compensation before task execution, we additionally report effective observation time $T_{\mathrm{eff}}$, and time from first pose estimate to manipulation $T_{\mathrm{mani}}$ (Fig.~\ref{fig:ablation_compensation_observation} and Tab.~\ref{tab:ablation_tap}). Here, $T_{\mathrm{eff}}$ refers to the duration during which the object remains within the camera's field of view without being occluded; $T_{\mathrm{mani}}$ refers to the time from the first available pose estimate to the start of manipulation.

\begin{table}[h]
\footnotesize\sf\centering
\caption{Ablation study of time-assured active perception (TAP) under different perception initialization delays. For each variant, results are grouped by mission outcome (Succ: success or failure), and we report the number of successful or failed missions (Cnt), together with the median and minimum (Min) values of effective observation time ($T_\mathrm{eff}$) and time from first pose estimate to manipulation ($T_\mathrm{mani}$).}
\label{tab:ablation_tap}
\setlength{\tabcolsep}{5.2pt}
\renewcommand{\arraystretch}{1.15}

\begin{tabular}{c|c|c|cc|cc}
\toprule
\multirow{2}{*}{\textbf{Variants}} &
\multirow{2}{*}{\textbf{Succ}} &
\multirow{2}{*}{\textbf{Cnt}} &
\multicolumn{2}{c}{$T_\mathrm{eff} (s)$} &
\multicolumn{2}{c}{$T_\mathrm{mani} (s)$} \\
& & & \textbf{Median} & \textbf{Min} & \textbf{Median} & \textbf{Min} \\
\midrule

\multirow{2}{*}{w/o TAP(D1$\,^*$)} &
\cmark & 217 & 3.28 & 0.00 & 2.40 & 0.00 \\
& \xmark & 23 & 1.03 & 0.00 & 0.00 & 0.00 \\
\midrule

\rowcolor{gray!15} 
& \cmark & 239 & 5.42 & 3.25 & 4.23 & 1.83 \\ 
\rowcolor{gray!15} 
\multirow{-2}{*}{Ours(D1$\,^*$)} & \xmark & 1 & 4.38 & 4.38 & 3.25 & 3.25 \\ \midrule

\multirow{2}{*}{w/o TAP(D2$\,^*$)} &
\cmark & 184 & 3.47 & 1.73 & 1.50 & 0.25 \\
& \xmark & 56 & 1.82 & 0.00 & 0.12 & 0.00 \\
\midrule

\rowcolor{gray!15} 
& \cmark & 238 & 5.44 & 3.40 & 3.30 & 1.38 \\
\rowcolor{gray!15}
\multirow{-2}{*}{Ours(D2$\,^*$)} & \xmark & 2 & 6.11 & 5.78 & 3.98 & 3.62 \\

\bottomrule
\end{tabular}
\par\vspace{2pt}
\raggedright\footnotesize
{$\,^*$D1 and D2 denote perception-module initialization delay of 1 s and 2 s, respectively.}
\vspace{-0.1cm}
\end{table}

As shown in Tab.~\ref{tab:ablation_tap}, our method maintains high success rates of 99.58\% (D1) and 99.17\% (D2), whereas \emph{w/o TAP} drops to 90.42\% and 76.67\%, respectively. The difference in guaranteed observation time explains this performance gap. In our formulation, we set the minimum required observation time to 3.0\,s during trajectory optimization. Correspondingly, the shortest effective observation time of our method observed in evaluation is 3.25 s (Fig.~\ref{fig:ablation_compensation_observation}(a) and Tab.~\ref{tab:ablation_tap}). In contrast, failures under \emph{w/o TAP} are associated with severely short effective observation time (median: 1.03\,s in D1 and 1.82\,s in D2). This leaves insufficient time for the perception pipeline to estimate the target pose before manipulation. As a result, the time gap between obtaining the first pose estimate and manipulation becomes near zero (Fig.~\ref{fig:ablation_compensation_observation}(b) and Tab.~\ref{tab:ablation_tap}; median: 0.00\,s in D1 and 0.12\,s in D2).

This failure mode arises from an efficiency--observability trade-off: enforcing observability only over a distance interval is inadequate, as it cannot ensure sufficient observation time to obtain a pose estimate. In contrast, by enforcing a time-guaranteed observation time in spatial-temporal trajectory optimization, our method eliminates this failure mode with only a modest 2.52\% increase in average operation time, from 120.05\,s for \emph{w/o TAP} to 123.08\,s for our method (Tab.~\ref{tab:ablation_compensation} and Fig.~\ref{fig:ablation_compensation_operation}).

\begin{figure*}[t]
\centering
\includegraphics[width=0.99\linewidth]{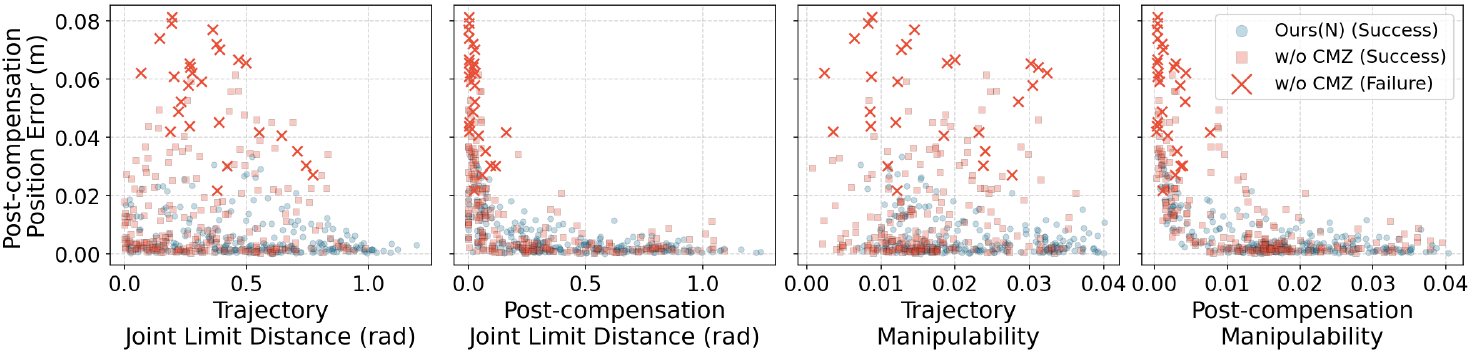}
\vspace{-0.2cm}
\caption{Scatter plots comparing position error against joint limit distance (left two) and manipulability (right two), evaluated along the global trajectory (“Trajectory”) and after compensation (“Post-compensation”) at the manipulation instant, comparing our method and the variant without compensation margin zone (CMZ). 
}
\label{fig:ablation_compensation_limit}
\vspace{-0.3cm}
\end{figure*}

\paragraph{Impact of Compensation Margin Zone (CMZ).}
\label{subsubsubsec:cmz}
CMZ requires that, at manipulation timestamps, the manipulator's base lies within a region from which the end-effector can reach a neighborhood of the task pose (rather than only the coarse initial pose), accounting for both manipulator configuration and joint limits, thereby preserving kinematic margin for correcting target displacements. To evaluate its contribution to reliability under target displacement, we disable CMZ (\emph{w/o CMZ}). Besides, to highlight the role of CMZ and eliminate interference from other variables, both this \emph{w/o CMZ} variant and our method (\emph{Ours(N)}) keep the trajectory generator conditioned on the coarse initial pose. As shown in Tab.~\ref{tab:ablation_compensation}, enabling CMZ increases the success rate from 88.75\% to 100.0\%, confirming that CMZ is crucial when the actual object pose deviates from the planned trajectory.

Fig.~\ref{fig:ablation_compensation_limit} further examines how post-compensation position error correlates with (i) joint-limit distance and (ii) manipulability, computed from 240 picking tasks. Here, post-compensation position error is defined as the difference between the target grasping pose of the object and the end-effector pose after online compensation. For each task, we compute these metrics from two manipulator configurations: (1) the configuration at the manipulation instant on the planned trajectory ("trajectory") and (2) the configuration at the manipulation instant after applying online compensation ("post-compensation"). Here, joint-limit distance measures how close the manipulator configuration is to its joint limits, defined as the minimum distance between each joint angle and its limit (smaller values indicate closer proximity to joint limits). Manipulability quantifies the local dexterity of the end-effector motion \cite{yoshikawa1985manipulability}, with larger values indicating more isotropic motion capacity and better conditioning. Across all four scatter plots, each point corresponds to a single grasping trial, with marker color/type indicating the success or failure of different variants. The results in Fig.~\ref{fig:ablation_compensation_limit} demonstrate that post-compensation position error is strongly correlated with joint-limit proximity. As shown in the left two panels in Fig.~\ref{fig:ablation_compensation_limit}, failures and large errors in the \emph{w/o CMZ} variant (red markers) predominantly occur when the manipulator is forced to operate near its joint limits during compensation. Notably, the right two panels in Fig.~\ref{fig:ablation_compensation_limit} show that high manipulability at the manipulation instant on the trajectory is not sufficient for successful compensation: several tasks exhibit high manipulability (e.g., $>0.03$) on trajectory yet still leave large position errors (e.g., $>0.06\,\mathrm{m}$) after compensation because the required corrective motion pushes the manipulator toward joint limits. This observation underscores a fundamental limitation of manipulability. As it is a local, velocity-based metric, it cannot determine whether the end-effector can reach a pose at a finite offset from the current state, given joint limit constraints and the current manipulator configuration.

CMZ addresses this limitation by sampling possible grasp candidates around the grasp pose derived from the coarse initial pose and constraining the manipulator base to lie within the region that can reach these candidates, calculated based on the inverse reachability map \cite{vahrenkamp2013robot} that accounts for both joint limits and configuration. This preserves the kinematic margin for corrective motion. Consequently, CMZ yields a substantially larger fraction of tasks with small error (94.17\% below 0.02 \,m versus 67.08\% without CMZ). The remaining outliers are mainly due to discretization in the sampling of grasp candidates and the resolution of the reachability map. Importantly, CMZ achieves these gains with only a minor increase of 2.99\% in the average operation time, from 119.97\,s for \emph{w/o CMZ} to 123.56\,s for our method (Tab.~\ref{tab:ablation_compensation} and Fig.~\ref{fig:ablation_compensation_operation}).

\subsubsection*{Ablation on Safe and Precise Motion Generation}
\label{subsubsec:ablation_safe}
Next, we evaluate the framework's ability to generate precise, safe motions. We ablate (i) Efficient Safe Interaction (ESI), (ii) Elastic Collision Spheres (ECS), and (iii) end-effector trajectory warping (WARP). For (i) and (ii), the experimental setting follows Sec.~\nameref{subsec:benchmark}, except that we use ground-truth task information (i.e., $d=0\,\mathrm{m}$) and disable both task-error compensation and the perception module, so that the results reflect motion-generation quality rather than compensation performance or pose-estimation accuracy. To evaluate (iii), we reuse the experimental setting of \textit{Impact of Compensation Margin Zone}: we inject an initial target displacement of $d=0.1\,\mathrm{m}$ to induce task-error compensation, and keep the trajectory generator conditioned on the coarse initial pose to highlight the role of WARP.

\begin{figure}[t]
\centering
\includegraphics[width=0.8\linewidth]{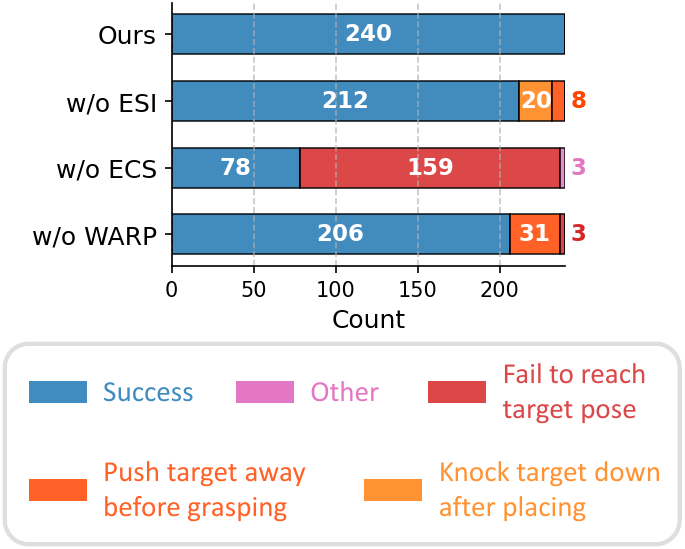}
\vspace{-0.1cm}
\caption{Stacked bar chart showing success count with failure modes across 240 missions for our method and three ablation variants.
}
\label{fig:ablation_safe_failuremode}
\vspace{-0.2cm}
\end{figure}

% Fig.~\ref{fig:ablation_safe_failuremode} reports success and failure modes over 240 missions, while Fig.~\ref{fig:ablation_safe_operation} shows the distributions of operation time.

\paragraph{Impact of Efficient Safe Interaction (ESI).}
ESI produces interaction-aware end-effector motions during the pre-grasp and post-place phases, aiming to prevent unintended object disturbance (e.g., toppling) during approach and retraction while still allowing the timing to be optimized. We disable ESI (\emph{w/o ESI}) so that the end-effector approaches the grasp pose and retracts from the placement pose without enforcing safe interaction motion.

\begin{figure}[h]
\centering
\includegraphics[width=0.99\linewidth]{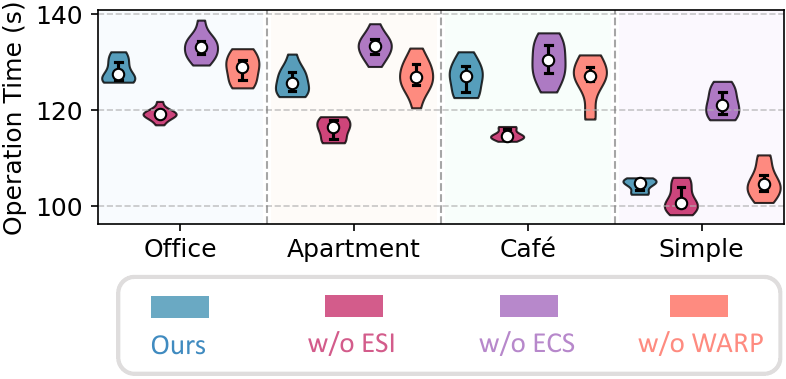}
\vspace{-0.1cm}
\caption{Distribution of operation time for four method configurations across four scenarios.
}
\label{fig:ablation_safe_operation}
\vspace{-0.2cm}
\end{figure}

As shown in Fig.~\ref{fig:ablation_safe_failuremode}, removing ESI reduces the success rate from 100\% to 88.34\%. Failures are dominated by object toppling, with 20 failures in post-place and 8 failures in pre-grasp. This failure pattern is attributable to the lack of explicit interaction-safety constraints between the end-effector and the manipulated object during approach and retraction, leading to unintended contacts that can cause the object to topple. In contrast, ESI constrains the end-effector to follow an interaction-safe motion primitive with optimized timing, which effectively suppresses toppling failures. Importantly, the associated operation time cost is modest: Fig.~\ref{fig:ablation_safe_operation} shows that ESI increases operation time by approximately 1.02\,s per ESI-constrained motion on average (8.42\,s total overhead across 12 tasks containing 8--10 ESI-constrained motions in each scenario), demonstrating a favorable trade-off between safety and efficiency.

\begin{figure}[h]
\centering
\includegraphics[width=0.99\linewidth]{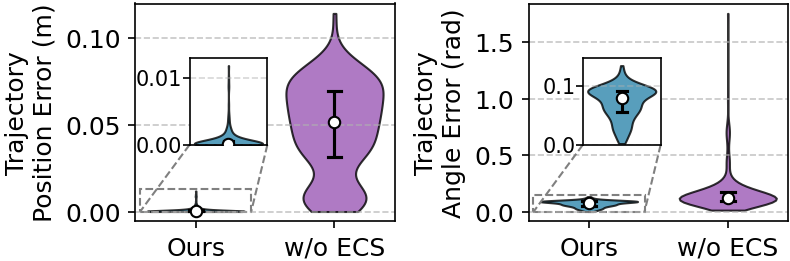}
\vspace{-0.2cm}
\caption{Distribution of end-effector position (left) and angle (right) errors between the planned and target manipulation pose, comparing our method with a variant without elastic collision spheres.
}
\label{fig:ablation_safe_precise}
\vspace{-0.2cm}
\end{figure}

\paragraph{Impact of Elastic Collision Sphere (ECS).}
We disable ECS (\emph{w/o ECS}) by removing the elastic relaxation mechanism and performing collision checking using fixed-radius collision spheres only. As shown in Fig.~\ref{fig:ablation_safe_failuremode}, this variant exhibits a sharp drop in success rate to 32.5\%, with 159 out of 240 trials failing because the end-effector cannot reach the task target pose. Fig.~\ref{fig:ablation_safe_precise} further shows markedly increased pose error at the manipulation instant: the median position error increases from 0.000181\,m (Ours) to 0.051491\,m (\emph{w/o ECS}), and the median orientation error increases from 0.079\,rad to 0.120\,rad.

These failures arise from a feasibility conflict near the supporting surfaces. Strict collision constraints tend to separate the object from the support surface, whereas the task requires contact with the surface (e.g., placing the object on the surface or grasping it from the surface). As a result, the desired task pose can become infeasible under purely fixed collision checking, preventing the end-effector from reaching it. ECS resolves this contact–safety conflict via state-dependent radius modulation, which (i) preserves reachability of intended-contact poses and (ii) biases subsequent motion toward increasing clearance through radius recovery to prevent unintended contacts between object and supporting surfaces.

\paragraph{Impact of End-effector Trajectory Warping (WARP).}
WARP re-targets the end-effector trajectory to the latest pose estimate while explicitly preserving the pre- and post-task safety motions generated by the planner. In particular, by smoothly blending into a rigidly re-targeted safe interaction segment, WARP maintains the relative motion pattern that avoids unintended contacts (e.g., end-effector pushing the target away before grasping) during compensation.

We disable WARP (\emph{w/o WARP}), and the desired end-effector pose is directly switched to the updated task target from 1\,s before manipulation until 0.5\,s after manipulation, without the safety-preserving warping and blending around the task-critical phase. As shown in Fig.~\ref{fig:ablation_safe_failuremode}, removing WARP reduces the success rate to 85.84\% (206/240), with failures dominated by the end effector pushing the target away before reaching the grasp pose (31/240).
The failures occur because, without WARP, the end-effector is driven directly toward the latest pose estimate without explicitly accounting for contacts between the end-effector and the object. As a result, the end-effector often pushes the object away from its original pose just before grasping, leading to grasp failures. 
In contrast, enabling WARP preserves the pre- and post-task safety segments while smoothly re-targeting them to the latest pose estimate, which avoids these failures. 

\section{Conclusion}

Our results validate a unified solution to the long-standing trade-off between operational efficiency and execution reliability in mobile manipulation. Whereas efficiency demands aggressive, continuous motion, reliability typically requires conservative stop-and-verify behaviors. By explicitly coupling these objectives, our framework demonstrates that (i) continuous whole-body coordination is sustainable even in constrained, dynamic environments, and (ii) efficient execution need not compromise robustness, even under dynamic disturbances, perception, and control errors.

The observed performance advantage over state-of-the-art methods stems from the departure of our framework from traditional decoupled approaches, which typically separate base navigation from manipulation or optimize motion efficiency without regard for execution constraints. Purely efficiency-driven planners often fail because they push robots to the limits of their kinematics, shrink perception windows, and neglect interaction safety. In contrast, our reliability-aware trajectory generation treats reliability as a proactive objective rather than a reactive afterthought. By explicitly encoding sufficient observability (TAP), kinematic margin (CMZ), and interaction safety (ESI/ECS) in trajectory optimization, the planner generates motions that are not only efficient but also inherently correctable and interaction-safe during execution.

Furthermore, our results highlight the need for a hierarchical architecture to address the frequency mismatch between whole-body planning and real-time compensation. The proposed high-frequency execution layer effectively converts the planner’s long-horizon structure into immediate physical responsiveness. By smoothly transitioning between global trajectory tracking and local task-error compensation, the system preserves the safe-interaction structure of the planned trajectory while adapting to disturbances and updated estimates online. Ultimately, this synergy allows the system to accommodate the large discrepancies between coarse priors and refined estimates (as seen in the domino experiments) without interrupting the continuous flow of successive tasks.

Beyond immediate performance gains, our framework can serve as a robust physical grounding layer for high-level task generation modules. We complement such generators~\cite{huang2024rekep,hsu2025spot}, which excel at semantic reasoning and predicating object-centric goals or desired end-effector motions, by translating the task output into safe and efficient whole-body motions. This capability bridges the critical gap between abstract task generation and physical reality, enabling efficient and reliable execution in complex 3D environments.

Although our framework supports online replanning, its computational efficiency could be further improved via GPU acceleration for collision checking, inverse kinematics, and trajectory optimization \cite{sundaralingam2023curobo}, which would accelerate trajectory generation by orders of magnitude compared to CPU-based approaches~\cite{schulman2014motion}. Furthermore, although our framework can react to dynamic obstacles, our long-horizon whole-body optimization currently assumes a coarse prior map of the static environment for collision checking. This mirrors human behavior: efficient motions typically rely on a rough model of the surroundings, while perception handles local, transient changes. Operating in a completely unknown complex 3D environment would require an active mapping strategy~\cite{pilania2018mobile} to actively resolve workspace occlusions and verify free space along the intended manipulation trajectory. To extend our framework to other platforms, one mainly needs to replace the platform-specific kinematic and dynamic models (e.g., forward kinematics and platform dynamic limits) used by the whole-body optimizer and controller. For platforms operating on uneven terrain, it should additionally incorporate terrain-aware reachability map \cite{birr2022oriented} and traversability representations \cite{xu2023efficient,li2025seb} to identify feasible regions and constrain base motion accordingly.

Overall, our framework demonstrates that reliability can be effectively embedded into efficient whole-body motion for continuous mobile manipulation. We hope this work will pave the way for robotic systems capable of seamlessly integrating efficiency and reliability for continuous mobile manipulation in the real world.

\section{Declaration of Conflicting Interests}
The author(s) declared no potential conflicts of interest with respect to the research, authorship, and/or publication of this article.

% \section{Funding}
% The author(s) received no financial support for the research, authorship, and/or publication of this article.

\bibliographystyle{SageH}
\bibliography{references}

@Misc{agilextracerminiurl,
  title        = {TRACER ROS},
  author       = {AgileX},
  url          = {https://github.com/agilexrobotics/tracer_ros},
  year         = {2026}
}

@Misc{unitreez1url,
  title        = {Unitree Z1 User Manual},
  author       = {Unitree},
  url          = {https://www.unitree.com/cn/z1},
  year         = {2026}
}

@Misc{mid360url,
  title        = {Livox Mid-360 User Manual},
  author       = {Livox},
  url          = {https://www.livoxtech.com/mid-360},
  year         = {2026}
}

@Misc{femtobolturl,
  title        = {Orbbec Femto Bolt User Manual},
  author       = {Orbbec},
  url          = {https://www.orbbec.com/products/tof-camera/femto-bolt},
  year         = {2026}
}

@Misc{isaacsimurl,
  title        = {NVIDIA Isaac Sim Documentation},
  author       = {NVIDIA},
  url          = {https://docs.isaacsim.omniverse.nvidia.com/4.2.0/index.html},
  year         = {2026}
}

@inproceedings{burgess2023architecture,
  title={An architecture for reactive mobile manipulation on-the-move},
  author={Burgess-Limerick, Ben and Lehnert, Chris and Leitner, J{\"u}rgen and Corke, Peter},
  booktitle={2023 IEEE International Conference on Robotics and Automation (ICRA)},
  pages={1623--1629},
  year={2023},
  organization={IEEE}
}

@Article{Verschueren2021,
  Title                    = {acados -- a modular open-source framework for fast embedded optimal control},
  Author                   = {Robin Verschueren and Gianluca Frison and Dimitris Kouzoupis and Jonathan Frey and Niels van Duijkeren and Andrea Zanelli and Branimir Novoselnik and Thivaharan Albin and Rien Quirynen and Moritz Diehl},
  Journal                  = {Mathematical Programming Computation},
  Year                     = {2021},
}

@article{burgess2024reactive,
  title={Reactive base control for on-the-move mobile manipulation in dynamic environments},
  author={Burgess-Limerick, Ben and Haviland, Jesse and Lehnert, Chris and Corke, Peter},
  journal={IEEE Robotics and Automation Letters},
  volume={9},
  number={3},
  pages={2048--2055},
  year={2024},
  publisher={IEEE}
}

@article{small_gicp,
  author = {Kenji Koide},
  title = {{small\_gicp: Efficient and parallel algorithms for point cloud registration}},
  journal = {Journal of Open Source Software},
  month = aug,
  number = {100},
  pages = {6948},
  volume = {9},
  year = {2024},
  doi = {10.21105/joss.06948}
}

@article{dolgov2010path,
  title={Path planning for autonomous vehicles in unknown semi-structured environments},
  author={Dolgov, Dmitri and Thrun, Sebastian and Montemerlo, Michael and Diebel, James},
  journal={The international journal of robotics research},
  volume={29},
  number={5},
  pages={485--501},
  year={2010},
  publisher={SAGE Publications Sage UK: London, England}
}

@article{frison2020hpipm,
  title={HPIPM: a high-performance quadratic programming framework for model predictive control},
  author={Frison, Gianluca and Diehl, Moritz},
  journal={IFAC-PapersOnLine},
  volume={53},
  number={2},
  pages={6563--6569},
  year={2020},
  publisher={Elsevier}
}

@article{pilania2018mobile,
  title={Mobile manipulator planning under uncertainty in unknown environments},
  author={Pilania, Vinay and Gupta, Kamal},
  journal={The International Journal of Robotics Research},
  volume={37},
  number={2-3},
  pages={316--339},
  year={2018},
  publisher={SAGE Publications Sage UK: London, England}
}

@inproceedings{wen2024foundationpose,
  title={Foundationpose: Unified 6d pose estimation and tracking of novel objects},
  author={Wen, Bowen and Yang, Wei and Kautz, Jan and Birchfield, Stan},
  booktitle={Proceedings of the IEEE/CVF Conference on Computer Vision and Pattern Recognition},
  pages={17868--17879},
  year={2024}
}

@inproceedings{hsu2025spot,
  title={Spot: Se (3) pose trajectory diffusion for object-centric manipulation},
  author={Hsu, Cheng-Chun and Wen, Bowen and Xu, Jie and Narang, Yashraj and Wang, Xiaolong and Zhu, Yuke and Biswas, Joydeep and Birchfield, Stan},
  booktitle={2025 IEEE International Conference on Robotics and Automation (ICRA)},
  pages={4853--4860},
  year={2025},
  organization={IEEE}
}

@article{yoshikawa1985manipulability,
  title={Manipulability of robotic mechanisms},
  author={Yoshikawa, Tsuneo},
  journal={The international journal of Robotics Research},
  volume={4},
  number={2},
  pages={3--9},
  year={1985},
  publisher={Sage Publications Sage CA: Thousand Oaks, CA}
}

@inproceedings{sundaralingam2023curobo,
  title={Curobo: Parallelized collision-free robot motion generation},
  author={Sundaralingam, Balakumar and Hari, Siva Kumar Sastry and Fishman, Adam and Garrett, Caelan and Van Wyk, Karl and Blukis, Valts and Millane, Alexander and Oleynikova, Helen and Handa, Ankur and Ramos, Fabio and others},
  booktitle={2023 IEEE International Conference on Robotics and Automation (ICRA)},
  pages={8112--8119},
  year={2023},
  organization={IEEE}
}

@article{heins2023keep,
  title={Keep it upright: Model predictive control for nonprehensile object transportation with obstacle avoidance on a mobile manipulator},
  author={Heins, Adam and Schoellig, Angela P},
  journal={IEEE Robotics and Automation Letters},
  volume={8},
  number={12},
  pages={7986--7993},
  year={2023},
  publisher={IEEE}
}

@inproceedings{birr2022oriented,
  title={Oriented surface reachability maps for robot placement},
  author={Birr, Timo and Pohl, Christoph and Asfour, Tamim},
  booktitle={2022 International Conference on Robotics and Automation (ICRA)},
  pages={3357--3363},
  year={2022},
  organization={IEEE}
}

@inproceedings{xu2023efficient,
  title={An efficient trajectory planner for car-like robots on uneven terrain},
  author={Xu, Long and Chai, Kaixin and Han, Zhichao and Liu, Hong and Xu, Chao and Cao, Yanjun and Gao, Fei},
  booktitle={2023 IEEE/RSJ International Conference on Intelligent Robots and Systems (IROS)},
  pages={2853--2860},
  year={2023},
  organization={IEEE}
}

@article{li2025seb,
  title={SEB-Naver: A SE (2)-based Local Navigation Framework for Car-like Robots on Uneven Terrain},
  author={Li, Xiaoying and Xu, Long and Huang, Xiaolin and Xue, Donglai and Zhang, Zhihao and Han, Zhichao and Xu, Chao and Cao, Yanjun and Gao, Fei},
  journal={arXiv preprint arXiv:2503.02412},
  year={2025}
}

@article{schulman2014motion,
  title={Motion planning with sequential convex optimization and convex collision checking},
  author={Schulman, John and Duan, Yan and Ho, Jonathan and Lee, Alex and Awwal, Ibrahim and Bradlow, Henry and Pan, Jia and Patil, Sachin and Goldberg, Ken and Abbeel, Pieter},
  journal={The International Journal of Robotics Research},
  volume={33},
  number={9},
  pages={1251--1270},
  year={2014},
  publisher={Sage Publications Sage UK: London, England}
}

@article{xu2021planning,
  title={Planning a minimum sequence of positions for picking parts from multiple trays using a mobile manipulator},
  author={Xu, Jingren and Domae, Yukiyasu and Ueshiba, Toshio and Wan, Weiwei and Harada, Kensuke},
  journal={IEEE Access},
  volume={9},
  pages={165526--165541},
  year={2021},
  publisher={IEEE}
}

@inproceedings{xu2020planning,
  title={Planning an efficient and robust base sequence for a mobile manipulator performing multiple pick-and-place tasks},
  author={Xu, Jingren and Harada, Kensuke and Wan, Weiwei and Ueshiba, Toshio and Domae, Yukiyasu},
  booktitle={2020 IEEE International Conference on Robotics and Automation (ICRA)},
  pages={11018--11024},
  year={2020},
  organization={IEEE}
}

@article{wu2023tidybot,
  title={Tidybot: Personalized robot assistance with large language models},
  author={Wu, Jimmy and Antonova, Rika and Kan, Adam and Lepert, Marion and Zeng, Andy and Song, Shuran and Bohg, Jeannette and Rusinkiewicz, Szymon and Funkhouser, Thomas},
  journal={Autonomous Robots},
  volume={47},
  number={8},
  pages={1087--1102},
  year={2023},
  publisher={Springer}
}

@article{zucker2013chomp,
  title={Chomp: Covariant hamiltonian optimization for motion planning},
  author={Zucker, Matt and Ratliff, Nathan and Dragan, Anca D and Pivtoraiko, Mihail and Klingensmith, Matthew and Dellin, Christopher M and Bagnell, J Andrew and Srinivasa, Siddhartha S},
  journal={The International journal of robotics research},
  volume={32},
  number={9-10},
  pages={1164--1193},
  year={2013},
  publisher={SAGE Publications Sage UK: London, England}
}

@article{pilania2015hierarchical,
  title={A hierarchical and adaptive mobile manipulator planner with base pose uncertainty},
  author={Pilania, Vinay and Gupta, Kamal},
  journal={Autonomous Robots},
  volume={39},
  number={1},
  pages={65--85},
  year={2015},
  publisher={Springer}
}

@phdthesis{diankov2010automated,
  title={Automated construction of robotic manipulation programs},
  author={Diankov, Rosen},
  year={2010},
  school={Carnegie Mellon University, USA}
}

@article{zhou2019robust,
  title={Robust and efficient quadrotor trajectory generation for fast autonomous flight},
  author={Zhou, Boyu and Gao, Fei and Wang, Luqi and Liu, Chuhao and Shen, Shaojie},
  journal={IEEE Robotics and Automation Letters},
  volume={4},
  number={4},
  pages={3529--3536},
  year={2019},
  publisher={IEEE}
}

@article{deng2025whole,
  title={Whole-body integrated motion planning for aerial manipulators},
  author={Deng, Weiliang and Chen, Hongming and Ye, Biyu and Chen, Haoran and Li, Ziliang and Lyu, Ximin},
  journal={IEEE Transactions on Robotics},
  volume={41},
  pages={6661--6679},
  year={2025},
  publisher={IEEE}
}

@article{xu2025topay,
  title={TopAY: Efficient Trajectory Planning for Differential Drive Mobile Manipulators via Topological Paths Search and Arc Length-Yaw Parameterization},
  author={Xu, Long and Wong, Choilam and Zhang, Mengke and Lin, Junxiao and Hou, Jialiang and Gao, Fei},
  journal={arXiv preprint arXiv:2507.02761},
  year={2025}
}

@inproceedings{thakar2020accelerating,
  title={Accelerating bi-directional sampling-based search for motion planning of non-holonomic mobile manipulators},
  author={Thakar, Shantanu and Rajendran, Pradeep and Kim, Hyojeong and Kabir, Ariyan M and Gupta, Satyandra K},
  booktitle={2020 IEEE/RSJ International Conference on Intelligent Robots and Systems (IROS)},
  pages={6711--6717},
  year={2020},
  organization={IEEE}
}

@inproceedings{thakar2018towards,
  title={Towards time-optimal trajectory planning for pick-and-transport operation with a mobile manipulator},
  author={Thakar, Shantanu and Fang, Liwei and Shah, Brual and Gupta, Satyandra},
  booktitle={2018 IEEE 14th International Conference on Automation Science and Engineering (CASE)},
  pages={981--987},
  year={2018},
  organization={IEEE}
}

@article{ichnowski2020deep,
  title={Deep learning can accelerate grasp-optimized motion planning},
  author={Ichnowski, Jeffrey and Avigal, Yahav and Satish, Vishal and Goldberg, Ken},
  journal={Science Robotics},
  volume={5},
  number={48},
  pages={eabd7710},
  year={2020},
  publisher={American Association for the Advancement of Science}
}

@article{thakar2020manipulator,
  title={Manipulator motion planning for part pickup and transport operations from a moving base},
  author={Thakar, Shantanu and Rajendran, Pradeep and Kabir, Ariyan M and Gupta, Satyandra K},
  journal={IEEE Transactions on Automation Science and Engineering},
  volume={19},
  number={1},
  pages={191--206},
  year={2020},
  publisher={IEEE}
}

@article{gao2023adaptive,
  title={Adaptive tracking and perching for quadrotor in dynamic scenarios},
  author={Gao, Yuman and Ji, Jialin and Wang, Qianhao and Jin, Rui and Lin, Yi and Shang, Zhimeng and Cao, Yanjun and Shen, Shaojie and Xu, Chao and Gao, Fei},
  journal={IEEE Transactions on Robotics},
  volume={40},
  pages={499--519},
  year={2023},
  publisher={IEEE}
}

@inproceedings{vahrenkamp2013robot,
  title={Robot placement based on reachability inversion},
  author={Vahrenkamp, Nikolaus and Asfour, Tamim and Dillmann, R{\"u}diger},
  booktitle={2013 IEEE International Conference on Robotics and Automation},
  pages={1970--1975},
  year={2013},
  organization={IEEE}
}

@INPROCEEDINGS{liang2025dynamicpose,
  author={Liang, Tingbang and Zeng, Yixin and Xie, JiaTong and Zhou, Boyu},
  booktitle={2025 IEEE/RSJ International Conference on Intelligent Robots and Systems (IROS)}, 
  title={DynamicPose: Real-time and Robust 6D Object Pose Tracking for Fast-Moving Cameras and Objects}, 
  year={2025},
  volume={},
  number={},
  pages={2424-2431},
  doi={10.1109/IROS60139.2025.11247403}
}

@inproceedings{vosylius2024instant,
title={Instant Policy: In-Context Imitation Learning via Graph Diffusion},
author={Vosylius, Vitalis and Johns, Edward},
booktitle = {Proceedings of the International Conference on Learning Representations (ICLR)},
year={2025}
}

@article{xu2022fast,
  title={Fast-lio2: Fast direct lidar-inertial odometry},
  author={Xu, Wei and Cai, Yixi and He, Dongjiao and Lin, Jiarong and Zhang, Fu},
  journal={IEEE Transactions on Robotics},
  volume={38},
  number={4},
  pages={2053--2073},
  year={2022},
  publisher={IEEE}
}

@article{reister2022combining,
  title={Combining navigation and manipulation costs for time-efficient robot placement in mobile manipulation tasks},
  author={Reister, Fabian and Grotz, Markus and Asfour, Tamim},
  journal={IEEE Robotics and Automation Letters},
  volume={7},
  number={4},
  pages={9913--9920},
  year={2022},
  publisher={IEEE}
}

@article{abdi2010principal,
  title={Principal component analysis},
  author={Abdi, Herv{\'e} and Williams, Lynne J},
  journal={Wiley interdisciplinary reviews: computational statistics},
  volume={2},
  number={4},
  pages={433--459},
  year={2010},
  publisher={Wiley Online Library}
}

@inproceedings{zimmermann2021go,
  title={Go fetch!-dynamic grasps using boston dynamics spot with external robotic arm},
  author={Zimmermann, Simon and Poranne, Roi and Coros, Stelian},
  booktitle={2021 IEEE International Conference on Robotics and Automation (ICRA)},
  pages={4488--4494},
  year={2021},
  organization={IEEE}
}

@article{du2023hierarchical,
  title={Hierarchical Task Model Predictive Control for Sequential Mobile Manipulation Tasks},
  author={Du, Xintong and Zhou, Siqi and Schoellig, Angela P},
  journal={IEEE Robotics and Automation Letters},
  volume={9},
  number={2},
  pages={1270--1277},
  year={2023},
  publisher={IEEE}
}

@inproceedings{jauhri2024active,
  title={Active-perceptive motion generation for mobile manipulation},
  author={Jauhri, Snehal and Lueth, Sophie and Chalvatzaki, Georgia},
  booktitle={2024 IEEE International Conference on Robotics and Automation (ICRA)},
  pages={1413--1419},
  year={2024},
  organization={IEEE}
}

@inproceedings{zhang2024gamma,
  title={Gamma: Graspability-aware mobile manipulation policy learning based on online grasping pose fusion},
  author={Zhang, Jiazhao and Gireesh, Nandiraju and Wang, Jilong and Fang, Xiaomeng and Xu, Chaoyi and Chen, Weiguang and Dai, Liu and Wang, He},
  booktitle={2024 IEEE International Conference on Robotics and Automation (ICRA)},
  pages={1399--1405},
  year={2024},
  organization={IEEE}
}

@article{haviland2022holistic,
  title={A holistic approach to reactive mobile manipulation},
  author={Haviland, Jesse and S{\"u}nderhauf, Niko and Corke, Peter},
  journal={IEEE Robotics and Automation Letters},
  volume={7},
  number={2},
  pages={3122--3129},
  year={2022},
  publisher={IEEE}
}

@inproceedings{wang2024arm,
  title={Arm-constrained curriculum learning for loco-manipulation of a wheel-legged robot},
  author={Wang, Zifan and Jia, Yufei and Shi, Lu and Wang, Haoyu and Zhao, Haizhou and Li, Xueyang and Zhou, Jinni and Ma, Jun and Zhou, Guyue},
  booktitle={2024 IEEE/RSJ International Conference on Intelligent Robots and Systems (IROS)},
  pages={10770--10776},
  year={2024},
  organization={IEEE}
}

@inproceedings{wang2025ehc,
  title={EHC-MM: Embodied Holistic Control for Mobile Manipulation},
  author={Wang, Jiawen and Jin, Yixiang and Shi, Jun and Li, Dingzhe and Sun, Fuchun and Luo, Dingsheng and Fang, Bin and others},
  booktitle={2025 IEEE International Conference on Robotics and Automation (ICRA)},
  pages={13330--13336},
  year={2025},
  organization={IEEE}
}

@article{spahn2023dynamic,
  title={Dynamic optimization fabrics for motion generation},
  author={Spahn, Max and Wisse, Martijn and Alonso-Mora, Javier},
  journal={IEEE Transactions on Robotics},
  volume={39},
  number={4},
  pages={2684--2699},
  year={2023},
  publisher={IEEE}
}

@inproceedings{missura2022fast,
  title={Fast-replanning motion control for non-holonomic vehicles with aborting A},
  author={Missura, Marcell and Roychoudhury, Arindam and Bennewitz, Maren},
  booktitle={2022 IEEE/RSJ International Conference on Intelligent Robots and Systems (IROS)},
  pages={10267--10274},
  year={2022},
  organization={IEEE}
}

@article{spahn2024demonstrating,
  title={Demonstrating adaptive mobile manipulation in retail environments},
  author={Spahn, Max and Pezzato, Corrado and Salmi, Chadi and Dekker, Rick and Wang, Cong and Pek, Christian and Kober, Jens and Alonso-Mora, Javier and Corbato, C Hernandez and Wisse, Martijn},
  journal={Proceedings of the Robotics: Science and System XX},
  year={2024}
}

@inproceedings{spahn2021coupled,
  title={Coupled mobile manipulation via trajectory optimization with free space decomposition},
  author={Spahn, Max and Brito, Bruno and Alonso-Mora, Javier},
  booktitle={2021 IEEE International Conference on Robotics and Automation (ICRA)},
  pages={12759--12765},
  year={2021},
  organization={IEEE}
}

@inproceedings{wu2024real,
  title={Real-time whole-body motion planning for mobile manipulators using environment-adaptive search and spatial-temporal optimization},
  author={Wu, Chengkai and Wang, Ruilin and Song, Mianzhi and Gao, Fei and Mei, Jie and Zhou, Boyu},
  booktitle={2024 IEEE International Conference on Robotics and Automation (ICRA)},
  pages={1369--1375},
  year={2024},
  organization={IEEE}
}

@article{burger2020mobile,
  title={A mobile robotic chemist},
  author={Burger, Benjamin and Maffettone, Phillip M and Gusev, Vladimir V and Aitchison, Catherine M and Bai, Yang and Wang, Xiaoyan and Li, Xiaobo and Alston, Ben M and Li, Buyi and Clowes, Rob and others},
  journal={Nature},
  volume={583},
  number={7815},
  pages={237--241},
  year={2020},
  publisher={Nature Publishing Group UK London}
}

@article{dai2024autonomous,
  title={Autonomous mobile robots for exploratory synthetic chemistry},
  author={Dai, Tianwei and Vijayakrishnan, Sriram and Szczypi{\'n}ski, Filip T and Ayme, Jean-Fran{\c{c}}ois and Simaei, Ehsan and Fellowes, Thomas and Clowes, Rob and Kotopanov, Lyubomir and Shields, Caitlin E and Zhou, Zhengxue and others},
  journal={Nature},
  volume={635},
  number={8040},
  pages={890--897},
  year={2024},
  publisher={Nature Publishing Group UK London}
}

@article{zhang2022learning,
  title={Learning garment manipulation policies toward robot-assisted dressing},
  author={Zhang, Fan and Demiris, Yiannis},
  journal={Science robotics},
  volume={7},
  number={65},
  pages={eabm6010},
  year={2022},
  publisher={American Association for the Advancement of Science}
}

@article{johns2023framework,
  title={A framework for robotic excavation and dry stone construction using on-site materials},
  author={Johns, Ryan Luke and Wermelinger, Martin and Mascaro, Ruben and Jud, Dominic and Hurkxkens, Ilmar and Vasey, Lauren and Chli, Margarita and Gramazio, Fabio and Kohler, Matthias and Hutter, Marco},
  journal={Science Robotics},
  volume={8},
  number={84},
  pages={eabp9758},
  year={2023},
  publisher={American Association for the Advancement of Science}
}

@inproceedings{yokoyama2021success,
  title={Success weighted by completion time: A dynamics-aware evaluation criteria for embodied navigation},
  author={Yokoyama, Naoki and Ha, Sehoon and Batra, Dhruv},
  booktitle={2021 IEEE/RSJ International Conference on Intelligent Robots and Systems (IROS)},
  pages={1562--1569},
  year={2021},
  organization={IEEE}
}

@article{WANG2022GCOPTER,
    title={Geometrically Constrained Trajectory Optimization for Multicopters}, 
    author={Wang, Zhepei and Zhou, Xin and Xu, Chao and Gao, Fei}, 
    journal={IEEE Transactions on Robotics}, 
    year={2022}, 
    volume={38}, 
    number={5}, 
    pages={3259-3278}, 
    doi={10.1109/TRO.2022.3160022}
}

@incollection{dijkstra2022note,
  title={A note on two problems in connexion with graphs},
  author={Dijkstra, Edsger W},
  booktitle={Edsger Wybe Dijkstra: his life, work, and legacy},
  pages={287--290},
  year={2022}
}

@article{rockafellar1974augmented,
  title={Augmented Lagrange multiplier functions and duality in nonconvex programming},
  author={Rockafellar, R Tyrrell},
  journal={SIAM Journal on Control},
  volume={12},
  number={2},
  pages={268--285},
  year={1974},
  publisher={SIAM}
}

@INPROCEEDINGS{ShenW-RSS-25, 
    AUTHOR    = {William Shen AND Caelan Reed Garrett AND Nishanth Kumar AND Ankit Goyal AND Tucker Hermans AND Leslie Pack Kaelbling AND Tomás Lozano-Pérez AND Fabio Ramos}, 
    TITLE     = {{Differentiable GPU-Parallelized Task and Motion Planning}}, 
    BOOKTITLE = {Proceedings of Robotics: Science and Systems}, 
    YEAR      = {2025}, 
    ADDRESS   = {Los Angeles, CA, USA}, 
    MONTH     = {June}, 
    DOI       = {10.15607/RSS.2025.XXI.050} 
}

@article{huang2024rekep,
  title={Rekep: Spatio-temporal reasoning of relational keypoint constraints for robotic manipulation},
  author={Huang, Wenlong and Wang, Chen and Li, Yunzhu and Zhang, Ruohan and Fei-Fei, Li},
  journal={arXiv preprint arXiv:2409.01652},
  year={2024}
}

@inproceedings{pan2025omnimanip,
  title={Omnimanip: Towards general robotic manipulation via object-centric interaction primitives as spatial constraints},
  author={Pan, Mingjie and Zhang, Jiyao and Wu, Tianshu and Zhao, Yinghao and Gao, Wenlong and Dong, Hao},
  booktitle={Proceedings of the Computer Vision and Pattern Recognition Conference},
  pages={17359--17369},
  year={2025}
}

% Please note that the files \textsf{SageH.bst} and \textsf{SageV.bst} are included with the class file
% for those authors using \BibTeX.
% The files work in a completely standard way, and you just need to uncomment one of the lines in the below example depending on what style you require:
% \begin{verbatim}
% %%Harvard (name/date)
% %\bibliographystyle{SageH}
% %%Vancouver (numbered)
% %\bibliographystyle{SageV}
% \bibliography{<YourBibfile.bib>}
% \end{verbatim}
% and remember to add the relevant option to the \verb+\documentclass[]{sagej}+ line as listed in Table~\ref{T1}. 

% \begin{acks}
% This class file was developed by Sunrise Setting Ltd,
% Brixham, Devon, UK.\\
% Website: \url{http://www.sunrise-setting.co.uk}
% \end{acks}

% \subsection{End of paper special sections}
% Depending on the requirements of the journal that you are submitting to,
% there are macros defined to typeset various special sections.

% The commands available are:
% \begin{verbatim}
% \begin{acks}
% To typeset an
%   "Acknowledgements" section.
% \end{acks}
% \end{verbatim}

% \begin{verbatim}
% \begin{biog}
% To typeset an
%   "Author biography" section.
% \end{biog}
% \end{verbatim}

% \begin{verbatim}
% \begin{biogs}
% To typeset an
%   "Author Biographies" section.
% \end{biogs}
% \end{verbatim}

% %\newpage

% \begin{verbatim}
% \begin{dci}
% To typeset a "Declaration of
%   conflicting interests" section.
% \end{dci}
% \end{verbatim}

% \begin{verbatim}
% \begin{funding}
% To typeset a "Funding" section.
% \end{funding}
% \end{verbatim}

% \begin{verbatim}
% \begin{sm}
% To typeset a
%   "Supplemental material" section.
% \end{sm}
% \end{verbatim}

\appendix
\section{Appendix A: Index to Multimedia Extensions}
The relevant extension videos associated with this paper are shown in Table \ref{tab:multimedia}.

% \begin{table}[!ht]
% \small\sf\centering
% \caption{Multimedia Extensions\label{tab:multimedia}}
% % 将表格的行高整体拉伸为原来的 1.3 倍
% \renewcommand{\arraystretch}{1.3} 
% \begin{tabular}{c c m{7.5cm}} 
% \toprule
% Extension & Type & Description\\
% \midrule
% \texttt{1} & Video & Overview of the proposed system. \\ 
% \texttt{2} & Video & Mobile manipulation in constrained and dynamic environments. \\  
% \texttt{3} & Video & Reliability validation in long-horizon continuous tasks. \\  
% \texttt{4} & Video & Base-arm coordination for complex tasks with diverse end-effector constraints. \\ 
% \bottomrule
% \end{tabular}
% \end{table}

\begin{table}[!ht]
\small\sf\centering
\caption{Multimedia Extensions\label{tab:multimedia}}
\begin{tabular}{c c l}
\toprule
Extension & Type & Description\\
\midrule
\texttt{1} & Video & Overview of the proposed system. \\[4pt] 
% \midrule
\multirow{2}{*}{\texttt{2}} & \multirow{2}{*}{Video} & Mobile manipulation in constrained \\
 & & and dynamic environments. \\  [4pt]
% \midrule
\multirow{2}{*}{\texttt{3}} & \multirow{2}{*}{Video} & Reliability validation in long-horizon \\
 & & continuous tasks. \\  [4pt]
% \midrule
\multirow{3}{*}{\texttt{4}} & \multirow{3}{*}{Video} & Base-arm coordination for complex \\
 & & tasks with diverse end-effector \\  
 & & constraints. \\ 
\bottomrule
\end{tabular}
\end{table}

% \textbf{Extension 1: Overview of the proposed system.} 
% The proposed system demonstrates its capability to (i) sustain reliable continuous whole-body coordination across successive tasks in constrained and dynamic real-world environments, (ii) maintain high reliability in long-horizon tasks under target pose uncertainty, and (iii) generalize base–arm coordination to complex tasks with diverse end-effector constraints.

\section{Appendix B: Task, Feasibility and Safety Constraints}
\label{app:constraints}
\subsection{Task Satisfaction}
To ensure the end-effector follows each task trajectory, task constraints have to be introduced (Fig.~\ref{fig:trajectory_optimization}(v)). For each task $i,\forall i\in{1,\cdots,N_\mathcal{T}}$, we constrain the end-effector pose $f_{\mathrm{FK,ee}}(\boldsymbol{x}(\bar{t}_{\kappa_{i,\mathrm{s}}}+\tau))$ to match task end-effector trajectory $\bar{\boldsymbol{P}}_{\mathrm{t},i}(\tau)$ over the task duration $\tau\in[0, T_{\mathcal{T},i}]$:
\begin{equation}
\begin{aligned}
\mathcal{H}_{t_p,i}&(\boldsymbol{q}^{[s]}(t),\boldsymbol{T}) \\
&= \| [f_{\text{FK},\mathrm{ee}}^{-1}(\boldsymbol{x}(\bar{t}_{\kappa_{i,\mathrm{s}}}+\tau))\cdot\bar{\boldsymbol{P}}_{\mathrm{t},i}(\tau)]_{\boldsymbol{p}} \|_2^2,\\
&\quad \quad\forall \tau \in [0, T_{\mathcal{T},i}],\forall i \in \{1,\cdots,N_\mathcal{T}\}
\end{aligned}
\label{eq:task_pos}
\end{equation}
\begin{equation}
\begin{aligned}
&\mathcal{G}_{t_R,i}(\boldsymbol{q}^{[s]}(t),\boldsymbol{T}) \\
&\ = f_{d,\boldsymbol{R}}([\bar{\boldsymbol{P}}_{\mathrm{t},i}(\tau)]_{\boldsymbol{R}},[f_{\mathrm{FK,ee}}(\boldsymbol{x}(\bar{t}_{\kappa_{i,\mathrm{s}}}+\tau))]_{\boldsymbol{R}})-d_o\\
&\qquad \quad\forall \tau \in [0, T_{\mathcal{T},i}],\forall i \in \{1,\cdots,N_\mathcal{T}\}
\end{aligned}
\label{eq:task_ori}
\end{equation}
where $[\cdot]_{\boldsymbol{p}}:\text{SE}(3)\rightarrow \mathbb{R}^3$ and $[\cdot]_{\boldsymbol{R}}:\text{SE}(3)\rightarrow \text{SO}(3)$ extract the position and rotation component of the transformation matrix, respectively, and $d_o$ is the orientation tolerance. Finally, to allocate sufficient time for task execution, we enforce:
\begin{align}
    \mathcal{H}_{T_\mathcal{T},i}(\boldsymbol{q}^{[s]}(t),\boldsymbol{T})=\sum_{j=\kappa_{i,\mathrm{s}}}^{\kappa_{i,\mathrm{e}}-1} T_j-T_{\mathcal{T},i}, \quad \forall i.
\end{align}

Furthermore, for instant tasks (e.g., pick or place tasks) where the task start and end times coincide, i.e., $\bar{t}_{\kappa_{i,\mathrm{s}}} = \bar{t}_{\kappa_{i,\mathrm{e}}}$, it is crucial to constrain the end-effector's instantaneous velocity to ensure stable grasping or placing. We enforce linear and angular velocity constraints at the specific time instance $t = \bar{t}_{\kappa_{i,\mathrm{s}}}$:
\begin{equation}
\begin{aligned}
&\mathcal{G}_{v}(\boldsymbol{q}^{[s]}(t),\boldsymbol{T}) = \| \boldsymbol{J}_{v}(\boldsymbol{x}(t))\boldsymbol{v}_{\mathrm{mm}}(t) \|_2 - d_{\mathrm{ee},v}, \\
&\mathcal{G}_{\omega}(\boldsymbol{q}^{[s]}(t),\boldsymbol{T}) = \| \boldsymbol{J}_{\omega}(\boldsymbol{x}(t))\boldsymbol{v}_{\mathrm{mm}}(t) \|_2 - d_{\mathrm{ee},\omega} , \\
&\ \forall t = \bar{t}_{\kappa_{i,\mathrm{s}}}, \text{ if } \bar{t}_{\kappa_{i,\mathrm{s}}} = \bar{t}_{\kappa_{i,\mathrm{e}}}, \forall i \in \{1,\cdots,N_\mathcal{T}\}
\end{aligned}
\label{eq:task_vel_ineq}
\end{equation}
where $\boldsymbol{J}_{v}$ and $\boldsymbol{J}_{\omega}$ correspond to the linear and angular parts of the geometric Jacobian. $\boldsymbol{v}_{\mathrm{mm}}(t)$ denotes the generalized velocity vector of the mobile manipulator (including base and manipulator joints). The terms $d_{\mathrm{ee},v}$ and $d_{\mathrm{ee},\omega}$ represent the tolerance for linear and angular velocities, respectively.

\subsection{Dynamic Feasibility}

To ensure that the planned trajectory is dynamically feasible for the mobile manipulator, we impose wheel and joint constraints along the entire trajectory $t\in[0,\bar{t}_M]$ (Fig.~\ref{fig:trajectory_optimization}(vi)). For the differential-drive base, we bound the angular velocity and angular acceleration of the left and right wheels:
\begin{subequations}
	\begin{align}
        \mathcal{G} _{\omega _{l(r)}}\bigl(\boldsymbol{q}^{[s]}(t),\boldsymbol{T}\bigr)
        &= \omega_{l(r)}^2\bigl(\dot{\boldsymbol{q}}_{\mathfrak{b}},\ddot{\boldsymbol{q}}_{\mathfrak{b}}\bigr)
        -\omega _{w,\max}^{2},\\
        \mathcal{G} _{\alpha _{l(r)}}\bigl(\boldsymbol{q}^{[s]}(t),\boldsymbol{T}\bigr)
        &= \alpha_{l(r)}^2\bigl(\dot{\boldsymbol{q}}_{\mathfrak{b}},\ddot{\boldsymbol{q}}_{\mathfrak{b}},\dddot{\boldsymbol{q}}_{\mathfrak{b}}\bigr)
        -\alpha _{w,\max}^{2},
	\end{align}
	\label{eq:constrain_feas_base}
\end{subequations}
where $\omega _{w,\max}$ and $\alpha _{w,\max}$ denote the maximum wheel's angular velocity and acceleration, respectively. The wheel's angular velocity and acceleration are given by
\begin{equation}
    \omega _{l(r)}(\dot{\boldsymbol{q}}_{\mathfrak{b}},\ddot{\boldsymbol{q}}_{\mathfrak{b}})
        =\frac{1}{2r_w}\left( 2 \left\| \dot{\boldsymbol{q}}_{\mathfrak{b}} \right\| _2
        \underset{(+)}{-}d_w\,\omega _{\mathfrak{b}}(\dot{\boldsymbol{q}}_{\mathfrak{b}},\ddot{\boldsymbol{q}}_{\mathfrak{b}}) \right), 
\label{eq:base_wheel_omega}
\end{equation}
\begin{equation}
\begin{aligned}
    \alpha _{l(r)}&(\dot{\boldsymbol{q}}_{\mathfrak{b}},\ddot{\boldsymbol{q}}_{\mathfrak{b}},\dddot{\boldsymbol{q}}_{\mathfrak{b}})\\
        &=\frac{1}{2r_w}\left( 2 \frac{\dot{\boldsymbol{q}}_{\mathfrak{b}}^{\top}\ddot{\boldsymbol{q}}_{\mathfrak{b}}}{\left\| \dot{\boldsymbol{q}}_{\mathfrak{b}} \right\| _2}
        \underset{(+)}{-}d_w\,\alpha _{\mathfrak{b}}(\dot{\boldsymbol{q}}_{\mathfrak{b}},\ddot{\boldsymbol{q}}_{\mathfrak{b}},\dddot{\boldsymbol{q}}_{\mathfrak{b}}) \right),
\label{eq:base_wheel_alpha}
\end{aligned}
\end{equation}
\begin{equation}
    \omega _{\mathfrak{b}}(\dot{\boldsymbol{q}}_{\mathfrak{b}},\ddot{\boldsymbol{q}}_{\mathfrak{b}})
=\frac{\ddot{\boldsymbol{q}}_{\mathfrak{b}}^{\top}\mathbf{B}\dot{\boldsymbol{q}}_{\mathfrak{b}}}{\dot{\boldsymbol{q}}_{\mathfrak{b}}^{\top}\dot{\boldsymbol{q}}_{\mathfrak{b}}},
\end{equation}
\begin{equation}
\begin{aligned}
    \alpha _{\mathfrak{b}}&(\dot{\boldsymbol{q}}_{\mathfrak{b}},\ddot{\boldsymbol{q}}_{\mathfrak{b}},\dddot{\boldsymbol{q}}_{\mathfrak{b}})\\
    &=\frac{\dddot{\boldsymbol{q}}_{\mathfrak{b}}^{\top}\mathbf{B}\dot{\boldsymbol{q}}_{\mathfrak{b}}}{\dot{\boldsymbol{q}}_{\mathfrak{b}}^{\top}\dot{\boldsymbol{q}}_{\mathfrak{b}}}
        -\frac{2\left( \ddot{\boldsymbol{q}}_{\mathfrak{b}}^{\top}\mathbf{B}\dot{\boldsymbol{q}}_{\mathfrak{b}} \right) \left( \ddot{\boldsymbol{q}}_{\mathfrak{b}}^{\top}\dot{\boldsymbol{q}}_{\mathfrak{b}} \right)}{\left( \dot{\boldsymbol{q}}_{\mathfrak{b}}^{\top}\dot{\boldsymbol{q}}_{\mathfrak{b}} \right) ^2},
\end{aligned}
\end{equation}
with $r_w$ the wheel radius and $d_w$ the distance between the two wheels. In our experiments, $r_w=0.06\,m$ and $d_w=0.2624\,m$. The operator $\underset{(+)}{-}$ indicates that the minus sign is used for the left wheel~$l$ and the plus sign for the right wheel~$r$.
\begin{equation}
    \mathbf{B}:=\begin{bmatrix}
    	0&		-1\\
    	1&		0
    \end{bmatrix}
\end{equation}
% , \qquad
% \bar{\mathbf{B}}:=\begin{bmatrix}
% 	0&		1\\
% 	1&		0
% \end{bmatrix}
is an auxiliary matrix introduced for notational convenience.

Moreover, when $\left\| \dot{\boldsymbol{q}}_{\mathfrak{b}} \right\|_2=0$, the expressions for wheel angular velocity and acceleration become ill-defined, effectively leading to unbounded wheel rates. To avoid this singularity, we enforce a lower bound on the base translational velocity:
\begin{equation}
	\mathcal{G} _{\bar{v}}\bigl(\boldsymbol{q}^{[s]}(t),\boldsymbol{T}\bigr)
	= v_{\min}^{2}-\dot{\boldsymbol{q}}_{\mathfrak{b}}^{\top}\dot{\boldsymbol{q}}_{\mathfrak{b}},
    \label{eq:constrain_feas_min_vel}
\end{equation}
where $v_{\min}\in\mathbb{R}_{>0}$ is a small positive constant.

For each joint $l$ of the manipulator, where $l\in\{1,\dots,\mathcal{L}\}$, we enforce joint position, velocity, and acceleration limits:
\begin{subequations}
	\begin{align}
        \mathcal{G} _{\mathfrak{q} _{\theta}}\bigl(\boldsymbol{q}^{[s]}(t),\boldsymbol{T}\bigr)&=q_l-q_{l,\max},\\
    	\mathcal{G} _{\mathfrak{q} _{\bar{\theta}}}\bigl(\boldsymbol{q}^{[s]}(t),\boldsymbol{T}\bigr)&=q_{l,\min}-q_l,\\
    	\mathcal{G} _{\mathfrak{q} _{\omega}}\bigl(\boldsymbol{q}^{[s]}(t),\boldsymbol{T}\bigr)&=\dot{q}_{l}^{2}-\omega _{l,\max}^{2},\\
        \mathcal{G} _{\mathfrak{q} _{\alpha}}\bigl(\boldsymbol{q}^{[s]}(t),\boldsymbol{T}\bigr)&=\ddot{q}_{l}^{2}-\alpha _{l,\max}^{2},
	\end{align}
	\label{eq:constrain_feas_mani}
\end{subequations}
where $q_{l,\min}$ and $q_{l,\max}$ denote the minimum and maximum joint angles of joint $l$, and $\omega _{l,\max}$ and $\alpha _{l,\max}$ denote the maximum joint angular velocity and angular acceleration, respectively.

\subsection{Collision Avoidance}
\label{subsec:collision_avoidance}

To ensure the safety of the mobile manipulator in complex 3D environments, we enforce collision avoidance constraints between the robot and obstacles (Fig.~\ref{fig:trajectory_optimization}(vii)). We approximate the robot's geometry by representing each link $l \in \{0, \dots, \mathcal{L}\}$ (where link $0$ denotes the mobile base) as a set of $m_l$ collision spheres. For the $j$-th sphere on link $l$ with radius $r_l$, its center position $\boldsymbol{p}_{l,j}(t)$ in the world frame is computed via forward kinematics:
\begin{equation}
    \boldsymbol{p}_{l,j}(t) = [f_{\mathrm{FK},l}(\boldsymbol{x}(t)) \cdot \,^l\boldsymbol{p}_{l,j}]_{\boldsymbol{p}},
\end{equation}
where $^l\boldsymbol{p}_{l,j}$ is the position of the sphere center relative to the link frame $l$.

The environmental safety is maintained using an ESDF map. To ensure safety, we impose a constraint requiring each collision sphere to maintain a minimum distance of $d_\mathrm{s}$ from the environment:
\begin{equation}
\begin{aligned}
    \mathcal{G}_{\mathrm{col},l,j}&(\boldsymbol{q}^{[s]}(t),\boldsymbol{T}) = r_l + d_\mathrm{s} - D_{\mathrm{ESDF}}(\boldsymbol{p}_{l,j}(t)), \\
    &\qquad\forall\, l \in \{0, \dots, \mathcal{L}\}, \forall\, j \in \{1, \dots, m_l\},
\end{aligned}
\label{eq:constrain_env_collision}
\end{equation}

To avoid collision with dynamic obstacles, we approximate each moving obstacle as a 2D cylinder with radius $r_{\mathrm{dy}}$. The obstacle's future position, denoted as $\boldsymbol{p}_{\mathrm{dy},i}(t)$, is provided by the perception module as a predicted trajectory, where $t$ represents the time elapsed relative to the start of the planned trajectory. Specifically, in this work, we estimate the obstacle's initial position and velocity, and derive $\boldsymbol{p}_{\mathrm{dy},i}(t)$ under the assumption of constant linear velocity. The collision avoidance constraint is enforced as:
\begin{equation}
\begin{aligned}
    \mathcal{G}_{\mathrm{dy},l,j,i}&(\boldsymbol{q}^{[s]}(t),\boldsymbol{T}) = r_l + r_{\mathrm{dy}}+ d_\mathrm{s} \\
    &\qquad\qquad\quad - \|\big[\boldsymbol{p}_{l,j}(t)\big]_{\mathrm{xy}}-\boldsymbol{p}_{\mathrm{dy},i}(t)\|_2, \\
    &\forall\, l \in \{0, \dots, \mathcal{L}\}, \forall\, j \in \{1, \dots, m_l\}, \forall\,i.
\end{aligned}
\label{eq:constrain_dyn_collision}
\end{equation}

\subsection{Self-Collision Avoidance}

To prevent self-collisions (Fig.~\ref{fig:trajectory_optimization}(vii)), we impose constraints that maintain a minimum distance of $d_{\mathrm{self}}$ between all pairs of collision spheres belonging to different links:
\begin{equation}
\begin{aligned}
    \mathcal{G}_{\mathrm{self}}&(\boldsymbol{q}^{[s]}(t),\boldsymbol{T}) 
    = r_{l^\prime}+r_l+d_{\mathrm{self}}
    \\&\qquad \qquad \quad \ -\left\|\,^{l^\prime}\boldsymbol{p}_{l^\prime,i}
      -\,^{l^\prime}\boldsymbol{T}_l(\boldsymbol{q}_\mathfrak{m})\,^l\boldsymbol{p}_{l,j}\right\|_2 ,\\
    &\forall\, l^\prime\in \{0,\dots ,l-1\},\ \forall\, l\in \{1,\dots ,\mathcal{L}\},\\
    &\forall\, i\in \{1,\dots ,m_{l^\prime}\},\ \forall\, j\in \{1,\dots ,m_{l}\},
\end{aligned}
\label{eq:constrain_self}
\end{equation}
where $\,^{l^\prime}\boldsymbol{p}_{l^\prime,i}$ is the $i$-th collision point on link $l^\prime$ expressed in frame $l^\prime$, $\,^{l}\boldsymbol{p}_{l,j}$ is the $j$-th collision point on link $l$ expressed in frame $l$, and $\,^{l^\prime}\boldsymbol{T}_l(\boldsymbol{q}_\mathfrak{m})$ is the homogeneous transform from frame $l$ to frame $l^\prime$ obtained from the mobile manipulator forward kinematics.

\section{Appendix C: Visibility Constraints}
\label{app:visibility}
In this section, we provide the detailed formulations for the visibility constraints $\mathcal{G}_{g,i}(\boldsymbol{q}^{[s]}(t),\boldsymbol{T},\boldsymbol{T}_{\mathrm{p}})$ mentioned in the main text, where $g \in \{\mathfrak{v}_a, \mathfrak{v}_d, \mathfrak{v}_o\}$. For a task necessitating task pose estimation $i \in \mathcal{I}_{\text{perc}}$, let $\boldsymbol{p}_{\mathrm{tgt},i} := [{\boldsymbol{P}}_{\mathrm{t},i}(0)]_{\boldsymbol{p}} \in \mathbb{R}^3$ denote the position of the target to be observed. We define the camera center position $\boldsymbol{p}_\mathrm{c}(t)$ and the camera optical axis direction $\boldsymbol{z}_\mathrm{c}(t)$ using the forward kinematics function for camera pose $f_{\mathrm{FK,c}}(\cdot)$ associated with the robot state $\boldsymbol{x}(t)$:
\begin{equation*}
    \boldsymbol{p}_\mathrm{c}(t) := [f_{\mathrm{FK,c}}(\boldsymbol{x}(t))]_{\boldsymbol{p}}, \quad
    \boldsymbol{z}_\mathrm{c}(t) := [f_{\mathrm{FK,c}}(\boldsymbol{x}(t))]_{\boldsymbol{z}}.
\end{equation*}
The vector from the camera center to the target is defined as $\boldsymbol{v}_{ct,i}(t):= \boldsymbol{p}_{\mathrm{tgt},i} - \boldsymbol{p}_\mathrm{c}(t)$. The constraints below are enforced over the perception interval $t \in [\bar{t}_{\kappa_{i,\mathrm{s}}}-T_{\mathrm{p},i},\, \bar{t}_{\kappa_{i,\mathrm{s}}})$.

\subsection{Field-of-View (FOV) Constraint}
To ensure the target projects onto the camera image plane, we approximate the camera's field of view as a cone with a half-angle of $\theta_{\mathrm{fov}}$ (derived from the camera intrinsics with a safety margin). Consequently, the angle between the camera optical axis $\boldsymbol{z}_\mathrm{c}(t)$ and the target vector $\boldsymbol{v}_{ct,i}(t)$ must be within this limit. The constraint is formulated as:
\begin{equation}
    \mathcal{G}_{\mathfrak{v}_a,i}(\boldsymbol{q}^{[s]}(t),\boldsymbol{T},\boldsymbol{T}_{\mathrm{p}}) = \cos(\theta_{\mathrm{fov}}) - \frac{\boldsymbol{z}_\mathrm{c}(t)^\top \boldsymbol{v}_{ct,i}(t)}{\|\boldsymbol{v}_{ct,i}(t)\|_2}.
\end{equation}

\subsection{Maximum Sensing Range Constraint}
To guarantee reliable depth measurement, the distance to the target must not exceed the maximum effective range, denoted by $d_{\max}$. This is enforced by:
\begin{equation}
    \mathcal{G}_{\mathfrak{v}_d,i}(\boldsymbol{q}^{[s]}(t),\boldsymbol{T},\boldsymbol{T}_{\mathrm{p}}) = \|\boldsymbol{v}_{ct,i}(t)\|_2^2 - d_{\max}^2.
\end{equation}

\subsection{Occlusion-free Constraint}
To ensure the line-of-sight (LoS) from the camera to the target is not occluded by obstacles, we approximate the viewing cone using a set of $K_v$ spheres positioned along the target vector $\boldsymbol{v}_{ct,i}(t)$.
For each $k \in \{1, \dots, K_v-1\}$, the center of the $k$-th sphere is given by:
\begin{equation}
    \boldsymbol{p}_{k,i}(t) = \boldsymbol{p}_\mathrm{c}(t) + \frac{k}{K_v} \boldsymbol{v}_{ct,i}(t).
\end{equation}
The radius of the spheres increases linearly from the camera to the target to conservatively cover the LoS, defined as $r_{k,i} = \frac{k}{K_v} r_{\mathrm{tgt},i}$, where $r_{\mathrm{tgt},i}$ is the bounding radius of the target region. The occlusion-free requirement is enforced using the ESDF:
\begin{equation}
    \mathcal{G}_{\mathfrak{v}_o,i,k}(\boldsymbol{q}^{[s]}(t),\boldsymbol{T},\boldsymbol{T}_{\mathrm{p}}) = r_{k,i} - D_{\mathrm{ESDF}}(\boldsymbol{p}_{k,i}(t)), \ \forall k.
\end{equation}
This formulation ensures that every sphere along the LoS maintains a safe clearance from the environment.

\section{Appendix D: Sampling Strategy for $\mathcal{P}_{\mathrm{cmz},i}$}
\label{app:cmz}

The set $\mathcal{P}_{\mathrm{cmz},i}(\tau)$ is designed to approximate the local neighborhood of task-pose perturbations that may arise during execution due to uncertainty in perception, control, and target motion. Its role is to encode the desired compensation margin around the nominal task pose. The specific sampling scheme is not unique and can be adapted to different robots, tasks, or uncertainty models. Here we describe the sampling strategy used in our implementation.

Given the nominal task end-effector pose $\bar{\boldsymbol{P}}_{\mathrm{t},i}(\tau)$, with position $\bar{\boldsymbol{p}}\in\mathbb{R}^3$ and orientation $\bar{\boldsymbol{R}}\in\mathrm{SO}(3)$, we construct $\mathcal{P}_{\mathrm{cmz},i}(\tau)$ by sampling perturbations in both position and orientation.

\paragraph{Position sampling.}
To model translational uncertainty, we sample positions on a sphere of radius $r_{\mathrm{cmz}}$ centered at $\bar{\boldsymbol{p}}$. Using spherical coordinates, the sampled positions are defined as
\begin{equation}
\begin{aligned}
\boldsymbol{p}_{j,k}
=
\bar{\boldsymbol{p}}
+
r_{\mathrm{cmz}}
\begin{bmatrix}
\cos\theta_k \sin\phi_j \\
\sin\theta_k \sin\phi_j \\
\cos\phi_j
\end{bmatrix},
\\
j=0,\dots,N_{\phi},\;\;
k=0,\dots,N_{\theta},
\end{aligned}
\end{equation}
where
\[
\phi_j=j\Delta\phi,\qquad
\theta_k=k\Delta\theta,
\]
and the index bounds are
\[
N_{\phi}=\left\lfloor \frac{2\pi}{\Delta\phi}\right\rfloor,
\qquad
N_{\theta}=\left\lfloor \frac{\pi}{\Delta\theta}\right\rfloor.
\]
Here, $r_{\mathrm{cmz}}$ determines the translational compensation margin, while $\Delta\phi$ and $\Delta\theta$ control the angular sampling resolution.

\paragraph{Orientation sampling.}
To capture practically relevant variations around the nominal task pose while keeping the sampling tractable, we first apply a fixed tilt offset $\delta_{\mathrm{tilt}}$ about the local $x$-axis:
\begin{equation}
\boldsymbol{R}'=\bar{\boldsymbol{R}}\boldsymbol{R}_{x}(\delta_{\mathrm{tilt}}),
\end{equation}
where $\boldsymbol{R}_{x}(\cdot)$ denotes the elemental rotation matrix about the $x$-axis. We then generate rotational variations by rotating $\boldsymbol{R}'$ around the nominal approach axis
\[
\boldsymbol{z}_{\mathrm{t}}=[\bar{\boldsymbol{R}}]_{\boldsymbol{z}}
\]
using discrete angles $\psi_m=m\Delta\psi$:
\begin{equation}
\boldsymbol{R}_{m}=\mathrm{Rot}(\boldsymbol{z}_{\mathrm{t}},\psi_m)\boldsymbol{R}',\quad m=0,\dots,M,
\end{equation}
where
\[
M=\left\lfloor \frac{2\pi}{\Delta\psi}\right\rfloor,
\]
and $\mathrm{Rot}(\boldsymbol{n},\theta)$ denotes the rotation matrix corresponding to a rotation of angle $\theta$ about axis $\boldsymbol{n}$. The parameter $\delta_{\mathrm{tilt}}$ specifies the magnitude of the nominal tilt perturbation, while $\Delta\psi$ controls the resolution of rotational sampling about the approach direction.

\paragraph{Valid CMZ samples.}
Each sampled pose is formed by combining a sampled position with a sampled orientation. Any sample that results in a collision with the environment is discarded. The remaining collision-free poses constitute the set $\mathcal{P}_{\mathrm{cmz},i}(\tau)$.

\paragraph{Adaptive margin reduction.}
In some workspace configurations, the reachable-set intersection induced by the requested compensation margin may be empty. This indicates that the prescribed translational margin $r_{\mathrm{cmz}}$ is too large to be supported kinematically at the current task pose. In such cases, we iteratively reduce $r_{\mathrm{cmz}}$ and reconstruct $\mathcal{P}_{\mathrm{cmz},i}(\tau)$ until the corresponding intersection set becomes nonempty. The resulting radius can be interpreted as the largest feasible local compensation margin under the current workspace constraints.

The parameters $(r_{\mathrm{cmz}},\delta_{\mathrm{tilt}},\Delta\phi,\Delta\theta,\Delta\psi)$ determine the size and resolution of the sampled task-pose neighborhood, and therefore control the trade-off between robustness and conservativeness of the resulting Compensation Margin Zone.

% \subsection{Numerical Approximation of the Intersection}
% To efficiently compute the intersection of the reachability regions for all poses in $\mathcal{P}_{\mathrm{cmz},i}(\tau)$, we employ a sampling-based approach. Let $\mathcal{E}_0$ be the reachability ellipse computed for the first sampled pose. We generate a set of candidate base positions $\mathcal{X}_{\mathrm{cand}}$ by densely sampling the interior of $\mathcal{E}_0$. We then filter this set to retain only the points that lie within the reachability ellipses of all other sampled poses in $\mathcal{P}_{\mathrm{cmz},i}(\tau)$. The final CMZ ellipse $\mathcal{E}_{\mathrm{cmz},i}(\tau)$ is obtained by fitting the minimum volume enclosing ellipse to the surviving intersection points.

\section{Appendix E: Controller Formulation}
\label{app:cascaded_mpc}
We adopt a cascaded MPC architecture (Fig.~\ref{fig:controller}A) to achieve high-rate reactivity. At each control cycle, a mobile-base MPC first computes an optimal predicted base trajectory by minimizing the tracking error relative to the global trajectory, with collision avoidance constraints enforced. Building on this optimal predicted base trajectory, a manipulator MPC then optimizes the joint control input to track the global trajectory and compensate for task error during the task-critical phase while preserving safe interaction motion. Both MPC modules operate in a receding-horizon manner, where only the first control input from each optimized result is executed.

\subsection{State and Input}
Define the whole-body control input as $\boldsymbol{u}=[\boldsymbol{u}_\mathfrak{b}^\top,\boldsymbol{u}_\mathfrak{m}^\top]^\top$, where $\boldsymbol{u}_\mathfrak{b}=[v_\mathfrak{b},\omega_{\mathfrak{b}}]^\top$ denotes the mobile-base commands consisting of the linear velocity $v_\mathfrak{b}$ and the angular velocity $\omega_{\mathfrak{b}}$, and $\boldsymbol{u}_\mathfrak{m}=\dot{\boldsymbol{q}}_\mathfrak{m}$ denotes the manipulator joint velocity command. Despite the robot receiving velocity-level commands $\boldsymbol{u}$, directly optimizing $\boldsymbol{u}$ in MPC typically yields piecewise-constant velocity profiles that abruptly change at each control cycle, leading to impulsive accelerations and poor dynamic feasibility. To obtain continuous and dynamically feasible executable commands $\boldsymbol{u}$ in the implementation of receding-horizon, we adopt an extended model by augmenting the nominal state $\boldsymbol{x}$ presented in Sec.~\nameref{subsubsec:traj_rep} to $\bar{\boldsymbol{x}}=[\bar{\boldsymbol{x}}_{\mathfrak{b}}^\top,\bar{\boldsymbol{x}}_\mathfrak{m}^\top]^\top$ \cite{heins2023keep}, where $\bar{\boldsymbol{x}}_{\mathfrak{b}}=[q_x,q_y,\psi,v_\mathfrak{b},\omega_{\mathfrak{b}}]^\top$ and $\bar{\boldsymbol{x}}_\mathfrak{m}=[{\boldsymbol{q}}_\mathfrak{m}^\top,\dot{\boldsymbol{q}}_\mathfrak{m}^\top]^\top$. Accordingly, the MPC decision variable is defined as $\bar{\boldsymbol{u}}=[\bar{\boldsymbol{u}}_{\mathfrak{b}}^\top,\bar{\boldsymbol{u}}_\mathfrak{m}^\top]^\top$, where $\bar{\boldsymbol{u}}_\mathfrak{b}=[a_\mathfrak{b},\alpha_{\mathfrak{b}}]^\top$ consists of the linear acceleration $a_\mathfrak{b}$ and the angular acceleration $\alpha_{\mathfrak{b}}$, and $\bar{\boldsymbol{u}}_\mathfrak{m}=\ddot{\boldsymbol{q}}_\mathfrak{m}$. This dynamic extension yields velocity-level commands $\boldsymbol{u}$ that are continuous, while enabling direct enforcement of acceleration bounds. Finally, define $\bar{\boldsymbol{x}}^{\mathrm{ref}}(t)=[\bar{\boldsymbol{x}}^{\mathrm{ref}}_{\mathfrak{b}}(t)^\top,\bar{\boldsymbol{x}}^{\mathrm{ref}}_{\mathfrak{m}}(t)^\top]^\top$ and $\bar{\boldsymbol{u}}^{\mathrm{ref}}(t)=[\bar{\boldsymbol{u}}^{\mathrm{ref}}_{\mathfrak{b}}(t)^\top,\bar{\boldsymbol{u}}^{\mathrm{ref}}_{\mathfrak{m}}(t)^\top]^\top$ as the reference state and control input derived from trajectory $\boldsymbol{x}(t)$.

\subsection{Mobile Base MPC}
\label{subsubsec:base_mpc}
The mobile base MPC computes the optimal mobile base input to track the reference $\boldsymbol{x}^{\mathrm{ref}}_{\mathfrak{b}}(t)$ while avoiding dynamic obstacles. The optimal control problem is formulated as:
\begin{equation}
\begin{aligned}
	\min_{\bar{\boldsymbol{U}}_{\mathfrak{b}}}\quad &		
		\sum_{k=0}^{N-1} l_{\mathfrak{b}}\big( \bar{\boldsymbol{x}}_{\mathfrak{b},k},\bar{\boldsymbol{u}}_{\mathfrak{b},k} \big)
		+ l_{\mathfrak{b},N}\big( \bar{\boldsymbol{x}}_{\mathfrak{b},N} \big) \\
	\mathrm{s.t.}\quad&		
		\bar{\boldsymbol{x}}_{\mathfrak{b},0} = \bar{\boldsymbol{x}}^{\mathrm{meas}}_{\mathfrak{b}},\\
	&	\bar{\boldsymbol{x}}_{\mathfrak{b},k+1}
		= \boldsymbol{f}_{\mathfrak{b}}\big( \bar{\boldsymbol{x}}_{\mathfrak{b},k},\bar{\boldsymbol{u}}_{\mathfrak{b},k} \big),
		\quad \forall k=0,\dots,N-1,\\
	&	\mathcal{X}_{\mathfrak{b}}^{-} \leqslant \bar{\boldsymbol{x}}_{\mathfrak{b},k} \leqslant \mathcal{X}_{\mathfrak{b}}^{+},
		\quad \forall k=0,\dots,N,\\
	&	\mathcal{U}_{\mathfrak{b}}^{-} \leqslant \bar{\boldsymbol{u}}_{\mathfrak{b},k} \leqslant \mathcal{U}_{\mathfrak{b}}^{+},
		\quad \forall k=0,\dots,N-1,
\end{aligned}
\label{eq:mobile_base_ocp}
\end{equation}
where $\bar{\boldsymbol{x}}^{\mathrm{meas}}_{\mathfrak{b}}$ is the current measured state, $\bar{\boldsymbol{U}}_{\mathfrak{b}}=\{\bar{\boldsymbol{u}}_{\mathfrak{b},0},\cdots,\bar{\boldsymbol{u}}_{\mathfrak{b},N-1}\}$ is the set of decision variables for the input of the mobile-base. $\boldsymbol{f}_{\mathfrak{b}}$ represents the differential drive dynamics. The sets $\mathcal{X}_{\mathfrak{b}}^{\pm}$ and $\mathcal{U}_{\mathfrak{b}}^{\pm}$ define admissible state and input bounds, respectively. The cost function is defined as:
\begin{equation}
\begin{aligned}
l_{\mathfrak{b}}\left( \bar{\boldsymbol{x}}_{\mathfrak{b} ,k},\boldsymbol{u}_{\mathfrak{b} ,k} \right) =&\left\| \bar{\boldsymbol{x}}_{\mathfrak{b} ,k}-\boldsymbol{x}_{\mathfrak{b}}^{\mathrm{ref}}(t_k) \right\| _{\boldsymbol{Q}_{\mathfrak{b}}}\\
&+\left\| \bar{\boldsymbol{u}}_{\mathfrak{b} ,k}-\boldsymbol{u}_{\mathfrak{b}}^{\mathrm{ref}}(t_k) \right\| _{\boldsymbol{R}_{\mathfrak{b}}}\\
&+l_{\mathfrak{b} ,\mathrm{dy}}\left( \bar{\boldsymbol{x}}_{\mathfrak{b} ,k} \right),
\end{aligned}
\end{equation}
\begin{equation}
\begin{aligned}
l_{\mathfrak{b},N}\big( \bar{\boldsymbol{x}}_{\mathfrak{b},N} \big)=&\left\| \bar{\boldsymbol{x}}_{\mathfrak{b} ,N}-\boldsymbol{x}_{\mathfrak{b}}^{\mathrm{ref}}(t_N) \right\| _{\boldsymbol{Q}_{\mathfrak{b},N}},
\end{aligned}
\end{equation}
where $\boldsymbol{Q}_{\mathfrak{b}},\boldsymbol{Q}_{\mathfrak{b},N}, \boldsymbol{R}_{\mathfrak{b}} \succeq 0$ are weighting matrices. The term $l_{\mathfrak{b},\mathrm{dy}}$ enforces collision avoidance with dynamic obstacles. Each dynamic obstacle is modeled as 2D cylinders with position $\boldsymbol{p}_{\mathrm{dy},i}(t)\in\mathbb{R}^2$ with radius $r_{\mathrm{dy}}$. The mobile base is modeled as $N_{\mathfrak{b}}$ collision spheres with radius $r_0$. $\boldsymbol{p}_{0,m}(\bar{\boldsymbol{x}}_{\mathfrak{b}})$ returns the position of the $m$-th collision sphere when the mobile base at state $\bar{\boldsymbol{x}}_{\mathfrak{b}}$. $d_{\mathrm{s},0}=r_0+r_{\mathrm{dy}}+d_\mathrm{s}$ as the required safe distance between the collision sphere and the dynamic obstacle. Define $D_{\mathrm{dy},m,i}(\bar{\boldsymbol{x}}_{\mathfrak{b}}):=d_{\mathrm{s},0}-\big\| \big[\boldsymbol{p}_{0,m}(\bar{\boldsymbol{x}}_{\mathfrak{b}})\big]_{\mathrm{xy}} -\boldsymbol{p}_{\mathrm{dy},i}(t_k) \big\|$. $l_{\mathfrak{b},\mathrm{dy}}$ is defined as:
\begin{equation}
l_{\mathfrak{b} ,\mathrm{dy}}\left( \bar{\boldsymbol{x}}_{\mathfrak{b} ,k} \right)
= \omega _{\mathrm{dy}}\sum_{i}\sum_{m=1}^{N_{\mathfrak{b}}} \max \Big( 0,\,
D_{\mathrm{dy},m,i}(\bar{\boldsymbol{x}}_{\mathfrak{b},k}) \Big)^2.
\end{equation}

Solving Eq.~\eqref{eq:mobile_base_ocp} yielded the optimal state trajectory for the mobile base, denoted as $\bar{\boldsymbol{X}}_{\mathfrak{b}}^{*}=\{\bar{\boldsymbol{x}}_{\mathfrak{b},0}^{*},\cdots,\bar{\boldsymbol{x}}_{\mathfrak{b},N}^{*}\}$.

\subsection{Manipulator MPC}
\label{subsubsec:mani_mpc}
Conditioned on the optimal base motion $\bar{\boldsymbol{X}}_{\mathfrak{b}}^{*}$, the manipulator MPC optimizes the best joint input. The manipulator MPC is formulated as:
\begin{equation}
\begin{aligned}
	\min_{\bar{\boldsymbol{U}}_{\mathfrak{m}}}\quad&
		\sum_{k=0}^{N-1} l_{\mathfrak{m}}\big( \bar{\boldsymbol{x}}_{\mathfrak{b},k}^{*},\bar{\boldsymbol{x}}_{\mathfrak{m},k},\bar{\boldsymbol{u}}_{\mathfrak{m},k} \big)
		+ l_{\mathfrak{m},N}\big( \bar{\boldsymbol{x}}_{\mathfrak{b},N}^{*},\bar{\boldsymbol{x}}_{\mathfrak{m},N} \big) \\
	\mathrm{s.t.}\quad&
		\bar{\boldsymbol{x}}_{\mathfrak{m},0} = \bar{\boldsymbol{x}}^{\mathrm{meas}}_{\mathfrak{m}},\\
	&	\bar{\boldsymbol{x}}_{\mathfrak{m},k+1}
		= \boldsymbol{f}_{\mathfrak{m}}\big( \bar{\boldsymbol{x}}_{\mathfrak{m},k},\bar{\boldsymbol{u}}_{\mathfrak{m},k} \big), \quad \forall k=0,\dots,N-1,\\
	&	\mathcal{X}_{\mathfrak{m}}^{-} \leqslant \bar{\boldsymbol{x}}_{\mathfrak{m},k} \leqslant \mathcal{X}_{\mathfrak{m}}^{+}, \quad \forall k=0,\dots,N\\
	&	\mathcal{U}_{\mathfrak{m}}^{-} \leqslant \bar{\boldsymbol{u}}_{\mathfrak{m},k} \leqslant \mathcal{U}_{\mathfrak{m}}^{+}, \quad \forall k=0,\dots,N-1.
\end{aligned}
\label{eq:manipulator_ocp}
\end{equation}
$\bar{\boldsymbol{x}}^{\mathrm{meas}}_{\mathfrak{m}}$ is the measured current state. The function $\boldsymbol{f}_{\mathfrak{m}}$ represents the manipulator dynamics. The sets $\mathcal{X}_{\mathfrak{m}}^{\pm}$ and $\mathcal{U}_{\mathfrak{m}}^{\pm}$ impose manipulator state and input bounds. The manipulator MPC employs a hybrid cost function that integrates smooth switching between whole-body trajectory tracking and end-effector task error compensation with dynamic collision avoidance, expressed as:
\begin{equation}
\begin{aligned}
l_{\mathfrak{m}}\big( \bar{\boldsymbol{x}}_{\mathfrak{b},k}^{*},\bar{\boldsymbol{x}}_{\mathfrak{m},k},\boldsymbol{u}_{\mathfrak{m},k} \big) =& \sigma_{\mathrm{s}}(t_k)\,l_{\mathrm{track}}(\bar{\boldsymbol{x}}_{\mathfrak{b},k}^{*},\bar{\boldsymbol{x}}_{\mathfrak{m},k},\boldsymbol{u}_{\mathfrak{m},k}) \\
&+ \sigma_{\mathrm{ee}}(t_k)\,l_{\mathrm{task}}(\bar{\boldsymbol{x}}_{\mathfrak{b},k}^{*},\bar{\boldsymbol{x}}_{\mathfrak{m},k}) \\
&+ l_{\mathfrak{m},\mathrm{dy}}(\bar{\boldsymbol{x}}_{\mathfrak{b},k}^{*},\bar{\boldsymbol{x}}_{\mathfrak{m},k}).
\label{eq:supp_mani_cost}
\end{aligned}
\end{equation}
% \begin{equation}
% \begin{aligned}
% l_{\mathfrak{m},N}\big( \bar{\boldsymbol{x}}_{\mathfrak{b},k}^{*},\bar{\boldsymbol{x}}_{\mathfrak{m},k} \big) =& \sigma_{\mathrm{s}}(t_k)\,l_{\mathrm{track}}(\bar{\boldsymbol{x}}_{\mathfrak{b},k}^{*},\bar{\boldsymbol{x}}_{\mathfrak{m},k},\boldsymbol{u}_{\mathfrak{m},k}) 
% \end{aligned}
% \end{equation}
\begin{equation}
\begin{aligned}
    l_{\mathrm{track}}(\bar{\boldsymbol{x}}_{\mathfrak{b},k}^{*},\bar{\boldsymbol{x}}_{\mathfrak{m},k},\boldsymbol{u}_{\mathfrak{m},k})=&\big\| \bar{\boldsymbol{x}}_{\mathfrak{m},k} - \bar{\boldsymbol{x}}_{\mathfrak{m}}^{\mathrm{ref}}(t_k) \big\|_{\boldsymbol{Q}_{\mathfrak{m}}}\\ 
    &+ \big\| \bar{\boldsymbol{u}}_{\mathfrak{m},k} - \bar{\boldsymbol{u}}_{\mathfrak{m}}^{\mathrm{ref}}(t_k) \big\|_{\boldsymbol{R}_{\mathfrak{m}}}.
\end{aligned}
\end{equation}
penalizes deviations from the global trajectory, where $\boldsymbol{Q}_{\mathfrak{m}}, \boldsymbol{R}_{\mathfrak{m}} \succeq 0$ are weighting matrices.
\begin{equation}
\begin{aligned}
l_{\mathrm{task}}(\bar{\boldsymbol{x}}_{\mathfrak{b},k}^{*},\bar{\boldsymbol{x}}_{\mathfrak{m},k}) = &\big\| [\boldsymbol{e}_{\mathrm{ee},k}]_{\boldsymbol{p}} \big\|_{\boldsymbol{Q}_{\mathrm{p}}} \\
&- \omega_{R} \cdot \mathrm{tr}\big([\boldsymbol{e}_{\mathrm{ee},k}]_{\boldsymbol{R}}\big),
\end{aligned}
\end{equation}
penalizes the error between the current end-effector pose $\boldsymbol{P}_{\mathrm{ee},k} = f_{\mathrm{FK},\mathrm{ee}}(\bar{\boldsymbol{x}}_{\mathfrak{b},k}^{*}, \bar{\boldsymbol{x}}_{\mathfrak{m},k})$ and the warped reference task trajectory $\boldsymbol{P}_{\mathrm{ee},k}^{\mathrm{warp}}=\boldsymbol{P}_{\mathrm{ee}}^{\mathrm{warp}}(t_k)$ (Eq.~\eqref{eq:warping_operator}), where $\boldsymbol{e}_{\mathrm{ee},k} = \boldsymbol{P}_{\mathrm{ee},k}^{-1} \boldsymbol{P}^{\mathrm{warp}}_{\mathrm{ee},k}$ represents the pose error. $\boldsymbol{Q}_{\mathrm{p}}\succeq 0$ and $\omega_{R}\geq0$ are position and orientation weight. The smooth transition between $l_{\mathrm{track}}$ and $l_{\mathrm{task}}$ is governed by complementary switching parameters $\sigma_{\mathrm{s}},\sigma_{\mathrm{ee}}\in[0,1]$ (Eq.~\eqref{eq:smoo_switch}).

Similar to the base MPC, $l_{\mathfrak{m},\mathrm{dy}}$ penalizes the proximity of the manipulator links to dynamic obstacles. We approximate the collision model of the manipulator as a set of collision spheres with $r_{j,m}$ the radius of the $m$-th sphere on link $j$. $\boldsymbol{p}_{j,m}(\bar{\boldsymbol{x}}_{\mathfrak{b}}, \bar{\boldsymbol{x}}_{\mathfrak{m}})$ denotes the position of the $m$-th collision sphere on link $j$ in world frame at state $(\bar{\boldsymbol{x}}_{\mathfrak{b}}, \bar{\boldsymbol{x}}_{\mathfrak{m}})$ using forward kinematics. Define $d_{\mathrm{s},j,m}=r_{j,m}+r_{\mathrm{dy}}+d_\mathrm{s}$, $D_{\mathrm{dy},j,m,i}(\bar{\boldsymbol{x}}_{\mathfrak{m},k}^{*},\bar{\boldsymbol{x}}_{\mathfrak{b},k})=d_{\mathrm{s},j,m}-\big\| \big[\boldsymbol{p}_{j,m}(\bar{\boldsymbol{x}}_{\mathfrak{b},k}^{*}, \bar{\boldsymbol{x}}_{\mathfrak{m},k})\big]_{\mathrm{xy}} - \boldsymbol{p}_{\mathrm{dy},i}(t_k) \big\|$.  $l_{\mathfrak{m},\mathrm{dy}}$ is defined as:
\begin{equation}
\begin{aligned}
    &l_{\mathfrak{m},\mathrm{dy}}(\bar{\boldsymbol{x}}_{\mathfrak{m},k}^{*},\bar{\boldsymbol{x}}_{\mathfrak{b},k}) \\
    &\ \ = \omega_{\mathrm{dy}}\sum_{i}\sum_{j=1}^{\mathcal{L}}\sum_{m=1}^{N_j} \max \Big( 0,\, D_{\mathrm{dy},j,m,i}(\bar{\boldsymbol{x}}_{\mathfrak{m},k}^{*},\bar{\boldsymbol{x}}_{\mathfrak{b},k}) \Big)^2.
\end{aligned}
\end{equation}

\section{Appendix F: Task-trajectory Consistency}
\label{app:front_end}
% \subsection{Trajectory Verification Between Keypoints}
When the search expands to a node corresponding to a subsequent keypoint $\mathfrak{p}_{\mathfrak{n}}$ within the same task (implying the existence of a predecessor $\mathfrak{p}_{\mathfrak{n}-1}$ in the same task), we must verify that the transition does not deviate excessively from the nominal task trajectory (e.g., staying within a tolerance region around the linear path for drawer opening).

Let $\mathcal{N}_{\mathrm{curr}}$ be the current node with the computed IK solution $\boldsymbol{q}_{\mathrm{end}}$ for keypoint $\mathfrak{p}_{\mathfrak{n}}$ with grasp pose $\,^{\mathrm{o}}\boldsymbol{P}_{\mathcal{C}}$. Let $\mathcal{N}_{\mathrm{prev}}$ be the ancestor node in the search tree corresponding to the previous keypoint $\mathfrak{p}_{\mathfrak{n}-1}$, which stores a set of valid configurations $\boldsymbol{Q}_{\mathrm{valid}}$. 
The transition is considered feasible if there exists a starting configuration $\boldsymbol{q}_{\mathrm{start}} \in \boldsymbol{Q}_{\mathrm{valid}}$ corresponding to the same grasp pose $\,^{\mathrm{o}}\boldsymbol{P}_{\mathcal{C}}$ that satisfies the constraints. The verification process for a candidate $\boldsymbol{q}_{\mathrm{start}}$ proceeds as follows:

\begin{enumerate}
    \item \textbf{Path Reconstruction:} We trace the mobile-base path segments from $\mathcal{N}_{\mathrm{curr}}$ back to $\mathcal{N}_{\mathrm{prev}}$ within the Hybrid A* search tree. Let this discrete sequence of base poses be denoted as $\mathcal{P}_{\mathrm{base}} = \{\boldsymbol{x}_{\mathfrak{b}}^{(0)}, \dots, \boldsymbol{x}_{\mathfrak{b}}^{(K)}\}$, where $\boldsymbol{x}_{\mathfrak{b}}^{(0)}$ corresponds to $\mathcal{N}_{\mathrm{prev}}$ and $\boldsymbol{x}_{\mathfrak{b}}^{(K)}$ to $\mathcal{N}_{\mathrm{curr}}$.
    
    \item \textbf{Synchronized Sampling:} We uniformly sample $K$ steps along the desired task trajectory between keypoints $\mathfrak{p}_{\mathfrak{n}-1}$ and $\mathfrak{p}_{\mathfrak{n}}$, and then calculate the task end-effector poses using $\,^{\mathrm{o}}\boldsymbol{P}_{\mathcal{C}}$, denoted as $\{\boldsymbol{P}_{\mathrm{task}}^{(k)}\}_{k=0}^K$. Simultaneously, using the candidate $\boldsymbol{q}_{\mathrm{start}}$, we perform linear interpolation in the joint space to obtain a sequence of arm configurations $\{\boldsymbol{q}_{\mathrm{arm}}^{(k)}\}_{k=0}^K$, where $\boldsymbol{q}_{\mathrm{arm}}^{(k)} = \boldsymbol{q}_{\mathrm{start}} + \frac{k}{K}(\boldsymbol{q}_{\mathrm{end}} - \boldsymbol{q}_{\mathrm{start}})$.
    
    \item \textbf{Whole-Body Validation:} For each step $k$, we compute the actual end-effector pose $\boldsymbol{P}_\mathrm{ee}^{(k)}$ using forward kinematics based on the combined state of the base pose $\boldsymbol{x}_{\mathfrak{b}}^{(k)}$ and arm configuration $\boldsymbol{q}_{\mathrm{arm}}^{(k)}$.
    
    \item \textbf{Constraint Check:} 
    We compute the deviation $e_k$ between the actual end-effector pose $\boldsymbol{P}_\mathrm{ee}^{(k)}$ and the target pose $\boldsymbol{P}_{\mathrm{task}}^{(k)}$. The deviation is defined as a weighted sum of translational and rotational errors:
    \begin{equation}
    \begin{aligned}
        e_k &= \|[\boldsymbol{P}_\mathrm{ee}^{(k)}]_{\boldsymbol{p}} - [\boldsymbol{P}_{\mathrm{task}}^{(k)}]_{\boldsymbol{p}}\|\\
        &\ + \lambda \cdot \angle([\boldsymbol{P}_\mathrm{ee}^{(k)}]_{\boldsymbol{R}}, [\boldsymbol{P}_{\mathrm{task}}^{(k)}]_{\boldsymbol{R}})
    \end{aligned}
    \end{equation}
    where $\angle(\cdot, \cdot)$ represents the geodesic distance (rotation angle) in $SO(3)$, and $\lambda$ is a weighting factor. The transition is valid if $\max_{k} (e_k) < \delta_{\mathrm{thresh}}$.
\end{enumerate}
If a valid solution is found for $\boldsymbol{q}_{\mathrm{end}}$, the IK solution $\boldsymbol{q}_{\mathrm{end}}$ is accepted; otherwise, it is discarded.
% We iterate this process over $\boldsymbol{Q}_{\mathrm{valid}}$. 

\end{document}